\documentclass[final,3p,times,twocolumn,authoryear]{elsarticle}

\usepackage{amssymb}
\usepackage{graphicx}
\usepackage{subfig}
\usepackage{float}
\usepackage{overpic}
\usepackage{amsmath, amsthm, amssymb, graphicx}
\usepackage{booktabs, multirow}
\usepackage{caption}
\usepackage{stfloats}
\usepackage{color}
\usepackage{hyperref}
\usepackage{CJKutf8}
\hypersetup{
    colorlinks=true,
    linkcolor=blue,
    filecolor=blue,      
    urlcolor=blue,
    citecolor=blue,
}

\usepackage{lineno}

\makeatletter
\def\ps@pprintTitle{%
     \let\@oddhead\@empty
     \let\@evenhead\@empty
     \def\@oddfoot{\footnotesize\itshape\hfill\today}%
     \let\@evenfoot\@oddfoot}
\makeatother

\begin{document}
\begin{CJK}{UTF8}{gbsn}
\begin{frontmatter}



\title{A Radiometric Correction based Optical Modeling Approach to Removing Reflection Noise in TLS Point Clouds of Urban Scenes}


\author[label1,label5]{Li Fang}
\ead{fangli@fjirsm.ac.cn}
\author[label1]{Tianyu Li}
\ead{swccj1998@gmail.com}
\author[label2,label5]{Yanghong Lin\corref{cor1}}
\ead{linyanghong21@mails.ucas.ac.cn}
\cortext[cor1]{Corresponding author}
\author[label1]{Shudong Zhou}
\ead{zsd@fjirsm.ac.cn}
\author[label3,label4]{Wei Yao}
\ead{wei.yao@ieee.org}

\address[label1]{Quanzhou Institute of Equipment Manufacturing Haixi Institutes, Chinese Academy of Sciences, Quanzhou, 362200, China}
\address[label2]{Fujian Institute of Research on the Structure of Matter, Chinese Academy of Sciences, Fuzhou, 350002, China}
\address[label3]{School of Engineering and Design, Technical University of Munich, Munich, 80333, Germany}
\address[label4]{GeoBIM \& GeoNexus Intelligence, Munich, 80997, Germany}
\address[label5]{University of Chinese Academy of Sciences, Beijing, 100049, China}
             
             

\begin{abstract}
In the critical domain of computer vision, encompassing 3D reconstruction, autonomous driving, and robotics, point clouds are extensively employed. However, point clouds acquired via terrestrial laser scanner (TLS) frequently exhibit numerous virtual points when scans involve highly reflective surfaces. These virtual points substantially disrupt subsequent processes. This study introduces a reflection noise elimination algorithm for TLS point cloud models. For the first time, we propose an innovative reflection plane detection algorithm based on a geometry-optical model coupled with physical properties. This algorithm directly identifies and categorizes reflection area points according to optical reflection theory. To preserve the reflection feature invariance of virtual points, we have adapted the LSFH feature descriptor. The RE-LSFH descriptor, which retains directional information, partially mitigates the interference from highly symmetrical and self-similar architectural structures in the virtual point removal algorithm. By integrating the Hausdorff feature distance, the algorithm's resilience to ghosting and deformation effects of virtual points is bolstered, thus enhancing the accuracy of virtual point detection. We conducted extensive experiments on the newly constructed 3DRN benchmark dataset, which encompasses virtual TLS reflection noise in diverse and intricate real-world urban environments, e.g., complex glass architectures, highly repetitive and symmetric components, and so on. Compared to 2D projection-based methods, our proposed algorithm improves the precision and recall rates for identifying 3D points in reflective regions by 57.03\% and 31.80\%, facilitating reflection noise removal through accurate detection of complete reflection planes. Furthermore, the experimental results confirm the effectiveness of our approach, attaining approximately 9.17\% better outlier detection rate and 5.65\% higher accuracy in the dataset compared to the leading method. The 3DRN dataset can be accessed publicly at \href{https://github.com/Tsuiky/3DRN}{https://github.com/Tsuiky/3DRN}.

\end{abstract}


\end{frontmatter}


\section{Introduction}

In computer vision, 3D point clouds are widely used for their simplicity, flexibility, and powerful representation capability \citep{POLEWSKI2017118,HUANG2021310,WANG202389}. The development of 3D scanning devices, such as Kinect and light detection and ranging(LiDAR), makes the acquisition of point clouds easier and more efficient, providing a data base for fields such as virtual reality \citep{wirth2019pointatme}, robotics \citep{kim2018slam}, and autonomous driving \citep{zheng2022global}. In the field of urban 3D modeling \citep{li2017hierarchical,chen20223d}, point clouds are obtained by terrestrial laser scanners(TLS), which simplifies the process of surface modeling and geometric reconstruction and has proven to be one of the most suitable data sources for mapping urban scenes. However, the captured point clouds are inevitably disturbed by noise and outliers due to the limitation of the scanner and the reflective properties of the object surface, etc. 

\begin{figure}[htbp]
    \centering
    \includegraphics[width=0.99\hsize, height=0.5\hsize]{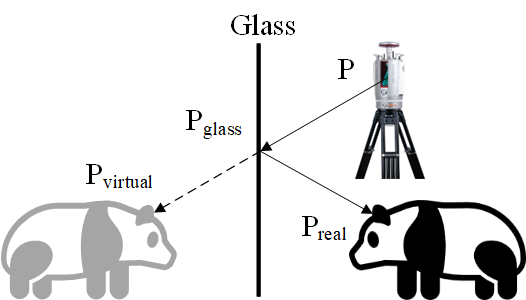}
    \caption{ The principle of reflection in 3D point clouds captured by TLS. Laser beams emitted by TLS bounce  off glass surfaces, producing 3D virtual points in the captured point cloud that do not exist in the real world.}
    \label{fig:panda}
\end{figure}
\begin{figure*}[!b]
    \centering
    \subfloat[]{\includegraphics[width=1.0\hsize, height=0.25\hsize]{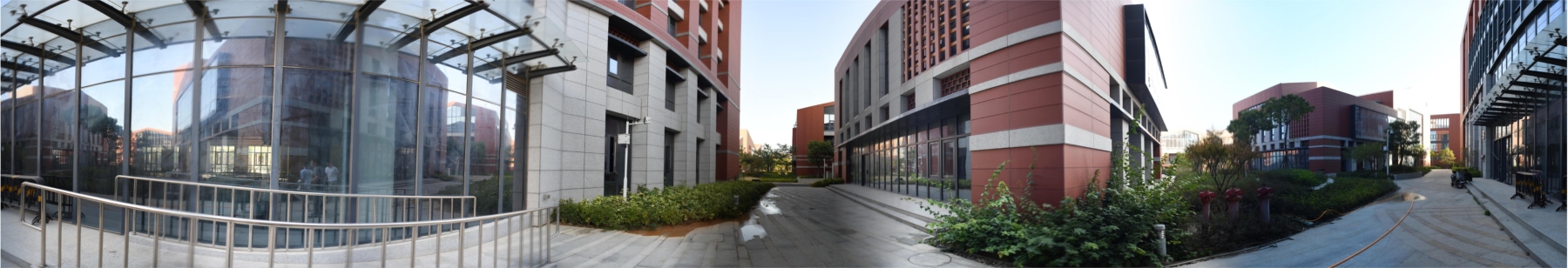}\label{fig:1a}}
    
    \subfloat[]{\includegraphics[width=1.0\hsize, height=0.5\hsize]{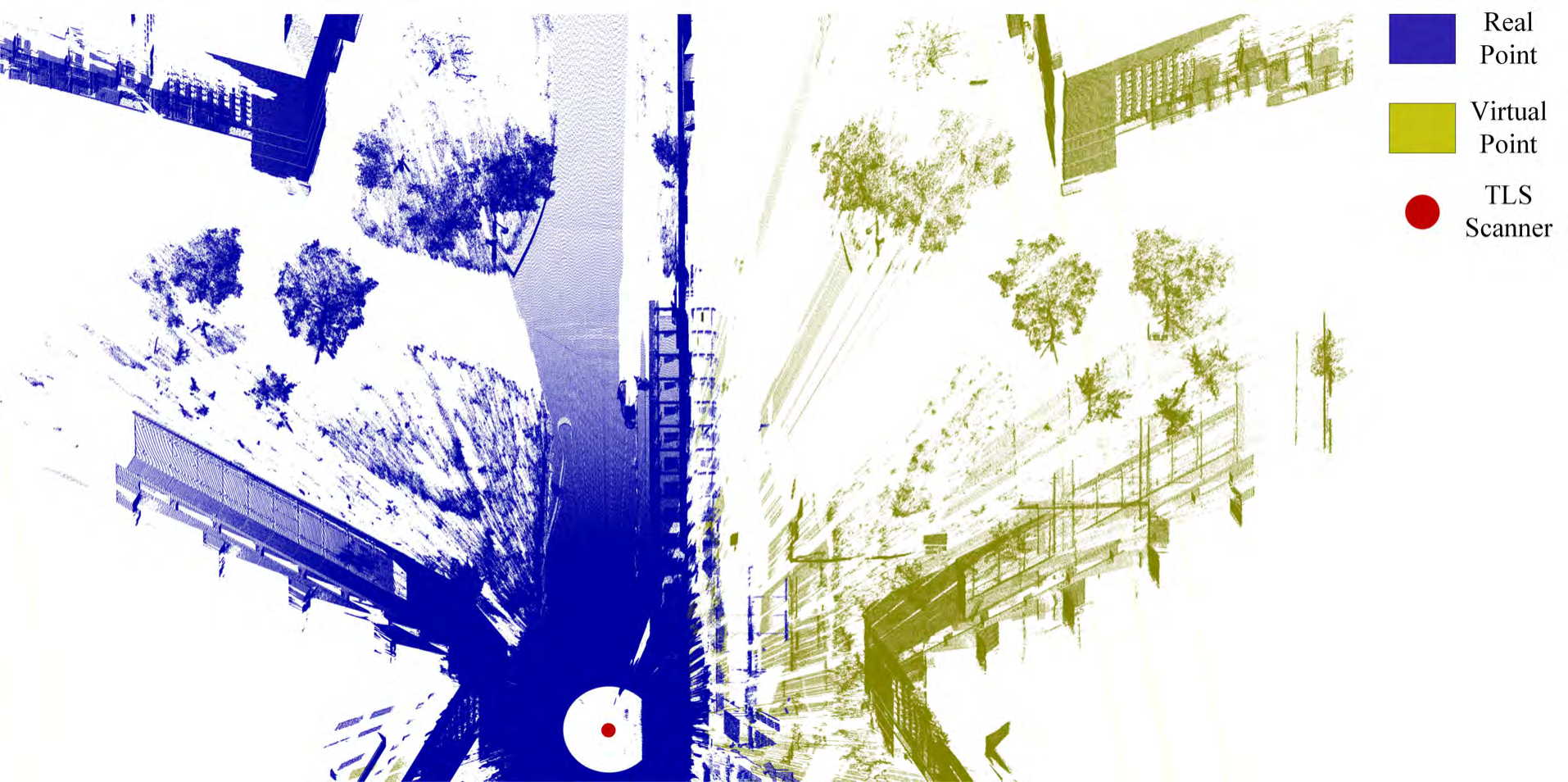}\label{fig: 1c}}
    \caption{Illustration of reflection noise in RGB panorama image and TLS point cloud. (a) Reflective noise in an RGB panorama image. (b) Reflective noise in a large-scale TLS point cloud is structured to exhibit geometric shapes and semantic information similar to those of real points.}
    \label{fig:1}
\end{figure*}

A variety of denoising algorithms have been developed to filter raw point clouds for downstream applications, with most focusing on specific types of noise or outliers \citep{10.1007/978-981-10-1721-6_31}. However, since reflective materials such as glass become more common in modern cities, a new type of reflection noise has drawn attention. Attempts \citep{RN70,RN72,RN73} have been undertaken to suppress reflection noise in images. Most of these methods view the image as a linear combination of transmission layer and reflection layer, then separate them to achieve reflection noise removal. However, the reflection noise in point clouds has not been effectively addressed. Fig.\ref{fig:panda} shows how reflection noise is generated. When a laser pulse hits the glass, part of it passes through and reaches the actual object behind it, whereas another part is reflected by the glass and hits the real object in front of it, creating a return pulse. Upon receiving the return pulse, the scanner is unaware of the presence of the glass and therefore calculates the distance as the sum of the distances from the scanner position $P$ to the glass $P_{glass}$ and from the glass $P_{glass}$ to the real object $P_{real}$, resulting in a virtual point behind the glass. Highly reflective surfaces are common in urban environments as a standard feature of modern architecture, making reflection noise a frequent issue in urban scene point clouds captured by TLS. The presence of this reflection noise significantly degrades the TLS data quality and presents a substantial challenge for subsequent processing tasks.

To deal with the reflection noise in point cloud, we first assume that, without the obstruction from the reflection, we can take a real point set, $B\in R^{n_1\times 3}$, and then model the reflection noise contaminated point cloud $P\in R^{n\times 3}$ as the union of $B$ and the virtual point set (called reflection noise) $R\in R^{n_2\times 3}$.

Noise and outliers differ from points on the actual surface in terms of distribution, density, etc. \citep{berger:hal-01017700}, which is a fundamental assumption of most current point cloud filtering algorithms. In contrast, reflection noise in point clouds is structural and shares similar point distribution characteristics, density, and semantic information with the real object that is reflected (Fig.\ref{fig:1}). Consequently, at these levels, existing point cloud filtering algorithms consistently fail to eliminate dense reflection noise. Several initial studies on reflection noise have been conducted \citep{RN60,RN71,RN63,RN64}. All of these studies involve projecting 3D point clouds into 2D panoramas to detect reflective areas and then applying some filtering criteria, offering new perspectives for reflection noise removal. However, these methods have not directly explored reflection plane estimation in the 3D space of LiDAR and still face significant challenges in effectively removing reflection noise. The completeness and precision of reflective plane estimations significantly affect the effectiveness of reflective noise removal. A challenge arises when 2D reflective pixels from panoramas are converted into 3D reflective area points based on the assumption that they are nearest to the scanner. This assumption is often ambiguous in multi-echo point clouds, leading to numerous outliers in the 3D reflective area points. Additionally, the resolution limitations of the panorama lead to a relatively sparse extraction of reflective area points, particularly in large-scale scenes. High outlier rates and sparse reflective area points result in incomplete and inaccurate orientation and position estimates for the reflective plane subsequently. When a reflective plane is incomplete or mispositioned, the method struggles to detect the virtual points it creates, thereby decreasing the accuracy of reflection noise removal. Furthermore, developing a feature descriptor suitable for reflection noise removal remains an open challenge. The descriptor must meet unique requirements, such as reflection invariance \citep{RN63} and the ability to adequately distinguish highly homogeneous building facades in urban scenes. Furthermore, multiple reflections caused by thick glass have already attracted attention in 2D imaging \citep{RN74}. However, existing methods in 3D point clouds assume ideal conditions, often ignoring factors such as the thickness and unevenness of reflection planes when faced with realistic scenarios. This study is motivated by these research gaps and aims to address the challenges of accurately estimating reflection planes and removing reflective noise in TLS point cloud data.

 Building upon prior research, we propose a novel approach for filtering dense reflection noise in TLS point clouds grounded in optical reflection theory. Initially, we present a reflection area detection module from the standpoint of the physical properties of laser reflection, enabling dense and reliable detection of reflective areas directly in 3D space. Subsequently, we develop an optical physical reflection model based on the identified reflective planes. In this model, virtual points are treated as geometrically similar to the actual object being reflected and symmetrical relative to the reflection plane. Driven by the identified research gaps, we created a compact, mirror-invariant feature descriptor that preserves orientation information. Additionally, we analyze and account for the phenomena of multiple refractions caused by thick glass and noise distortion due to uneven reflective planes. Consequently, we employ a novel virtual point removal module to assess the virtual point score for each query point.

To sum up, the main contributions of this paper are as follows:
\begin{itemize}
\setlength{\itemsep}{0pt}
\setlength{\parsep}{0pt}
\setlength{\parskip}{0pt}
  \item A novel reflection surface detection module based on radiometric correction model operates directly on the 3D point cloud space for preventing unreliable detection caused by compression of 3d data structure;
  \item An improved virtual point elimination module that integrates reflection-invariant and orientation-aware RE-LSFH with Hausdorff distance, aimed at minimizing disruptions from highly symmetrical and self-similar architectural structures and deformation virtual points;
  \item A new benchmark TLS dataset for testing of virtual point elimination is introduced. This dataset includes 12 large-scale urban LiDAR point clouds, featuring complex glass structures with various transparency and shapes, and highly repetitive and symmetrical elements.
\end{itemize}
\section{Related Work}

\subsection{Traditional Point Cloud Denoising}
Denoising has long been a conventional focus in computer vision. Noise points not only add to the volume of point cloud data but also compromise the precision of modeling and information extraction. The primary sources of this noise are systemic and environmental factors, encompassing random and systematic errors, as well as environmental conditions like rain, snow, and dust.

To eliminate noise, the point cloud can be handled using various 3D spatial filters \citep{han2017review}. Additionally, different clustering techniques can be applied, such as Principal Component Analysis \citep{duan2021low,mattei2017point} and DBSCAN \citep{schubert2017dbscan}. Since noise points typically have a much lower density compared to the point cloud itself, \cite{swatantran2016rapid} identifies noise points by assessing the local density. Point cloud denoising encompasses different categories of algorithms. PDE-based methods \citep{RN145,RN156} utilize kernel density estimation techniques for point clustering and filter point clouds using mean shift. \cite{RN152,RN148}, drawing inspiration from non-local image filtering techniques based on similarity, proposed a new non-local similarity measure that evaluates similarity using not only the position or normal of two surface points but also by examining the surrounding area of the vertices. \cite{RN158} presents a new framework for processing point-sampled objects based on spectral methods. By employing Fourier transform to create a spectral decomposition of the model, their method denoises by analyzing spectral coefficients. Projection-based methods \citep{RN154,RN153,RN155} have also gained increasing attention recently. Their primary concept is to adjust the position of points in point clouds by defining projection operators, thereby achieving point cloud denoising. Traditional statistical techniques \citep{barnett1994outliers,maimon2005data,rousseeuw2011robust} are also employed in the denoising domain. \cite{RN147} models the measurement process and prior assumptions about the measured object as probability distributions and then applies Bayesian statistics to generate smooth point clouds from noisy ones. \cite{RN145} defines a smooth likelihood function and locates each point on the smooth surface by moving each sample from the noisy dataset to the maximum likelihood position.

\subsection{Learning-based point cloud denoising}
With the advancement of deep learning, numerous efforts in point cloud denoising are being introduced. PointCleanNet \citep{rakotosaona2020pointcleannet} is the pioneer in utilizing the local neighborhood of point clouds for supervised training, assuming that the denoising results derive solely from points within the local vicinity. PointASNL \citep{yan2020pointasnl} introduces an adaptive sampling module that enhances noise resistance during sampling. In the realm of unsupervised learning, \cite{hermosilla2019total} accomplishes point cloud denoising by extending unsupervised image denoising techniques to unstructured 3D point clouds. Given that many existing denoising methods lead to over-smoothing, \cite{zeng20193d} maintains the sharp features of the point cloud using a low-dimensional manifold model. To explicitly identify noise points and recover the surface, \cite{luo2020differentiable} reconstructs the point cloud through manifold reconstruction. \cite{hu2020feature} proposes a fully convolutional neural network based on graph convolution, which effectively handles irregular domains and permutation invariant issues typical of point clouds. A novel approach for denoising point clouds based on the noisy point cloud distribution model is introduced by \cite{luo2021score}. \cite{mao2022pd} integrates normalizing flows and noise disentanglement techniques to achieve superior denoising accuracy. Additionally, some research \citep{zhou2019dup,li2023high,alliegro2021denoise} focuses on denoising in the context of upsampling and completing point clouds.

\subsection{Reflection noise removal for TLS point clouds}

Work closely related to ours includes \cite{RN63,RN64} and \cite{RN71,RN60}. \cite{RN63} employs the LiDAR multiple echo reflection mechanism and suggests a reflection noise removal algorithm for a single glass plane. Building on \cite{RN63}'s work, \cite{RN64} identifies multiple reflections caused by multiple glass planes. However, their methods require corresponding real and virtual points to detect reflective areas. If these corresponding points are missing, reflective plane detection will fail. Our proposed algorithm for reflection plane detection leverages information such as laser echo intensity to directly identify reflection areas without needing corresponding real and virtual points. The algorithms proposed by \cite{RN63,RN64} are based on point clouds from multiple echo data. \cite{RN71,RN60} suggest reflection noise removal algorithms for single echo data. \cite{RN71} uses a sliding window and a multi-site strategy to detect virtual points. However, this strategy of identifying virtual points through multiple sites is ineffective for legacy point cloud data, particularly in scenarios with low overlap rates. \cite{RN60} introduces a deep learning approach, utilizing a transformer network to label noise areas and eliminating the need to set manual parameters.

\begin{figure*}[htbp]
    \centering
    \includegraphics[width=0.96\hsize, height=0.58\hsize]{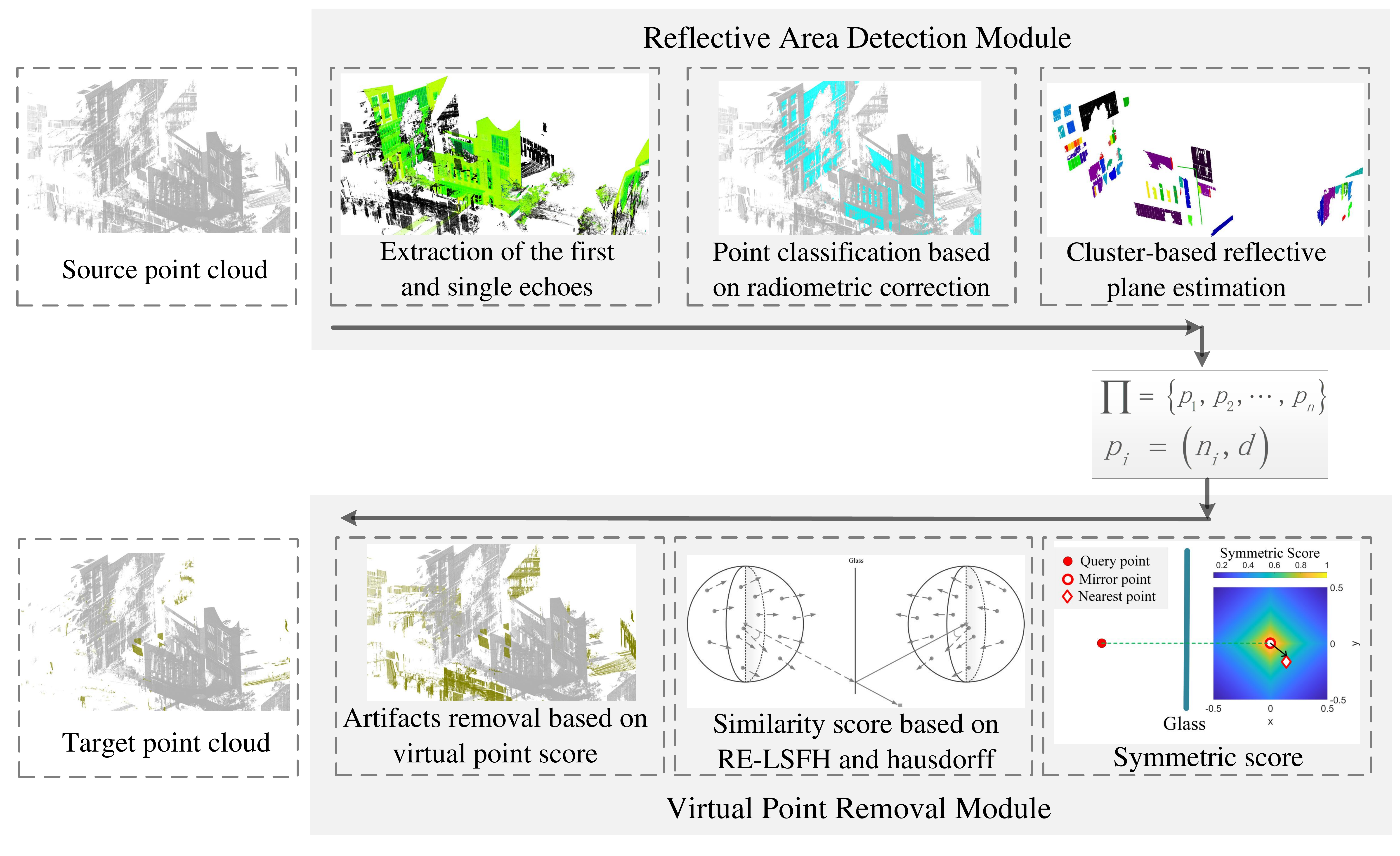}\label{fig:sub_figure1}
  \caption{The workflow of the proposed method features a reflective region detection module for precise estimation of reflective surfaces and a virtual point detection and removal module aimed at maximizing the removal of virtual points while maintaining real ones.}    \label{fig:2}\end{figure*}

\section{Proposed Method}

\subsection{Overview}

A novel method for eliminating virtual points in TLS point clouds is presented, which detects reflective surfaces and removes virtual reflective laser points while preserving the maximum number of real points.

The denoising algorithm described in this paper comprises a reflective surface detection module and a virtual point detection and removal module (Fig.\ref{fig:2}). The reflective surface detection module relies on physical principles related optical reflection, which identifies reflection area points directly in the 3D space and then adaptively estimates reflective planes associated with virtual points. The virtual point detection and removal module exploits the symmetry and geometric similarity between virtual points and their corresponding real points. A reflection-invariant RE-LSFH is introduced to be used in conjunction with Hausdorff distance for geometric similarity measurement, enabling a more detailed local shape description and enhanced robustness against virtual point deformation. Ultimately, points with high virtual point scores are identified as virtual points and automatically removed.

\subsection{Reflection area detection}\label{glass detect}

This module aims to identify reflective surfaces by utilizing point cloud attributes, as illustrated at the top of Fig.\ref{fig:rad}. The module is divided into two stages: extraction of reflective area points and estimation of the reflection plane.

\subsubsection{Extraction of reflection area points}
\begin{figure*}[htbp]
    \centering
    \includegraphics[width=0.96\hsize, height=0.463\hsize]{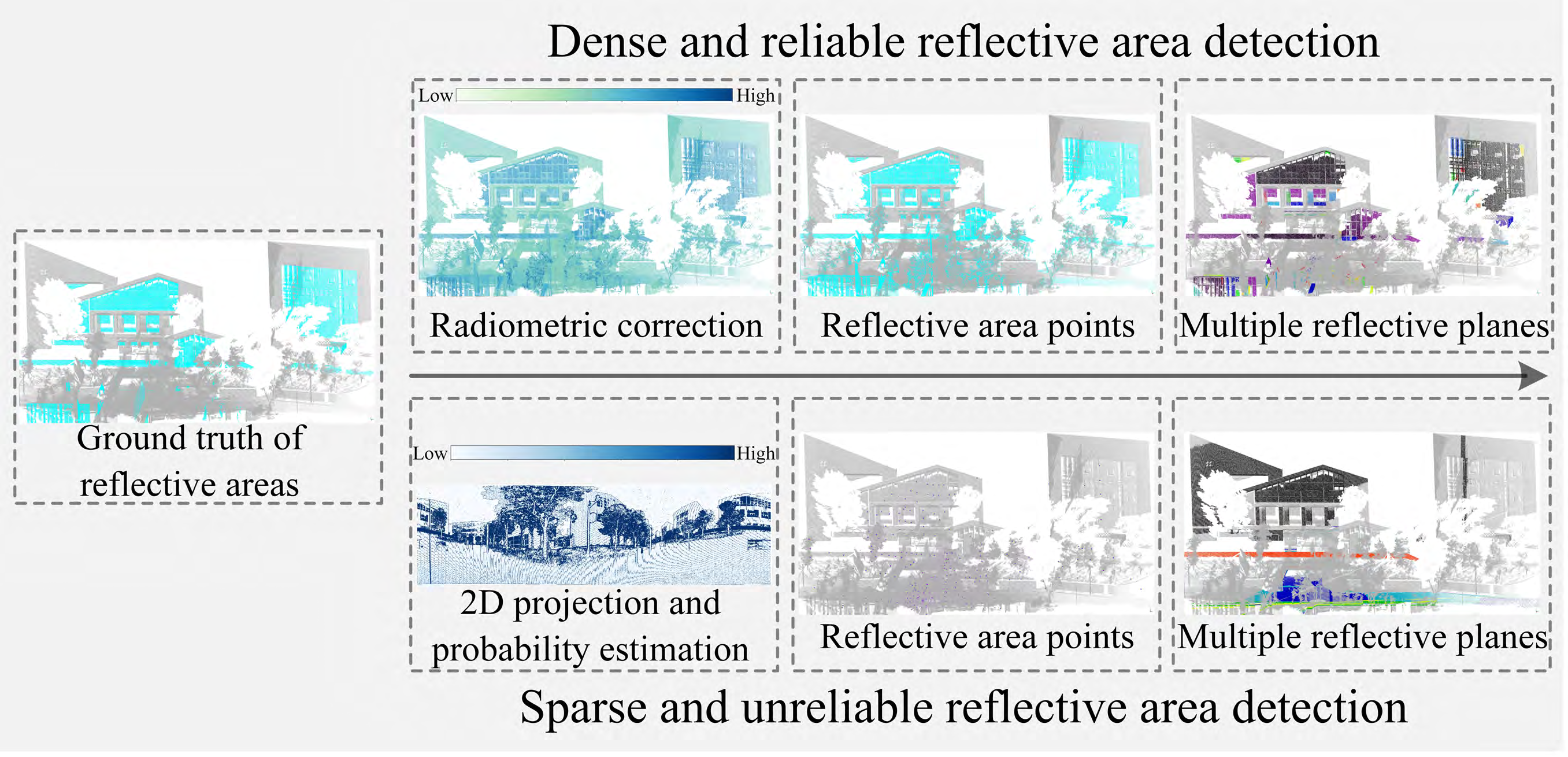}
  \caption{Comparison of the proposed reflective area detection module (top) with the 2D projection-based approach (bottom). By taking full advantage of optical properties, the proposed module can directly extract dense points of possible reflective areas from 3D space, enabling accurate reflective plane estimation.}    \label{fig:rad}
\end{figure*}

As illustrated at the bottom of Fig.\ref{fig:rad}, the detection of reflective areas using 2D projection methods (e.g., \cite{RN64}) is limited by the panorama resolution in extensive scenes. This constraint leads to the extraction of sparse points in reflective regions, causing unreliable estimation of reflective planes. By leveraging optical properties, we directly extract dense points from reflective regions in 3D space, enabling reliable estimation of reflective planes.

Intensity is an additional attribute provided by LiDAR to measure the power of the received reflected echoes. However, the TLS scanner is not specifically designed to record actual reflection intensity, and the so-called intensity is recorded only to control the quality of the point cloud. Therefore, a pre-processing step is necessary before utilizing the intensity information provided by the scanner. The reflectance $\rho$ corresponds to the ratio of the power of the reflected beam to that of the incident beam. For a fixed wavelength, it is an intrinsic property of the target surface, depending on various parameters such as surface color and material properties. The reflectance and intensity recorded by the scanner are neither equal nor proportional due to various factors. According to \cite{sanchiz2021radiometric}, two important external factors that affect intensity are distance $R$ and angle of incidence $\alpha$. The goal of radiometric correction is to remove the interference of these external factors so that the intensity values are related only to the surface properties of the target.

For an extended Lambertian target, the LiDAR equation can be classically simplified as follows:
\begin{align}
\phi_{r} = \phi_{i}\frac{D^{2}\rho\cos\alpha}{4R^{2}}\eta_{sys}\eta_{atm}    
\end{align}
where $\eta_{sys}$ denotes the system transmission factor, $\eta_{atm}$ represents the atmospheric transmission factor, $D$ is the diameter of the receiver aperture, and $\phi_{r}$ and $\phi_{i}$ denote the received and emitted signal power, respectively. In the case of TLS, $\eta_{atm}$ can be disregarded. The parameters related to the scanning system can be treated as constants $C$. Thus, the classical LiDAR equation can be further reduced to

\begin{align}
\phi_{r} = C\cdot \rho\cdot \cos\alpha \cdot  R^{-2} 
\label{simplied}
\end{align}

However, in real-world scenarios, most targets do not display ideal Lambertian surface properties. We consider the effects of distance and incidence angle on intensity to be independent of each other and rewrite Eq.\ref{simplied} as
\begin{align}
I_{raw}(\rho, \alpha, R) = I_{c}(\rho)\cdot f_{2}(\cos\alpha) \cdot  f_{3}(R) 
\end{align}
where $I_{raw}$ is the received echo intensity which is proportional to $\phi_{r}$, $I_{c}$, $f_{2}$ and $f_{3}$ respectively represents a function of $\rho$, $\alpha$ and $R$. We fit $f_{2}$ and $f_{3}$ sequentially using polynomial functions based on the experimental data:
\begin{align}
f_{2}(\cos \alpha )&=\sum_{k=0}^{N_{2}} \beta_{k}\cos^{k}\alpha \\  
f_{3}(R)&=\sum_{k=0}^{N_{3}} \gamma _{k}R^{k} 
\end{align}
Therefore, the corrected intensity $I_{c}$, considering the reference angle $\alpha_{s}$ and reference distance $R_{s}$, can be determined as
\begin{align}
I_{c} =I_{raw}\tfrac{f_{2}(\cos\alpha_{s})f_{3}(R_{s})}{f_{2}(\cos\alpha)f_{3}(R)} 
\end{align}

Following radiometric correction, an intensity threshold is established to isolate points with high reflectance. On natural surfaces, both diffuse and specular reflections occur concurrently, and the reflection intensity can be further broken down into diffuse and specular components. This implies that high reflectance can manifest in two forms: high diffuse reflectance and high specular reflectance, which are mutually exclusive. In laser-based remote sensing, however, the reflected laser is captured by the sensor in only one specific direction, not across the entire hemisphere. For high diffuse reflectance, the reflection adheres to Lambert's law, scattering the reflected beam uniformly in all directions, meaning only the diffuse component in one direction is detected by the sensor. According to the Bidirectional Reflection Distribution Function (BRDF), the ratio of reflected light to incident light in the received direction is $\rho \setminus\pi$, indicating that even perfectly diffuse surfaces will show low reflectance in the scanner. Therefore, we can infer that the extracted high-reflectance points are all points with high specular reflectance. However, we observe that trees also display high specular reflectance in addition to traditional specular materials, aligning with the findings in \cite{tian2021analysis} and remote sensing of vegetation. To mitigate the interference from leaves, we fit planes directly from the high reflectance points, as trees typically do not form planes due to their irregular structure. Additionally, to enhance computational efficiency, we extract points from the first and single echoes, which represent the surface information of the scan and may include reflective area points.

\subsubsection{Reflection plane estimation}
Former approaches generally estimated reflective planes from points within the extracted reflection region using iterative RANSAC. However, the number of reflective planes linked to virtual points in point clouds is unknown and must be manually determined in iterative RANSAC. Additionally, RANSAC selects plane parameters with the highest number of fitted points in each iteration, which is merely a statistical measure and may not be reliable for complex structures \citep{XU2019106}. In our work, exact boundaries of the reflective planes are not necessary, but the normal direction of the reflective planes is more critical. Drawing inspiration from \cite{RN200} and \cite{XU2019106}, our reflective plane estimation employs a bottom-up approach with a cluster-based incremental segmentation strategy. This method allows for adaptive determination of the number of reflective planes. The process includes clustering-based reflection plane estimation and merging of similar planar clusters.

During the reflective plane estimation phase, DBSCAN clustering is applied to the classification outcomes to segment the reflective area points into several clusters based on Euclidean distances and densities. The criteria for selecting clusters, namely smoothness and flatness, are determined by their linearity and curvature. These are computed using the eigenvalues obtained from the eigenvalue decomposition (EVD) of the 3D structure tensor of the point coordinates. The curvature and linearity of each cluster are derived from the eigenvalues of the cluster's covariance matrix. matrix:
\begin{align}
    \begin{split}
        \mathrm{curvature}&=\frac{e_3}{e_1+e_2+e_3}\\
        \mathrm{linearity}&=\frac{e_1-e_2}{e_1}
    \end{split}
\end{align}
where $\mathrm{e_k}$ denotes the k-th eigenvalue of the cluster covariance matrix. A lower $\mathrm{curvature}$ indicates a more planar surface. Similarly, a lower $\mathrm{linearity}$ suggests a more uniform area distribution within the planar cluster, leading to a more dependable plane orientation. Consequently, clusters with fewer points, higher curvature, or higher linearity (such as points misclassified in the first stage, like trees) are discarded. The remaining clusters are deemed reliable and are utilized for plane estimation. To improve the robustness of the plane estimation, we use RANSAC to determine the plane parameters for each reliable cluster individually, resulting in the set of reflective planes $\prod =\left \{ P_1,P_2, \cdots,P_{m-1}\right \} $, where $P_i=(n_i, d)$ specifies the parameters of the $i_{\mathrm{th}}$ plane, and $m$ represents the number of reliable clusters. $n_i$ is the normal of the plane determined by RANSAC, and $d$ is the perpendicular distance from the origin to the estimated plane.

During the phase of combining similar planar clusters, we aim to minimize the computational expense of later virtual point detection by merging clusters that have similar plane parameters. This bottom-up method results in several complete reflective planes. Adjacent clusters with comparable plane parameters are combined into a single cluster, using metrics such as the angle and the perpendicular distance from the centroid to the origin, calculated as follows:
\begin{align}
    \begin{split}
        \mathrm{cos_{\text {sim}}}&=n_i\cdot n_j\\
        \mathrm{dis_{\text {sim}}}&=\left \|n_i\cdot \overline{p_i} - n_j\cdot\overline{ p_j}  \right \| 
    \end{split}
\end{align}
where $n_i$ and $\overline{p_i}$ denote the normal vector and the centroid of the respective cluster plane. The adjacent planar clusters will be merged only if the aforementioned metrics are below the specified threshold. Ultimately, the reflective planes within the point cloud that could produce virtual points are fully identified.

In contrast to iterative RANSAC-based approaches that necessitate manually specifying the number of reflective planes to estimate, the proposed method can dynamically determine both the number and size of the reflective planes extracted, and it can also decrease the number of iterations when exact planar boundaries are not essential.
\begin{figure*}[htbp]
    \centering
    \subfloat[]{\includegraphics[width=0.57\hsize, height=0.22\hsize]{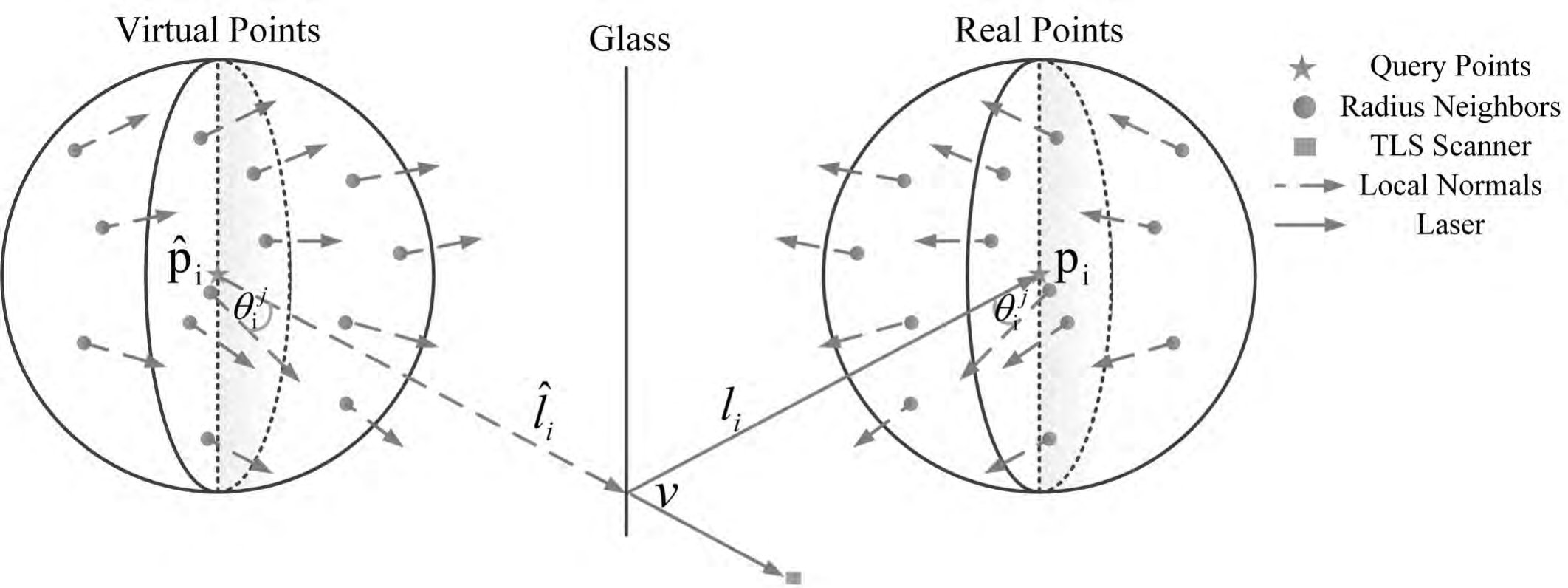}\label{fig:5a}}
    \subfloat[]{\includegraphics[width=0.43\hsize, height=0.22\hsize]{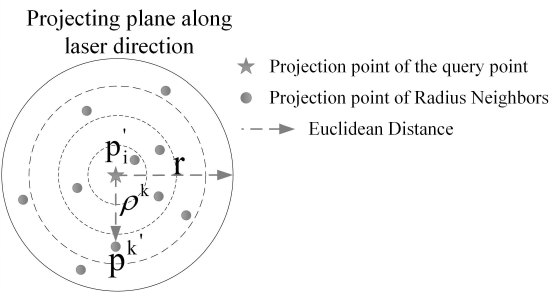}\label{fig:5b}}
  \caption{Construction of the RE-LSFH feature descriptor. (a) Angle of deviation between the laser path and the normals. (b) Projection distance from the query point to the neighboring radii along the laser path.}
  \label{fig:6}
\end{figure*}

\subsection{Virtual Point Detection and Removal}

We integrated reflection-invariant RE-LSFH descriptor with Hausdorff feature distance to assess local geometric similarity. By utilizing directional information from incident and reflected laser beams, this descriptor overcomes the limitations of traditional methods such as FPFH, which struggle to distinguish building facades in different directions. The feature distance measures the overall shape disparity between local point clouds, offering enhanced robustness against ghosting and deformation effects.

\subsubsection{RE-LSFH Feature Descriptor}

FPFH, which ignores directional information, becomes less effective in virtual point detection, as facades with varying orientations yield similar feature responses.

In this study, we construct an improved RE-LSFH feature descriptor with the local reference axes of the direction of the incident laser beams for query points (refer to $l_i$ for $p_i$ in Fig.\ref{fig:5a}) and reflected laser beams for symmetry points (refer to $\hat{l_i}$ for $\hat{p_i}$ in Fig.\ref{fig:5a}) respectively. Inspired by \cite{YANG2016163}, the feature descriptor includes the angle feature calculated by the deviation angle between the incident laser beam and the local normals, the local point density calculated by projecting the point cloud along the direction of the laser beam onto the plane. This results in an improved RE-LSFH feature descriptor which is reflection-invariant and distinctive to various planes. The descriptor is used for the description of local geometric shapes.

Part 1 is the deviation angle feature, which is used to describe the surface variation of the local point cloud. See Fig.\ref{fig:5a}, the input point cloud $P=\left \{p_1,p_2,\cdots,p_N  \right \}$ with N points contains virtual points $\hat{A}$ and the corresponding real points $A$. Let us pick a query point $\hat{p_i}$ from $\hat{A}$ which is a virtual point corresponding to the real point $p_i$. For the query point $\hat{p_i}$, a radius neighbors of $\hat{p_i}$ is defined as $p^k_i=\left \{p^1_i,p^2_i,\dots,p^k_i\right \} $. The laser emitted by the scanner at $o$ is reflected at $v$ on the glass. $\hat{l_i}$ denotes the direction of incident laser calculated by $\hat{l_i}=\hat{p_i}-v$ and $l_i$ denotes the direction of reflect laser calculated by $l_i=p_i-v$. $n^k_i$ denote the local normal vectors of $p^k_i$. 

The calculation of the local normals uses the method of \cite{10.1145/777792.777840}, and for the radius neighbors, the covariance matrix is calculated as Eq.\ref{8}:
\begin{equation}
\label{8}
    \begin{split}
        \operatorname{Cov}\left(p_{i}\right)=\left[\begin{array}{c}
        p_{i}^{1}-\overline{p_{i}} \\
        \cdots \\
        p_{i}^{k}-\overline{p_{i}}
        \end{array}\right]^{T} \cdot\left[\begin{array}{c}
        p_{i}^{1}-\overline{p_{i}} \\
        \cdots \\
        p_{i}^{k}-\overline{p_{i}}
        \end{array}\right]
    \end{split}
\end{equation}
where $\overline{p_i}$ represents the centroid of $p^k_i$. The covariance matrix is obtained by eigenvalue decomposition, and the minimum eigenvector indicates the normal direction $n^k_i$ of $p^k_i$. Our method does not rely on the local reference frame (LRF), but uses a stable incident laser beam $\hat{l_i}$ (for query point) or a reflection laser beam $l_i$(for symmetric point) as the local reference axis (LRA) to construct the deviation angle between the local normal and the laser. For each neighboring point $p^k_i$ of the query point $\hat{p}$, the deviation angle between $n^k_i$ and $\hat{l_i}$ can be expressed as the Eq. \ref{111}, with a range of $\left [0,\frac{\pi}{2} \right ] $.
\begin{equation}
\label{111}
    \begin{split}
        \theta^j=\arccos(\hat{l_i}\cdot n^k_i)
    \end{split}
\end{equation}

Finally, the number of neighboring points within different deviation angle ranges is counted to obtain the local deviation angle feature. Likewise, the deviation angle feature of the symmetry point $p_i$ is also calculated with $l_i$ as LRA.

Part 2 focuses on the point density feature. Despite the proposed deviation angle feature being able to differentiate planes with various orientations, it remains relatively uniform and lacks sufficient informativeness in urban environments. Previous research \citep{6784345} has demonstrated that projection of 3D point cloud onto a 2D plane is an effective and efficient method to represent local geometric shapes. Projecting the local point cloud onto a 2D plane in different directions can yield distinct 2D point distributions. Due to the mirror symmetry between virtual points and their corresponding real points, similar local geometric shapes can be observed when viewed along the directions of the incident and reflected lasers. Consequently, the calculation of the point density feature in this study is illustrated in Fig.\ref{fig:5b}. The query point $\hat{p_i}$ and its radius neighbors $p^k_i$ are projected onto the plane in the direction of the incident laser $\hat{l_i}$, resulting in a new point ${\hat{p_i}}'$ and a new point set ${p^k_i}'$ (see Fig.\ref{fig:5b}). A circle with radius $r$ is then drawn with the projection point ${\hat{p_i}}'$ as the center to calculate the projection distance $\rho^k$ of the neighboring points ${p^k_i}'$ to the projection point ${\hat{p}}'$, which is calculated as\begin{equation}
    \begin{split}
        \rho^{k}=\sqrt{\left\|\hat{p_i}-p_{i}^{k}\right\|^{2}-\left(\hat{l_{i}} \cdot\left(\hat{p_i}-p_{i}^{k}\right)\right)^{2}}
    \end{split}
\end{equation}
where $\hat{p_i}$ and $p^k_i$ are the query point and neighboring point, respectively, and $\hat{l_i}$ represents the direction vector of the incident laser beam. The number of points within different projection distance ranges is counted to obtain the local point density feature. Likewise, the point density feature of the symmetry point $p_i$ is also calculated with $l_i$ as the projected direction.

\begin{figure*}[htbp]
	\centering
        \subfloat[]{
         \begin{minipage}[t]{0.32\linewidth}
                \centering
			\includegraphics[width=1\linewidth]{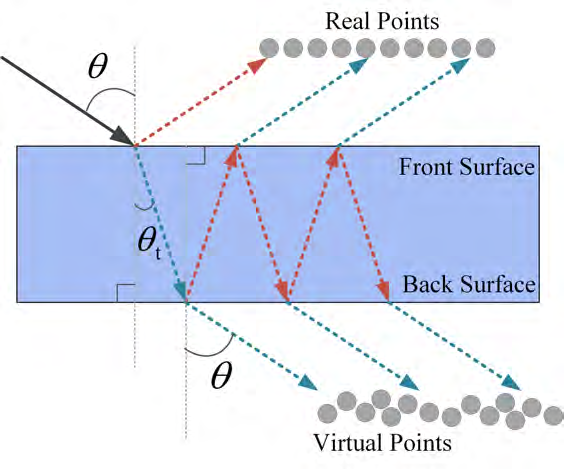} 
		\end{minipage}
		\label{fig:8a}
        }
	\subfloat[]{
            \begin{minipage}[t]{0.32\linewidth}
                \centering
			\includegraphics[width=1\linewidth, height=0.956\hsize]{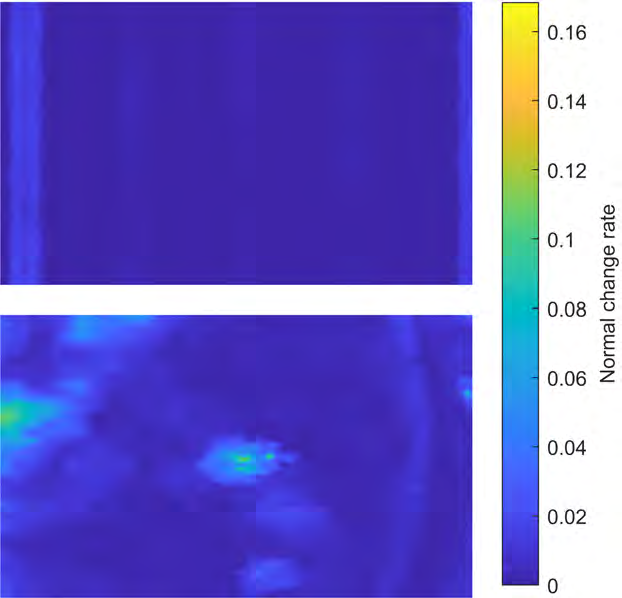} 
		\end{minipage}
		\label{fig:8b}
	}
	\subfloat[]{
		\begin{minipage}[t]{0.32\textwidth}
                \centering
			\includegraphics[width=1\linewidth, height=0.956\hsize]{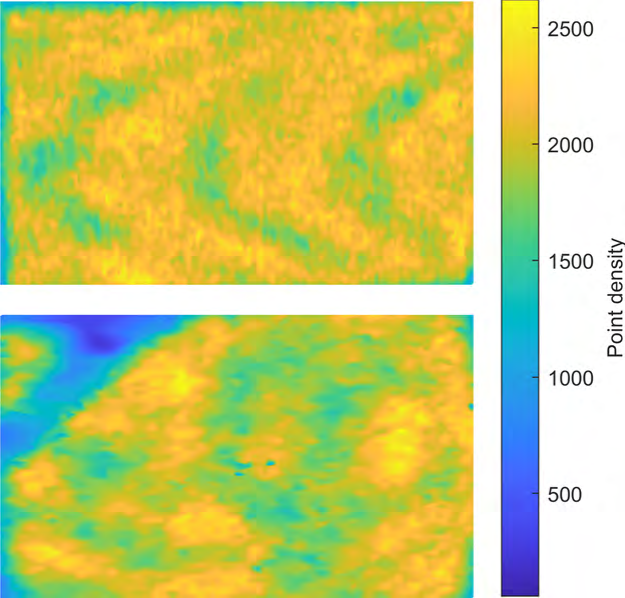} 
		\end{minipage}
		\label{fig:8c}
	}
	\caption{Depiction of various reflections and their impacts. (a) Diagram of multiple reflections. (b) Rate of local normal change for the flat facade (top) and its associated virtual points (bottom). (c) Point density distribution of the flat facade (top) and its corresponding virtual points (bottom). It is evident that real points on flat facades exhibit a low local normal change rate and a uniform point distribution, while the corresponding virtual points show significant variations.}
    \label{fig:4}
\end{figure*}

The deviation angle feature and the point density feature are aggregated through histograms to derive the feature vectors for bins $N_1$ and $N_2$, respectively. To ensure robustness against variations in point cloud density, the angle feature and the point density feature are individually normalized. Ultimately, the two subhistograms merge to form a $N_1+N_2$ bin RE-LSFH feature vector.

Note that the RE-LSFH proposed should be reflection-invariant. The point density feature of RE-LSFH is obviously reflection invariant. The reflection invariance of the angle feature can be proved by Eq.\ref{11}. For a pair of virtual point $\hat{p}$ and real point $p$ which satisfy $\hat{p}=A\cdot p$. Let $l_i$ and $l_j$ represent the laser incidence directions of $\hat{p}$ and $p$, which are the LRA for the RE-LSFH. $n_i$ and $n_j$ denote the normal vectors for the neighbors of $\hat{p}$ and its symmetric point $p$, respectively. Hence, the angle features of $\hat{p}$ and $p$ are calculated as follows.
\begin{align}
\label{11}
    \begin{split}
        \hat{\theta}&=\left \langle l_i,n_i \right \rangle \\
        \theta&=\left \langle l_j,n_j \right \rangle=l^T_j \cdot A \cdot n_j\\
        &=l^T_j \cdot A^T \cdot A \cdot n_j=l^T_i\cdot n_j\\
        &=\hat{\theta}
    \end{split}
\end{align}
Therefore, $\hat{\theta}=\theta$. This indicates that the deviation angle feature of RE-LSFH remains invariant under reflection. As a result, RE-LSFH demonstrates reflection invariance.

\subsubsection{Virtual point score}
The virtual point score is composed of two components: the symmetry score and the feature similarity score. The symmetry score evaluates the level of mirror symmetry between the query points $\hat{p}$ and the symmetric point $p$ relative to the reflective plane. The feature similarity score assesses the feature resemblance between the query points $\hat{p}$ and the symmetric point $p$. When both scores are high, a high virtual point score is observed, suggesting that the query point is more likely to be a virtual point.

\begin{figure}[t]
    \centering
    {
    \subfloat[]{\includegraphics[width=1.5in, height=1.2in]{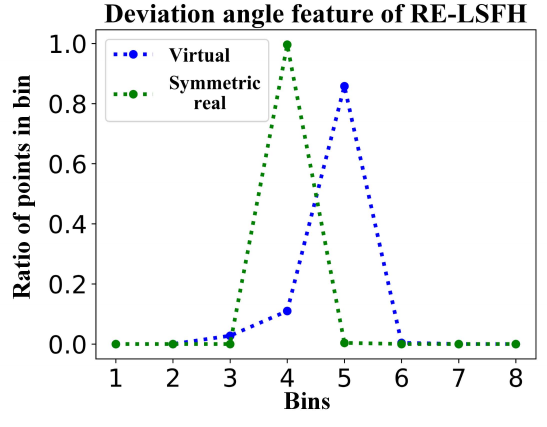}}
    \subfloat[]{\includegraphics[width=1.5in, height=1.2in]{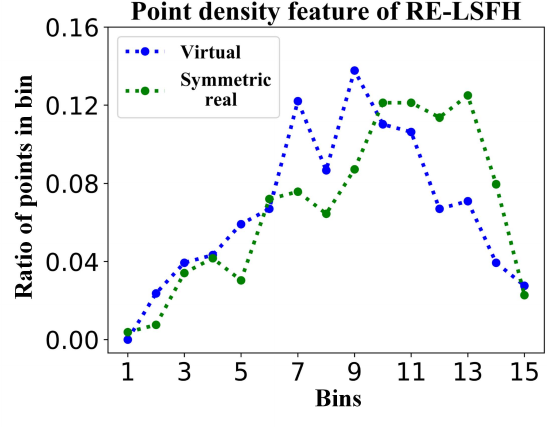}}
    }
    \caption{An example of RE-LSFH for a distorted virtual point and its corresponding real point. (a) Deviation angle feature of RE-LSFH (b) Point density feature of RE-LSFH. Note that both components have been L1-normalized.}
    \label{fig:8d}
\end{figure}

Nevertheless, reflection surfaces with a certain thickness cause multiple reflections, as illustrated in Fig.\ref{fig:8a}. These reflections lead to ghosting and deformation effects that alter the feature histogram of the virtual point. For a deformed virtual point and its corresponding real point from the flat facade (local geometric changes shown in Fig.\ref{fig:8b} and Fig.\ref{fig:8c}), their RE-LSFH are presented in Fig.\ref{fig:8d}. The ghosting and deformation effects shift the positions of the local peaks in the sub-histogram and decrease the overlap of the feature histograms of $\hat{p}$ and $p$, a phenomenon referred to as feature-drift in this study. Consequently, the KL distance, JS distance, and Hellinger distance are not suitable for directly measuring the distance between the two distributions due to the low overlap of the sub-histogram. These feature distances are highly susceptible to feature drift, leading to inaccurate estimations of virtual points.

\begin{figure}[t]
    \centering
    \subfloat[]{\includegraphics[width=1.5in, height=1.2in]{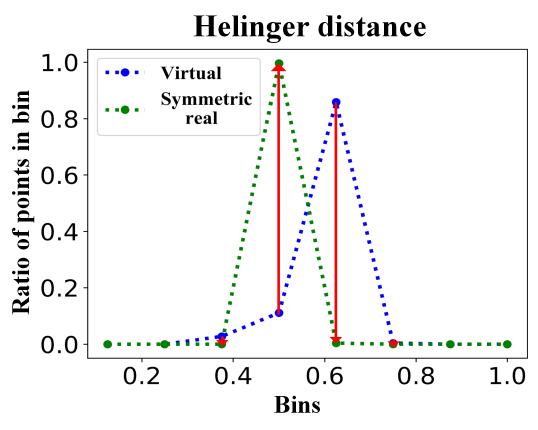}\label{fig:9a}}
    \subfloat[]{\includegraphics[width=1.5in, height=1.2in]{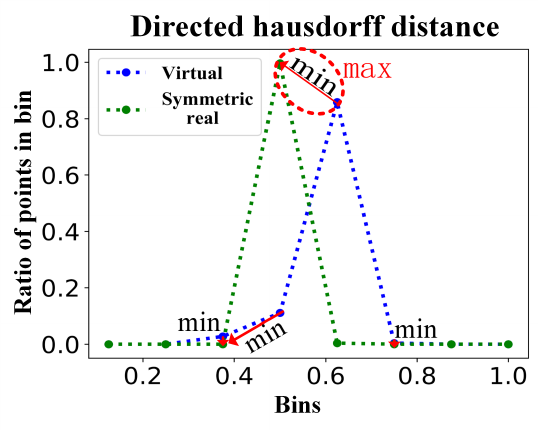}\label{fig:9b}}
  \caption{Comparison of the Helinger and Hausdorff distances between a virtual point and a real point. (a) The Helinger distance, shown by the clustering of red lines. (b) The directed Hausdorff distance, represented by a red circle.}    \label{fig:9}
\end{figure}

As shown in Fig.\ref{fig:8d}, the shape of the subhistogram remains fairly consistent despite feature drift. Both the distorted planar virtual points and the corresponding real planes exhibit angular feature sub-histograms with a single peak, and the point density feature sub-histograms rise to a peak and then fall. For the angular and point density features, the peak position differences between the virtual point and the real point on the x-axis are 1 bin and 3 bins, respectively. This suggests that the deviations caused by such effects are relatively minor in the feature histogram.

A minor deviation might decrease the overlap between features, making the KL distance and the Hellinger distance unsuitable for measuring feature similarity. This research also accounts for feature drift by using Hausdorff distances to quantify the maximum difference between feature histogram distributions as a measure of feature similarity. The Hausdorff feature distance captures the overall shape disparity between local point clouds, providing greater resilience to ghosting and deformation effects.

This study employs the Hausdorff distance to quantify the maximum discrepancy between the feature curves of the query point and its symmetric counterpart, aiding in achieving a more resilient feature similarity. Considering two feature descriptors (deviation angle feature or point density feature) for the query point and the symmetric point,
\begin{align}
    \begin{split}
        A=\left \{\alpha_i\mid i \in \left [ 1,N \right ]  \right \} \\
        B=\left \{\beta_i\mid i \in \left [ 1,N \right ]  \right \}
    \end{split}
\end{align}
where $A$ represents descriptor of the query point and $B$ represents descriptor of the symmetric point, $N$ is the number of dimensions of the feature histogram. The discrete Hausdorff distance between $A$ and $B$ is computed as follows:
\begin{equation}
    \begin{split}
        H(A,B)=\max(h(A,B),h(B,A))
    \end{split}
\end{equation}
where $h(A,B)$ is the directed Hausdorff distance from $A$ to $B$, defined as 
\begin{align}
    \begin{split}
        h(A,B)&= \max_{\alpha_i\in A}\min_{\beta_i \in B}\left \| \alpha_i-\beta_i \right \|\\
        h(B,A)&=\max_{\beta_i\in B}\min_{\alpha_i \in A}\left \| \beta_i-\alpha_i \right \|
    \end{split}
\end{align}
with $\left \| \alpha_i-\beta_i \right \|$ denoting Euclidean distance. Hence, the directed Hausdorff distance $h(A,B)$ ranks each bin of $A$ based on its Euclidean distance to the closest bin of $B$ and then takes the maximum $\left \| \alpha_i-\beta_i \right \|$ which actually comes from the bin of $A$ that is the most mismatched. The Hausdorff distance $H(A,B)$ is the maximum of the directed Hausdorff distances $h(A,B$ and $h(B,A)$. Note that, prior to calculation, both components of RE-LSFH have been individually normalized to the range of 0 to 1 along the horizontal axis (bins). Fig.\ref{fig:9} illustrates the Helinger distance and the Hausdorff distance between a pair of virtual point and real point, respectively, taking the deviation angle feature in Fig.\ref{fig:8d} as an example. The Hausdorff distance significantly reduces the feature distance between a pair of deformed virtual points and real points in this case.

For each query point $\hat{p}$, the final virtual point score is represented as
\begin{equation}
    \begin{split}
        Virtual(\hat{p})=\gamma_{\text sym}(\hat{p})\cdot \gamma_{\text sim}(\hat{p})
    \end{split}
\end{equation}
where $\gamma_{\text sym}(\hat{p})$ is the symmetric score, $\gamma_{\text sim}(\hat{p})$ is the geometric similarity score. The symmetric score represents the likelihood that the query point $\hat{p}$ is a mirror-symmetric point produced by the reflection of the glass surface $\prod$, and is calculated as
\begin{align}
    \begin{split}
        \gamma_{\text sym}(\hat{p})&=e^{-\frac{d}{\sigma}}\\
        &=e^{-\frac{\left \| p-\widetilde{p} \right \| }{\sigma}}
    \end{split}
\end{align}
where $d$ denotes the distance between the predicted location of the symmetric point $p$ and its nearest neighbor $\widetilde{p}$, with $\sigma$ being a constant. A higher symmetric score indicates a greater probability that the query point $\hat{p}$ is a symmetric point arising from the reflection on the glass surface.

$\gamma_{sim}(\hat{p})$ is the geometric similarity score. This score quantifies the resemblance of geometric characteristics between the query point and its symmetric counterpart, and is determined by
\begin{equation}
    \begin{split}
        \gamma_{\text sim}(\hat{p})=e^{-\frac{H(\hat{p},\widetilde{p}) }{\sigma}}
    \end{split}
\end{equation}
where H is the Hausdorff distance between the feature curves of the query point $\hat{p}$ and $\widetilde{p}$, and $\mu$ is a fixed value. A higher geometric similarity score indicates a higher similarity between the geometric features of the query point and its symmetric point.

\section{Experiments and results}

Extensive TLS point cloud data derived from actual environments are utilized for both qualitative and quantitative evaluation of the proposed algorithm. The testing environment consists of a desktop computer equipped with the following specifications: Windows 10 OS, Intel Core X processor, Nvidia RTX A6000 graphics card, and Python 3.8.

\subsection{Data acquisition}\label{data}
\begin{table}[b]
\centering
\caption{Parameter settings of TLS dataset.}
\label{table:para}
\begin{tabular}{@{}cc@{}}
\toprule
Scanner setting                & Parameter     \\ \midrule
Scan angle range (Vertical)   & 60°$\sim$-40° \\
Scan angle range (Horizontal) & 0°$\sim$360°  \\
Laser pulse repetition rate    & 1200kHz       \\
Scan range                 & 600m          \\
Scanning accuracy               & 5mm           \\ \bottomrule
\end{tabular}
\end{table}

\begin{figure*}[t]
    \centering
    \includegraphics[width=0.9\linewidth]{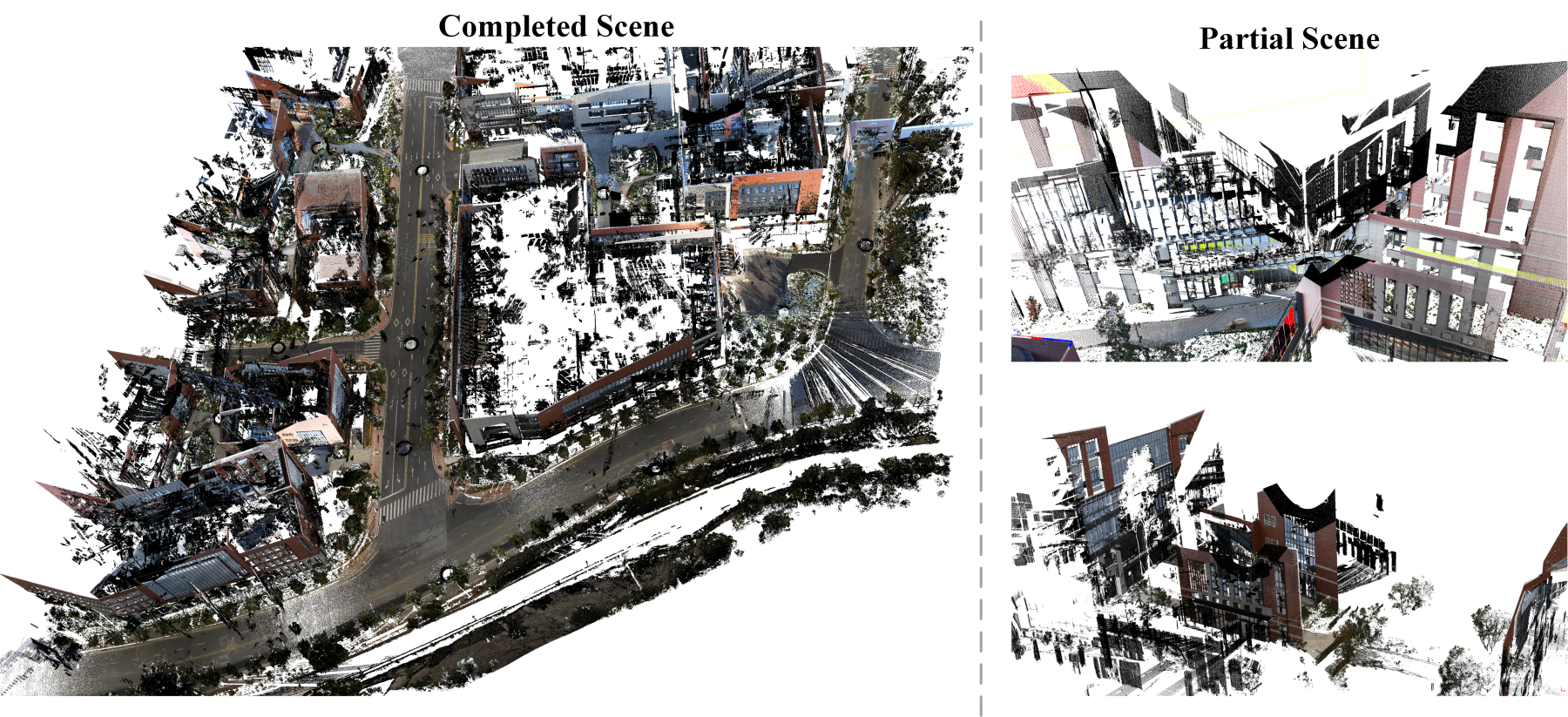}
  \caption{Sample images of the registered scene and partial scenes from Sanlinkou Innovation Park in our dataset, showing a considerable mix of virtual and real points.}    \label{fig:paronoma}
\end{figure*}

To the best of our knowledge, there is no publicly available benchmark TLS dataset for reflection noise removal. In this study, a new Beckmark dataset, namely 3DRN, is proposed for testing the reflection noise removal algorithm. The 3DRN contains 12 point cloud models with more than 55 million 3D points, collected by a terrestrial laser scanner (RIEGL VZ-2000i) with the parameter settings shown in Table \ref{table:para}. All point cloud models are captured from real urban scene containing highly reflective areas with significant reflection noise, half of which are from Sanlinkou Innovation Park and the other half from Quanzhou Equipment Center. Each point in the point cloud model provides its XYZ position information, RGB color information, reflectance intensity, echo order information, and ground truth of the virtual points annotated by the professional. Fig.\ref{fig:paronoma} shows a completed aerial view and partial scenes obtained after registration of all scan positions captured from the Sanlinkou Innovation Park, which clearly consists of real points and reflected virtual points.

\begin{table*}[htbp]
\centering
\caption{Details of TLS point cloud datasets, encompassing scene count, coverage area, total number of points, virtual points, and real points..}
\label{table:icpc}
\begin{tabular}{@{}ccccc@{}}
\toprule
   & Coverage(m) & Total points & Virtual points & Real points \\ \midrule
Scan 01 & $150.02\times124.27\times31.55$ & 6,021,813& 1,096,366& 4,925,447\\
Scan 02 & $194.32\times146.47\times35.01$ & 9,058,789& 197,734& 8,861,055\\
Scan 03 & $105.43\times172.41\times32.10$ & 8,319,624& 143,048& 8,176,576\\
Scan 04 & $102.04\times157.37\times31.29$ & 5,038,858& 1,638,517& 3,400,341\\
Scan 05 & $72.08\times122.68\times28.60$ & 3,342,466& 648,819& 2,693,647\\
Scan 06 & $147.35\times132.90\times24.91$ & 7,157,389& 627,539& 6,529,850\\
Scan 07 & $117.52\times119.71\times45.08$ & 3,509,231& 54,770& 3,454,461\\
Scan 08 & $103.27\times132.08\times55.30$ & 5,573,643& 244,010& 5,329,633\\
Scan 09 & $101.54\times49.47\times20.85$ & 1,612,160& 118,282& 1,493,878\\
Scan 10 & $48.60\times74.50\times28.15$ & 1,880,837& 412,547& 1,468,290\\
Scan 11 & $114.72\times79.55\times31.90$ & 1,942,031& 150,913& 1,791,118\\
Scan 12 & $31.80\times66.23\times9.24$ & 636,061& 38,490& 597,571\\ \bottomrule
\end{tabular}
\end{table*}

The proposed 3DRN dataset faces several challenges for the practical application of reflection noise removal algorithms: (1) Some scenes in the dataset exhibit highly repetitive and symmetrical features (for instance, multiple buildings within the same scene sharing identical structures and colors). (2) Buildings are situated closely together, leading to significant occlusions in the point cloud data. (3) Point cloud scenes include numerous irregularly shaped glass curtain walls. (4) Some real laser points corresponding to reflection-induced virtual points are occluded. (5) There are distortions between actual objects and their reflected virtual counterparts. (6) Glass transmittance differs between the Jinjiang Creative Entrepreneurship and Innovation Park dataset and the Quanzhou Equipment Manufacturing Research Center dataset, with the former having lower glass transmittance and the latter having higher glass transmittance. Research Center dataset, with the former having lower glass transmittance and the latter having higher glass transmittance.

RISCAN PRO is utilized for initial data processing. Initially, outliers (like airborne dust) were eliminated using a radius filter with a 0.5 radius and a 500 threshold. Subsequently, points exhibiting waveform deviations exceeding 25 were filtered out to eliminate filamentous noise caused by dense objects (such as foliage). Lastly, a voxel filter with a voxel size of 0.02 was applied to downsample the extensive point cloud, thereby reducing computational demands.

To test the performance of the virtual point removal algorithm, virtual points and real points are manually labeled to generate point cloud labels. Table \ref{table:icpc} introduces the specific information about the collected point cloud, including the point cloud scene number, coverage, total points, virtual points, and real points.

\subsection{Evaluation metrics}
For the quantitative assessment of the proposed virtual point removal algorithm, we adhere to the evaluation criteria outlined in \cite{RN71,RN60}. These criteria are used to compare the quantitative performance of the specified methods and include the outlier detection rate (ODR), inlier detection rate (IDR), false positive rate (FPR), and false negative rate (FNR). The evaluation metrics are defined as follows:
\begin{align}
ODR &=\frac{TN}{FP+TN} \\
IDR &=\frac{TP}{TP+FN} \\
FPR &=\frac{FN}{TP+FN} \\
FNR &=\frac{FP}{FP+TN} 
\end{align}
In this context, TP, FP, TN, and FN represent the counts of true positives, false positives, true negatives, and false negatives, respectively. Elevated ODR and IDR values guarantee that the majority of virtual points can be eliminated when searching for virtual points. Conversely, low FPR and FNR values emphasize the accuracy of virtual point detection and minimize the number of actual points that are wrongly removed. The accuracy, which combines these four ratios, reflects the overall effectiveness of the proposed method in eliminating virtual points. The formula for its calculation is as follows:\begin{align}
    Accuracy=\frac{TP+TN}{TP+FN+FP+TN} 
\end{align}
Furthermore, to assess the quality of the denoised point cloud, we propose an evaluation metric inspired by the peak signal-to-noise ratio concept \citep{schwarz2018common}. The formula for this calculation is given below.\begin{align}
SNR=10 \cdot \lg \frac{TP+FN}{FP+FN}  
\end{align}

\subsection{Result of Virtual Point Removal}\label{denoise exp}
We performed comprehensive experiments across two categories of indoor and outdoor environments.
\begin{figure*}[b]
  \begin{minipage}{0.24\linewidth}
     \vspace{3pt}  
     \centerline{\includegraphics[width=\textwidth]{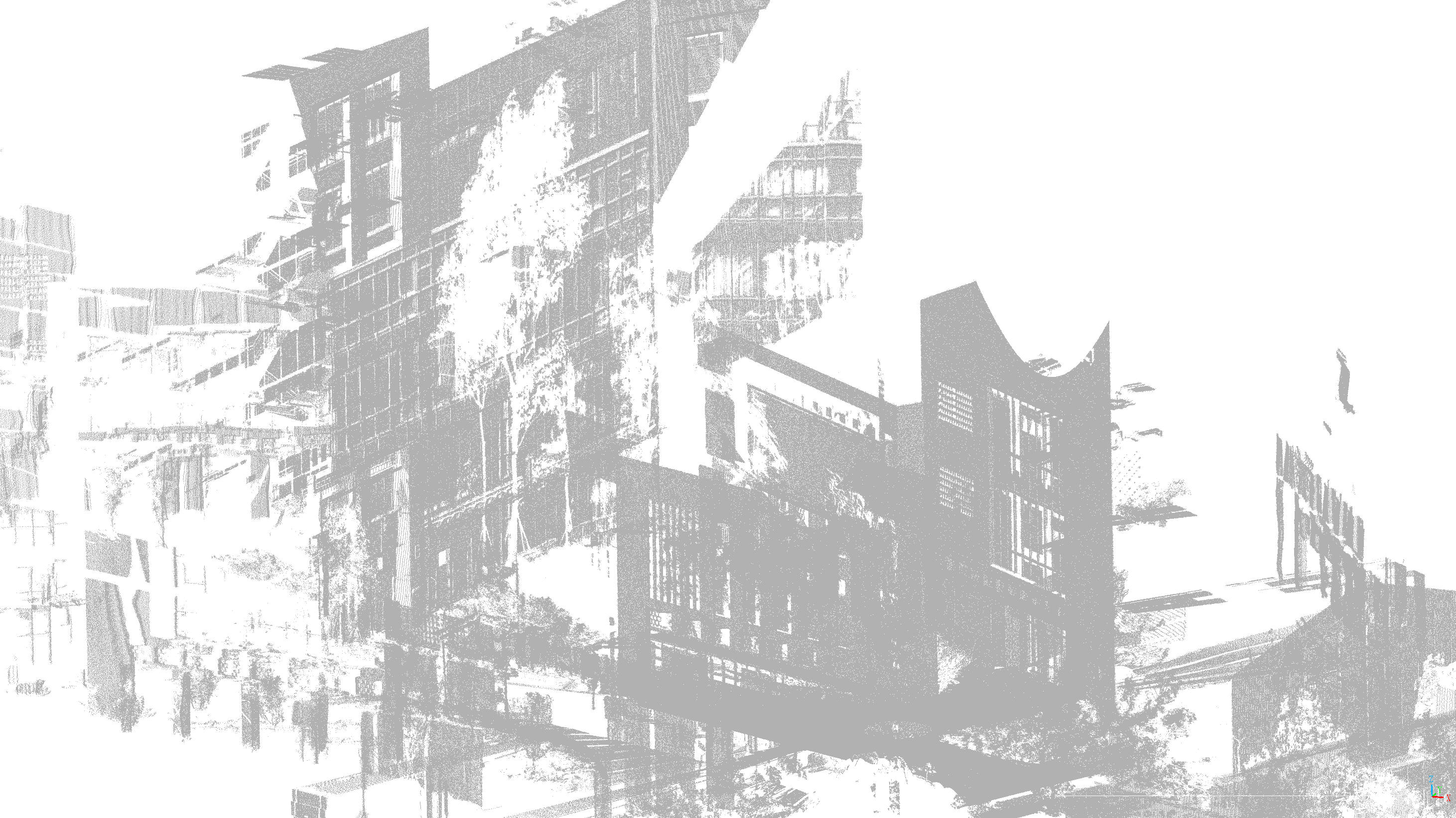}}
     \vspace{3pt}
     \centerline{\includegraphics[width=\textwidth]{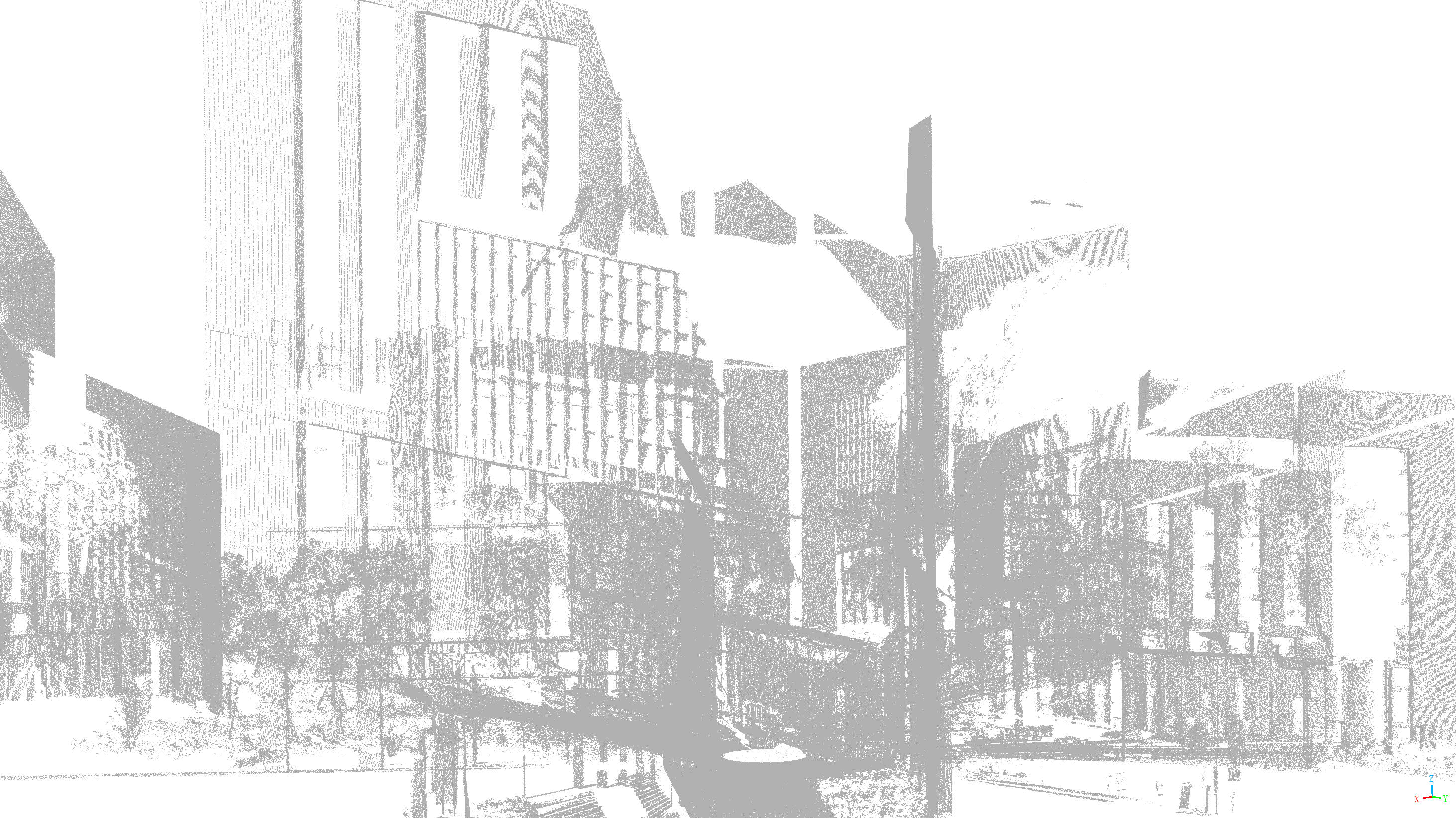}}
     \vspace{3pt}
     \centerline{\includegraphics[width=\textwidth]{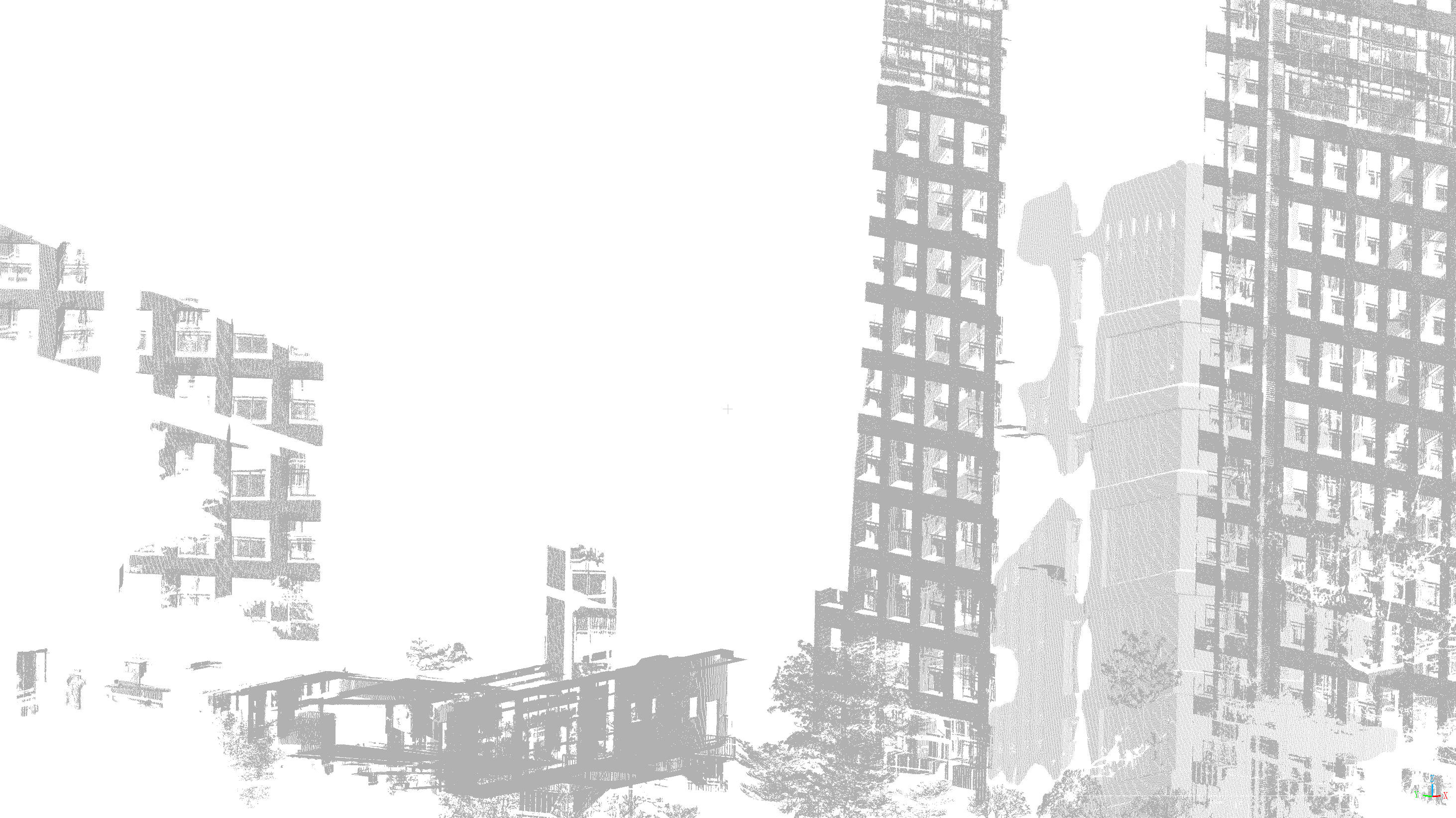}}
     \vspace{3pt}
     \centerline{\includegraphics[width=\textwidth]{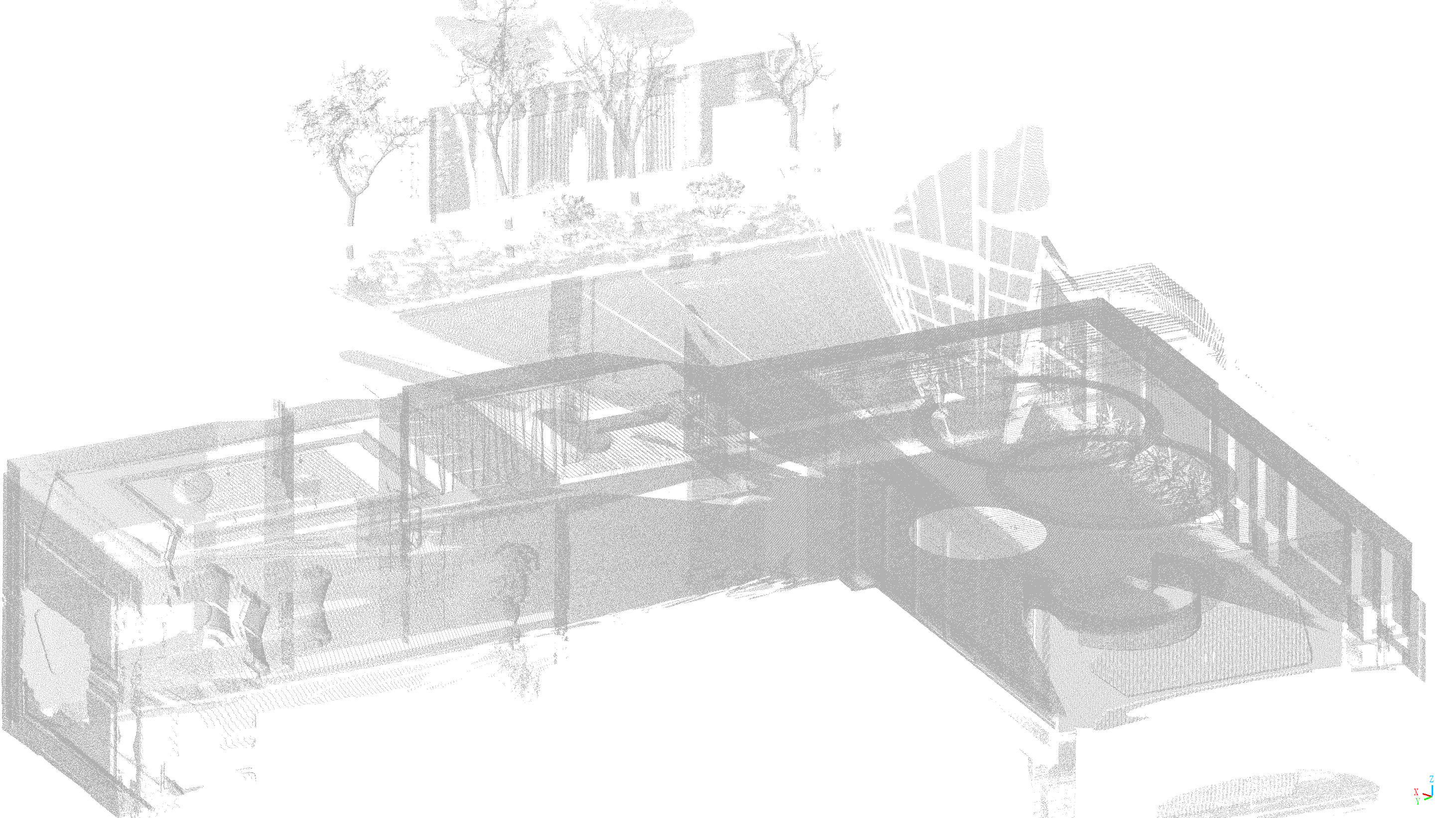}}
     \vspace{3pt}
     \centerline{(a)}
 \end{minipage}
    \begin{minipage}{0.24\linewidth}
     \vspace{3pt}
     \centerline{\includegraphics[width=\textwidth]{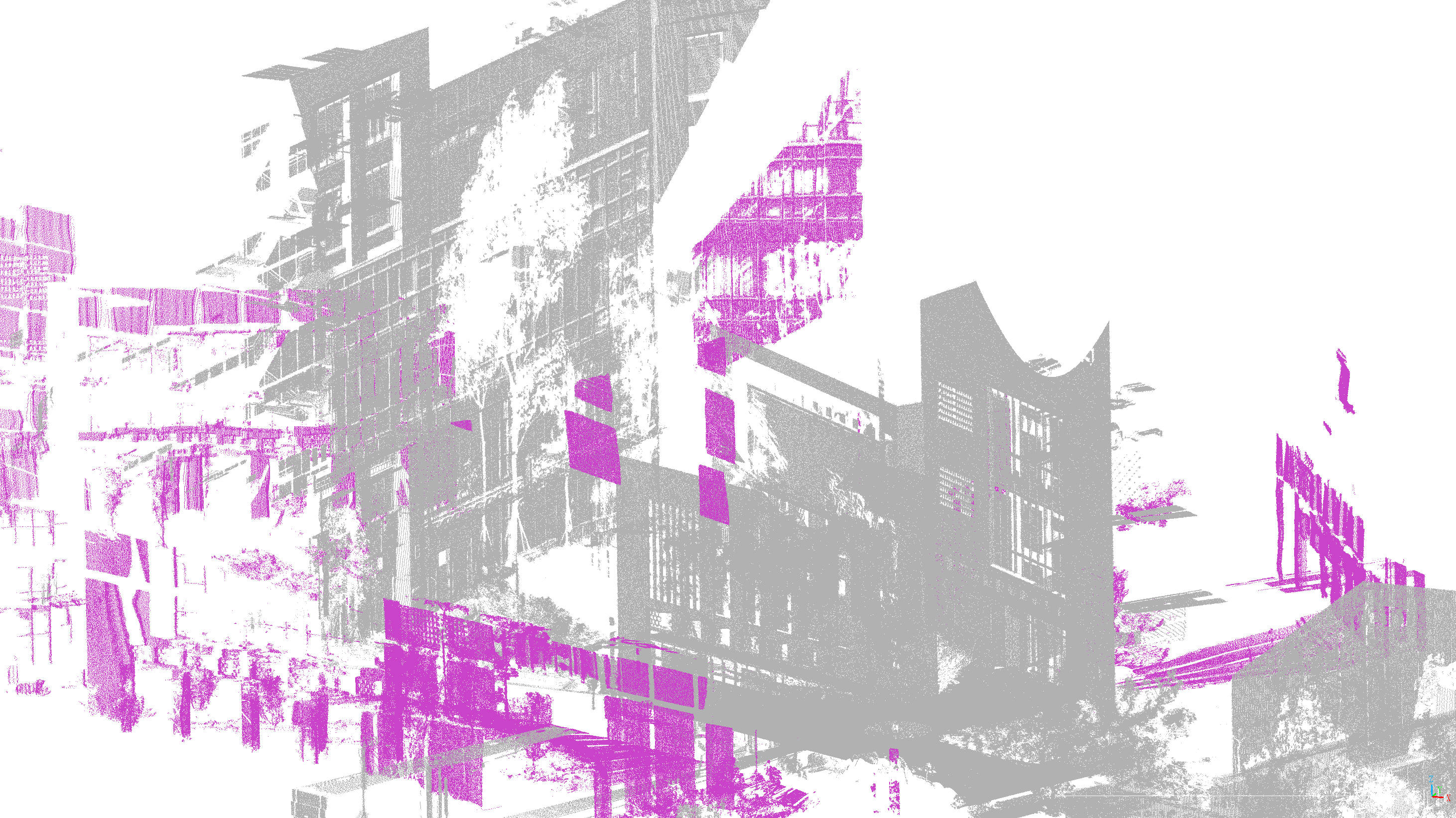}}
     \vspace{3pt}
     \centerline{\includegraphics[width=\textwidth]{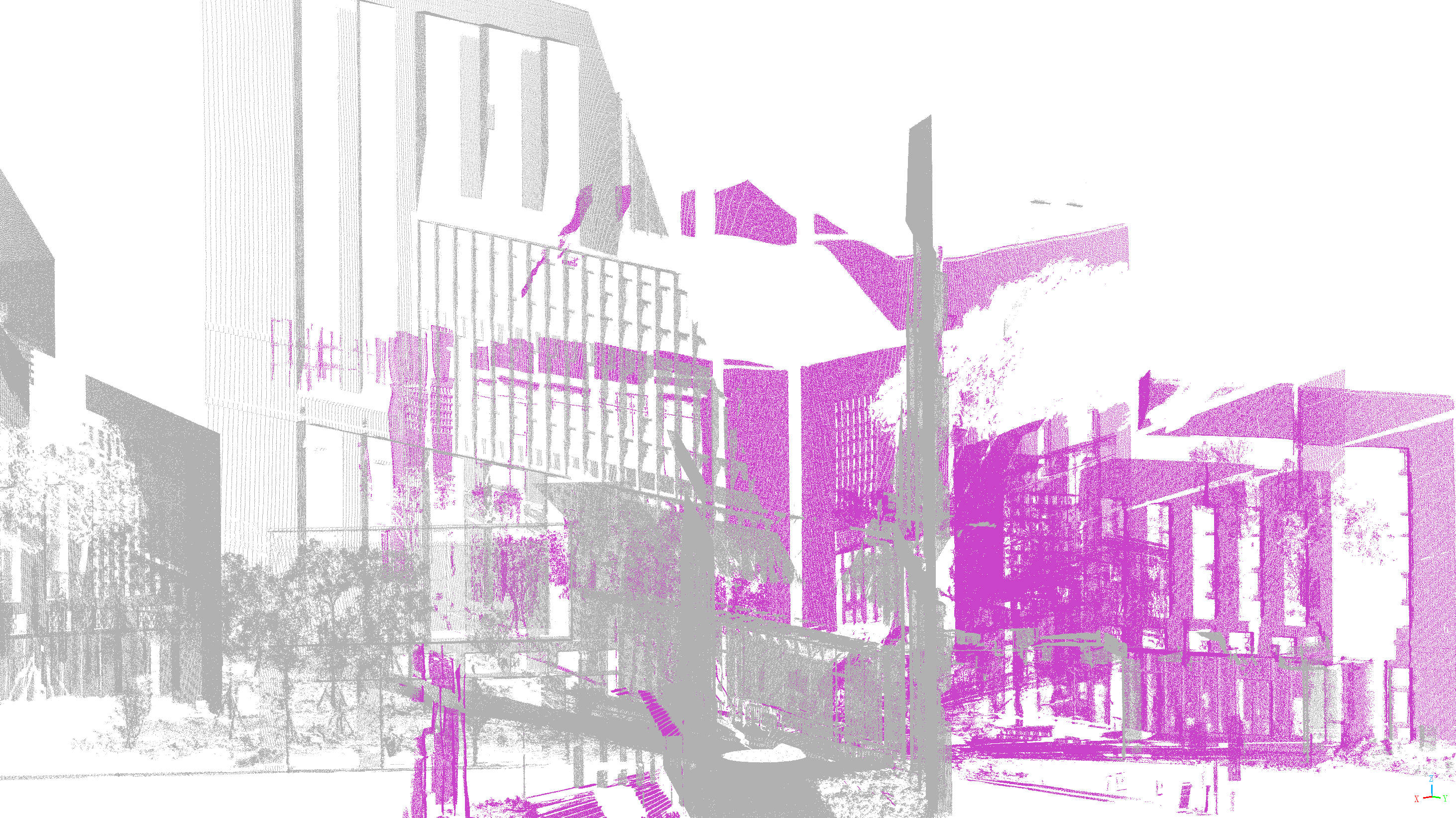}}
     \vspace{3pt}
     \centerline{\includegraphics[width=\textwidth]{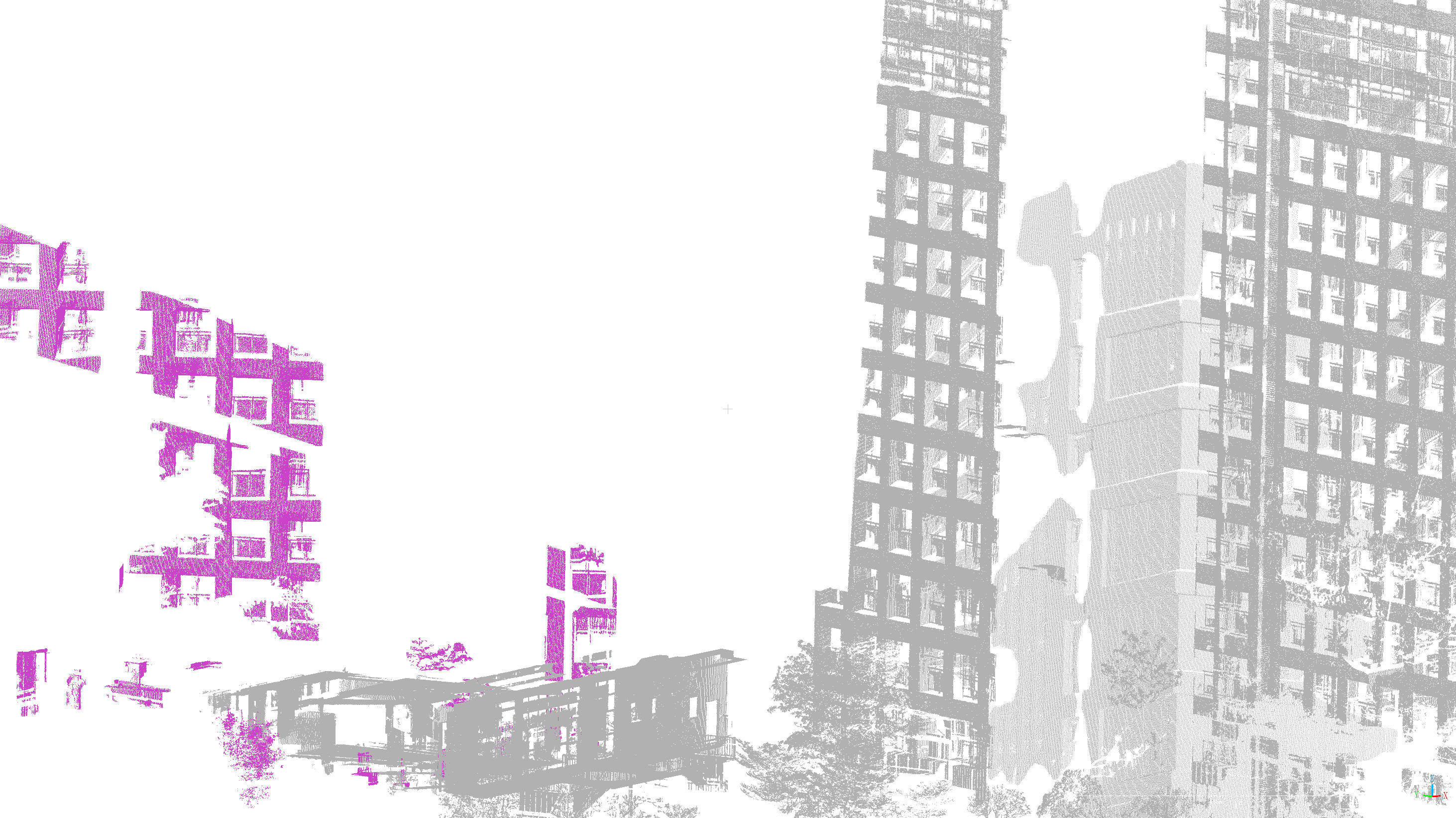}}
     \vspace{3pt}
     \centerline{\includegraphics[width=\textwidth]{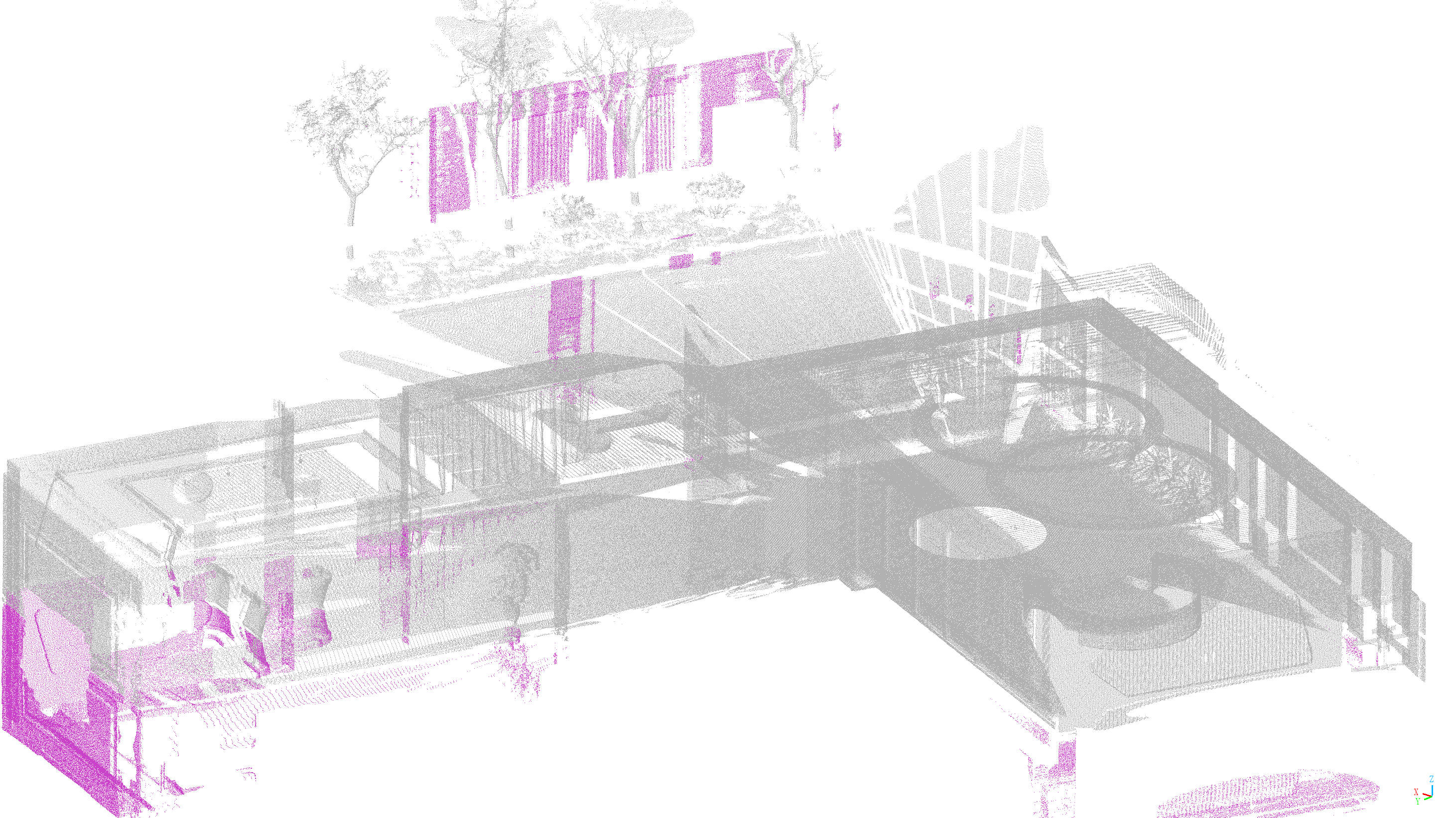}}
     \vspace{3pt}
     \centerline{(b)}
 \end{minipage}
    \begin{minipage}{0.24\linewidth}
     \vspace{3pt}
     \centerline{\includegraphics[width=\textwidth]{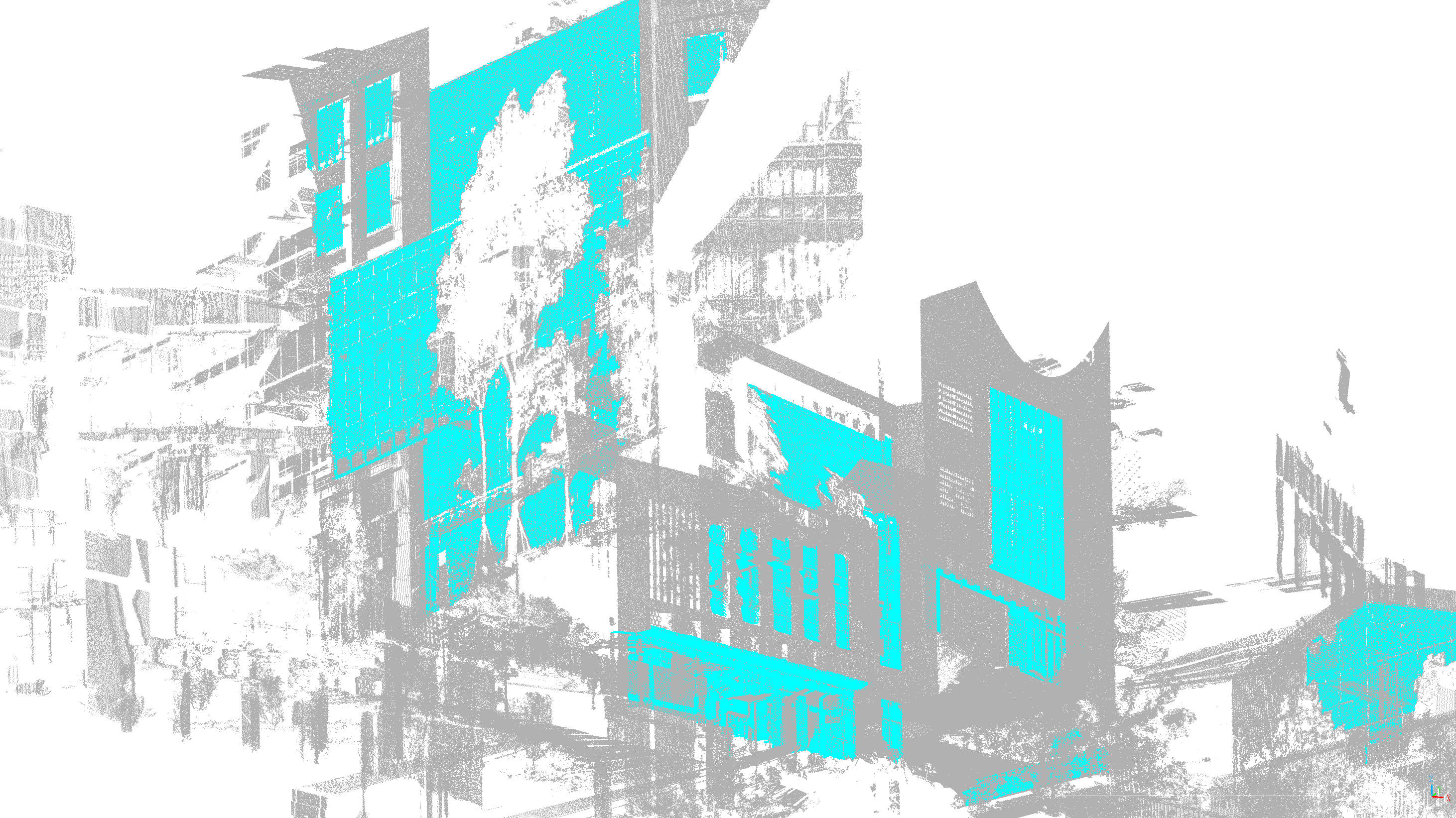}}
     \vspace{3pt}
     \centerline{\includegraphics[width=\textwidth]{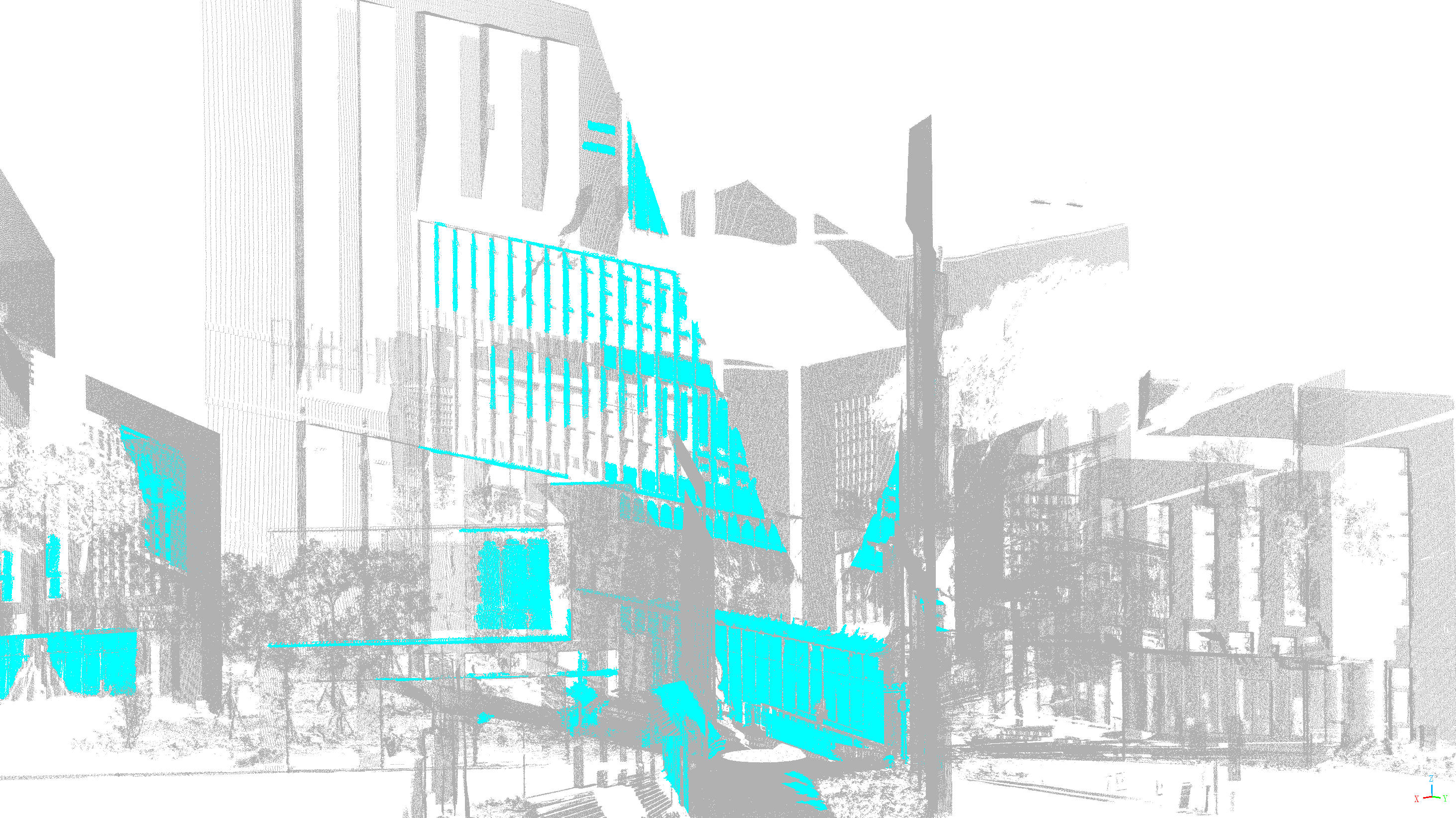}}
     \vspace{3pt}
     \centerline{\includegraphics[width=\textwidth]{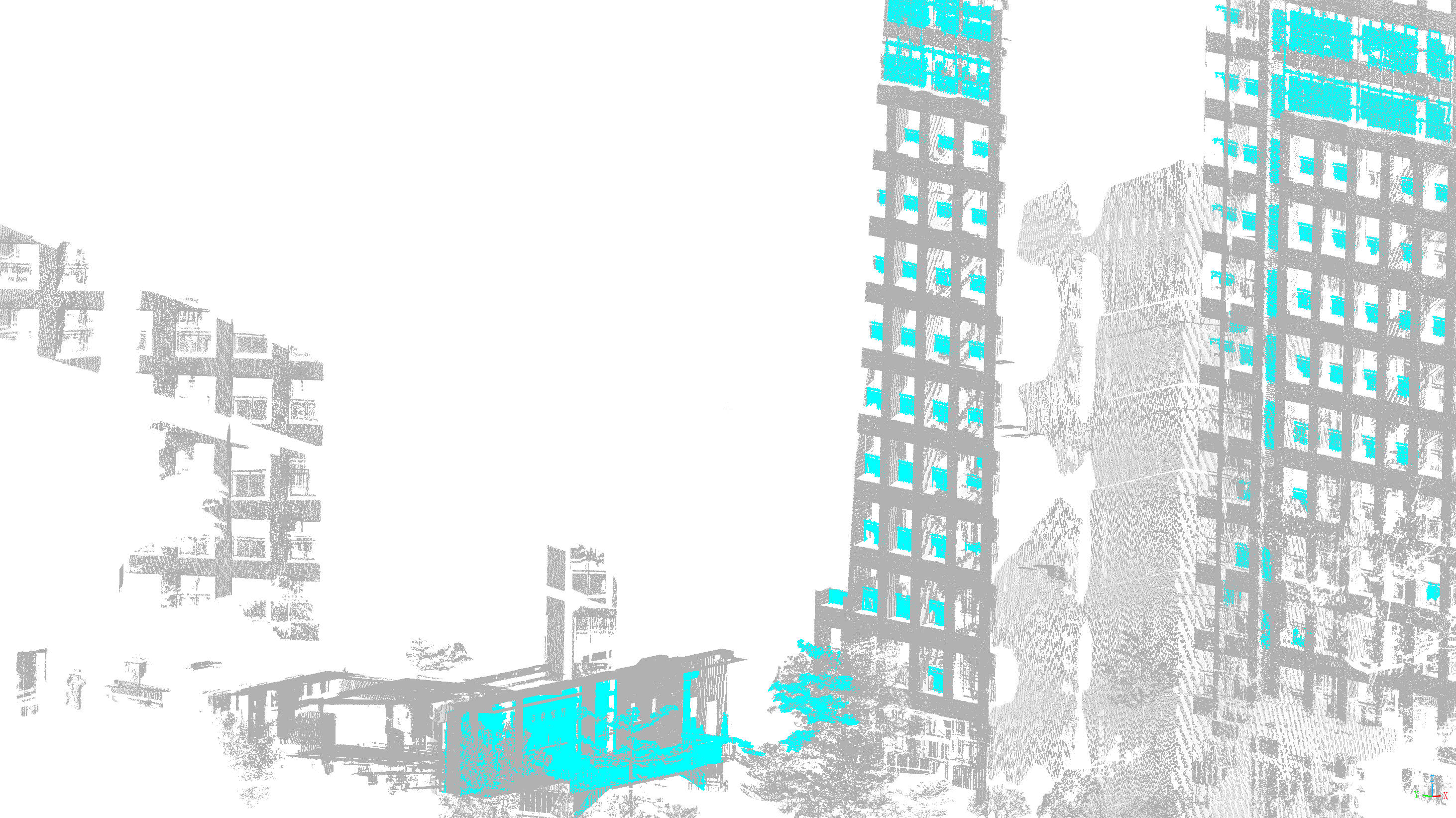}}
     \vspace{3pt}
     \centerline{\includegraphics[width=\textwidth]{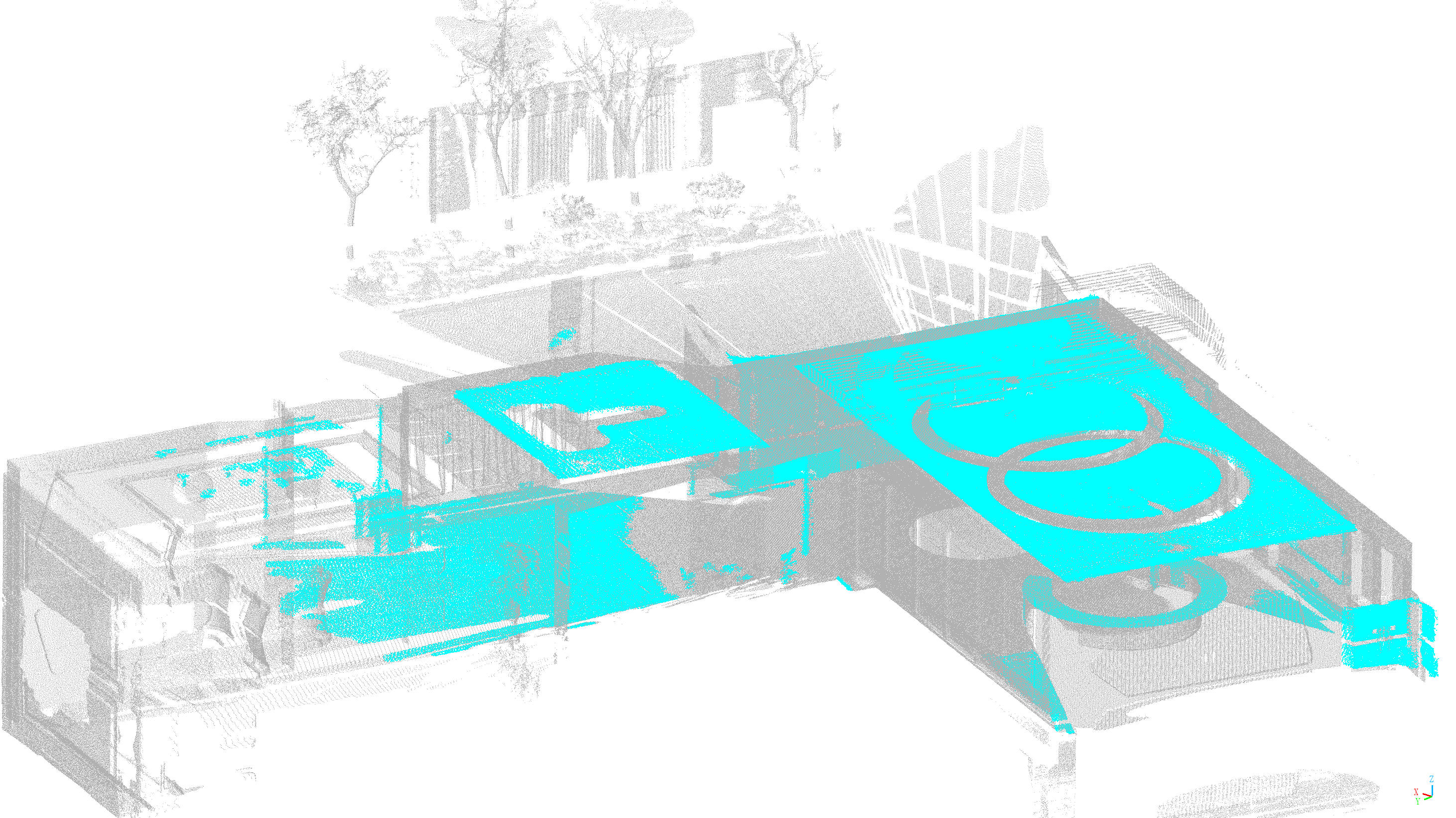}}
     \vspace{3pt}
     \centerline{(c)}
 \end{minipage}
     \begin{minipage}{0.24\linewidth}
     \vspace{3pt}
     \centerline{\includegraphics[width=\textwidth]{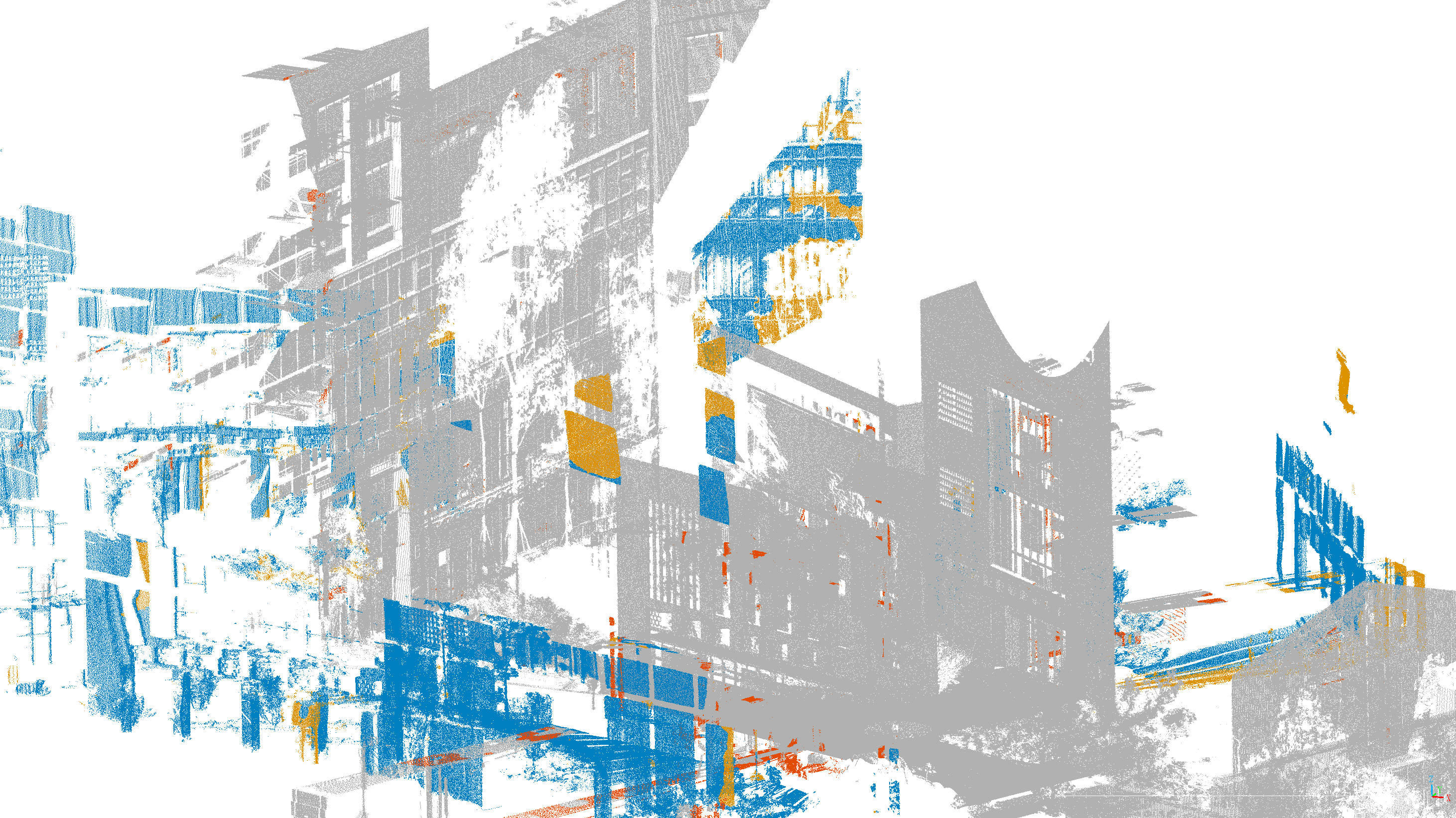}}
     \vspace{3pt}
     \centerline{\includegraphics[width=\textwidth]{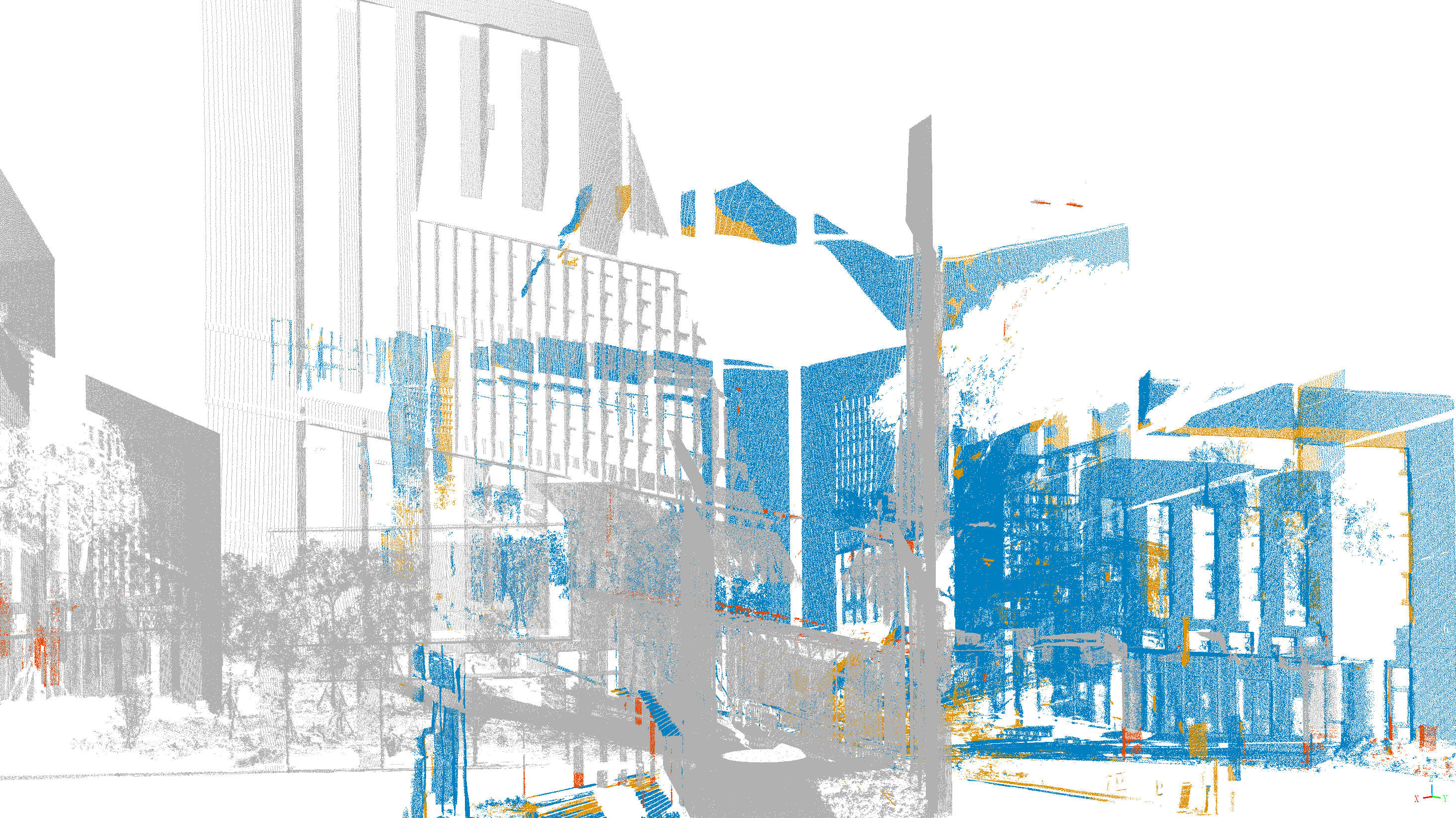}}
     \vspace{3pt}
     \centerline{\includegraphics[width=\textwidth]{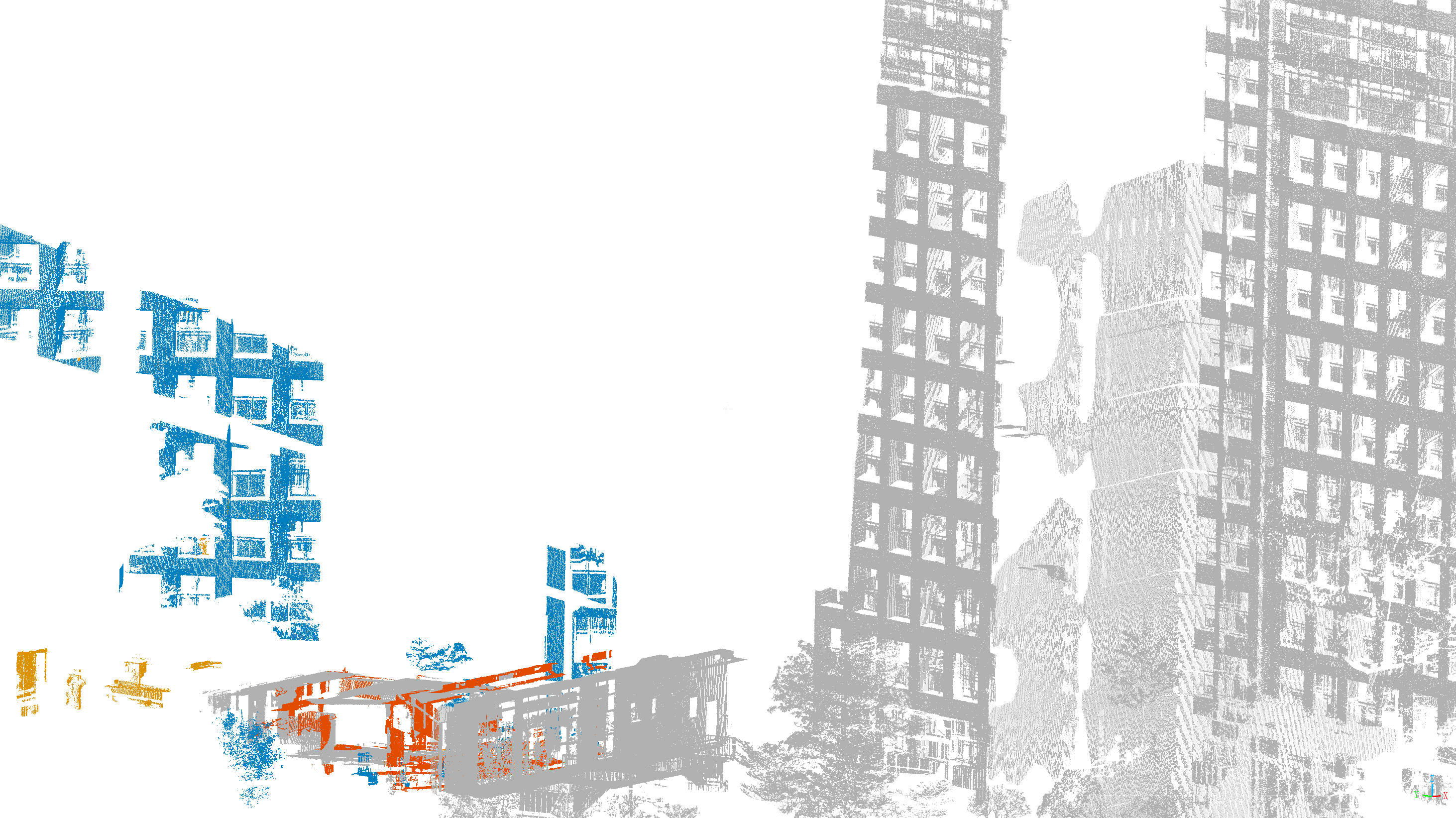}}
     \vspace{3pt}
     \centerline{\includegraphics[width=\textwidth]{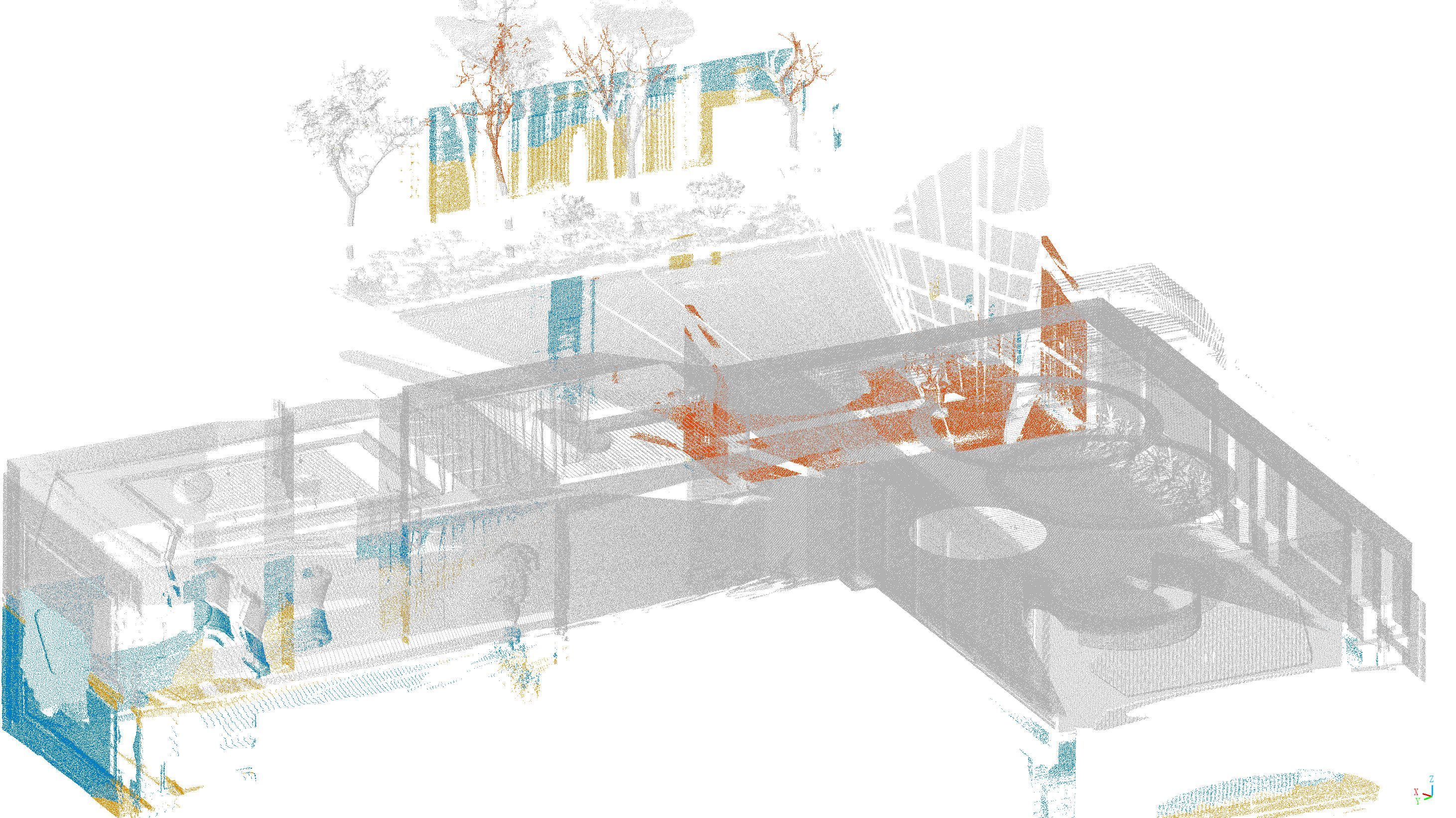}}
     \vspace{3pt}
     \centerline{(d)}
 \end{minipage}
   \begin{minipage}{0.98\linewidth}
    \vspace{3pt}
    \centerline{\includegraphics[width=\textwidth]{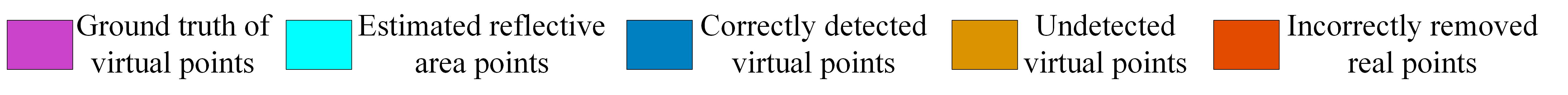}}
  \end{minipage}
	\caption{Results of the proposed method. (a) Input point cloud. (b) Ground truth of the virtual points. (c) Predicted multiple reflection planes. (d) Virtual point detection results. }
	\label{fig:result}
\end{figure*}
\begin{figure*}[b]   
\centering
 \begin{minipage}{0.24\linewidth}
     \vspace{3pt}  
     \centerline{\includegraphics[width=\textwidth]{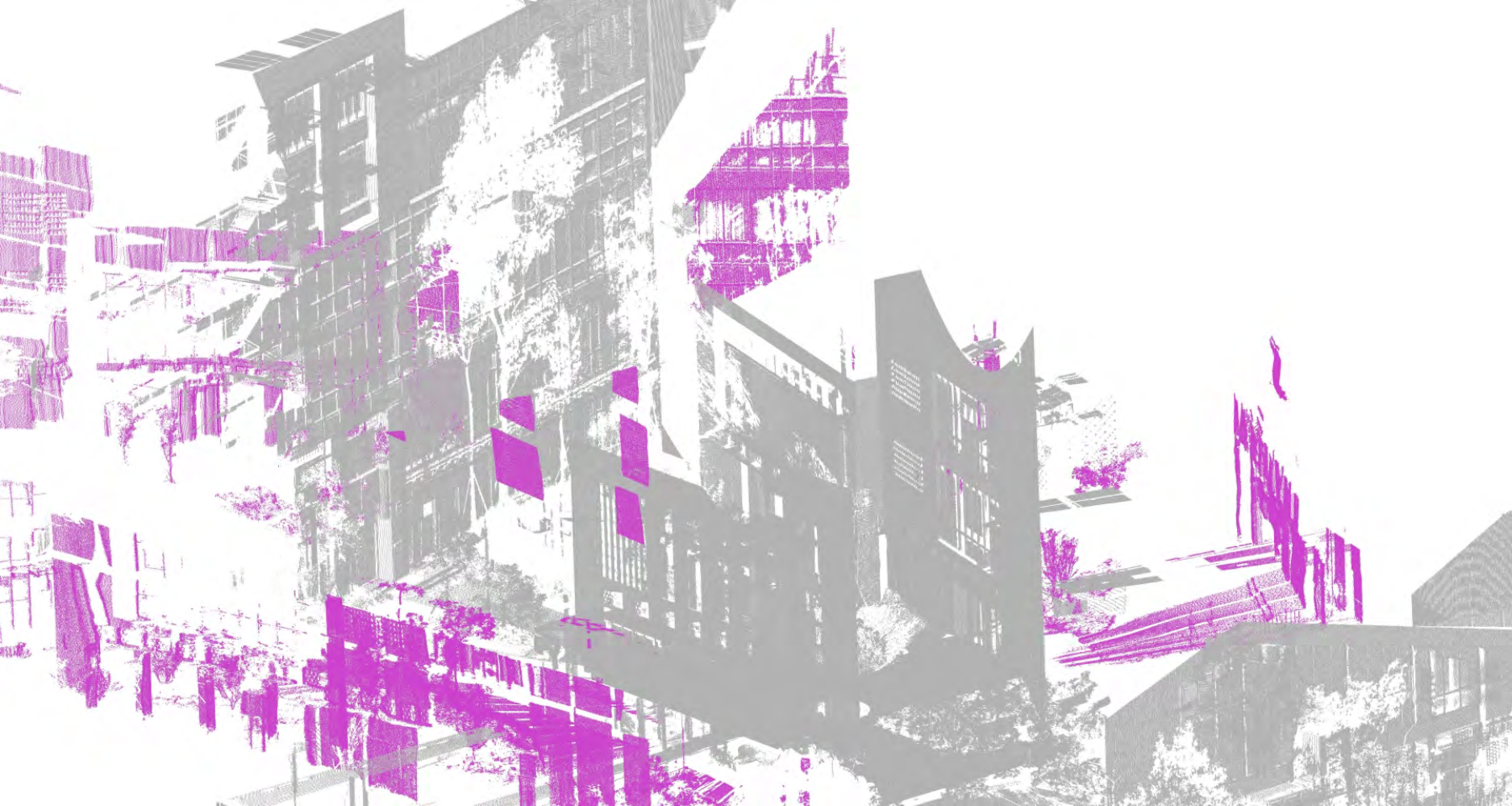}}
     \vspace{3pt}
     \centerline{\includegraphics[width=\textwidth]{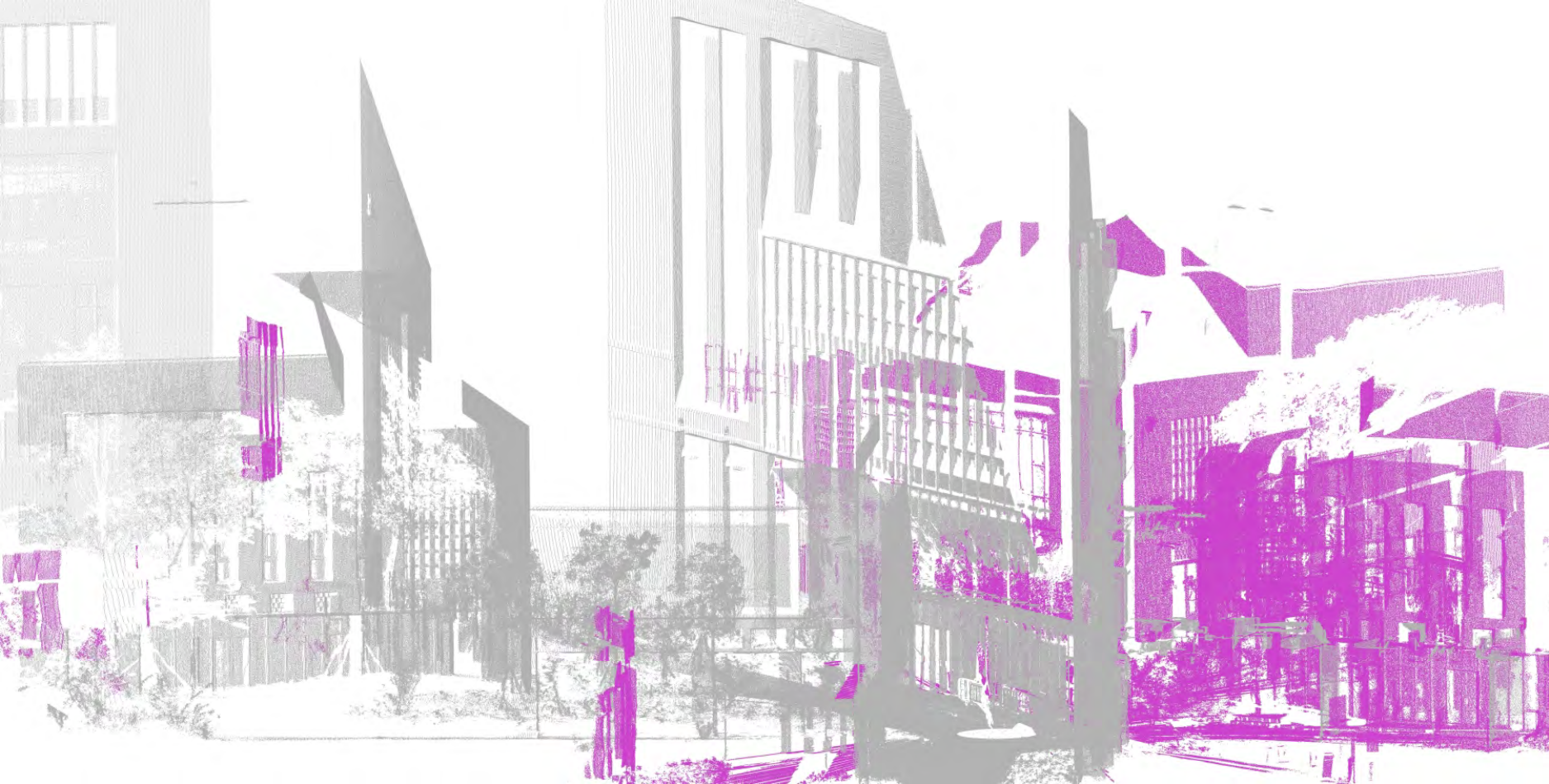}}
     \vspace{3pt}
     \centerline{\includegraphics[width=\textwidth]{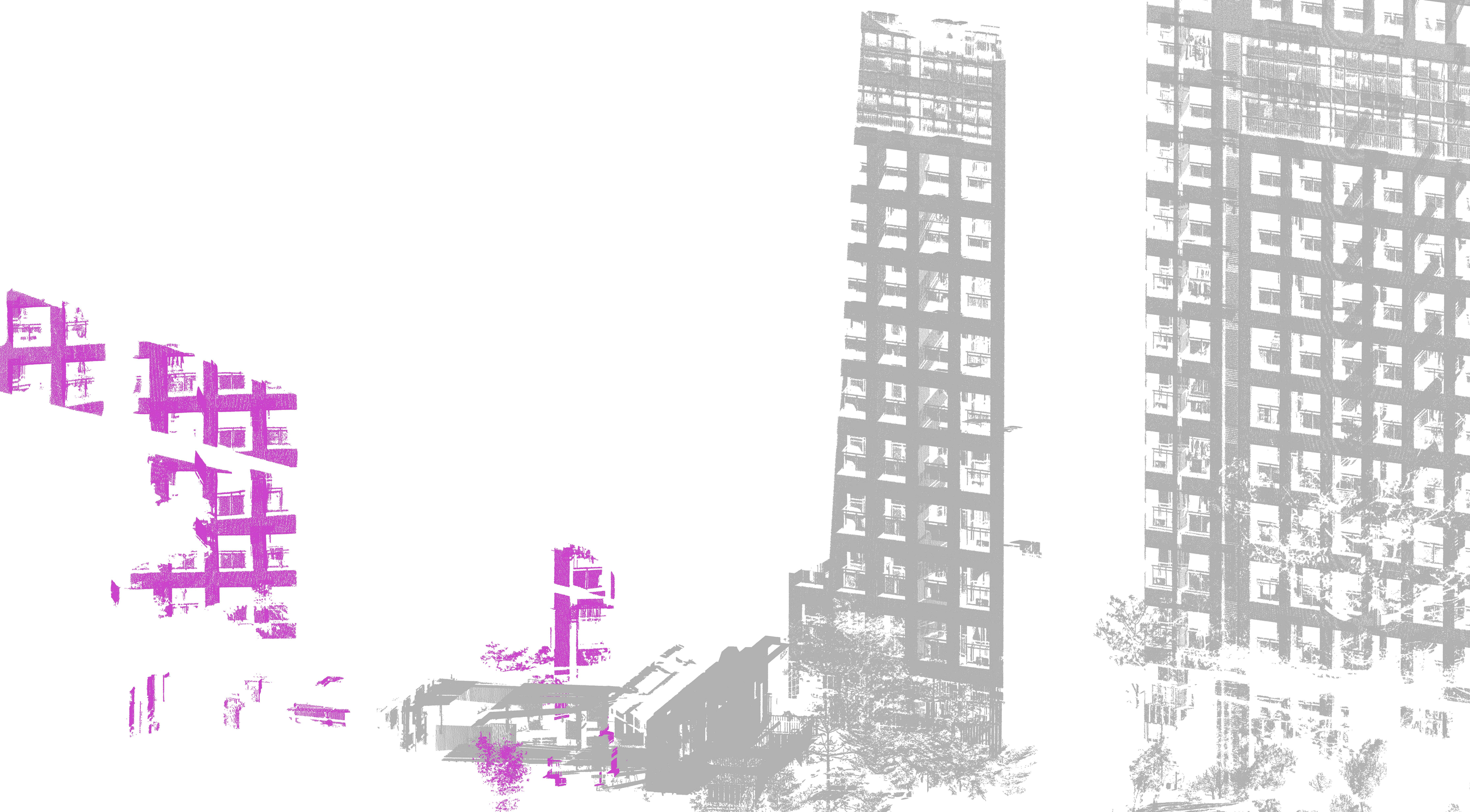}}
     \vspace{3pt}
     \centerline{\includegraphics[width=\textwidth]{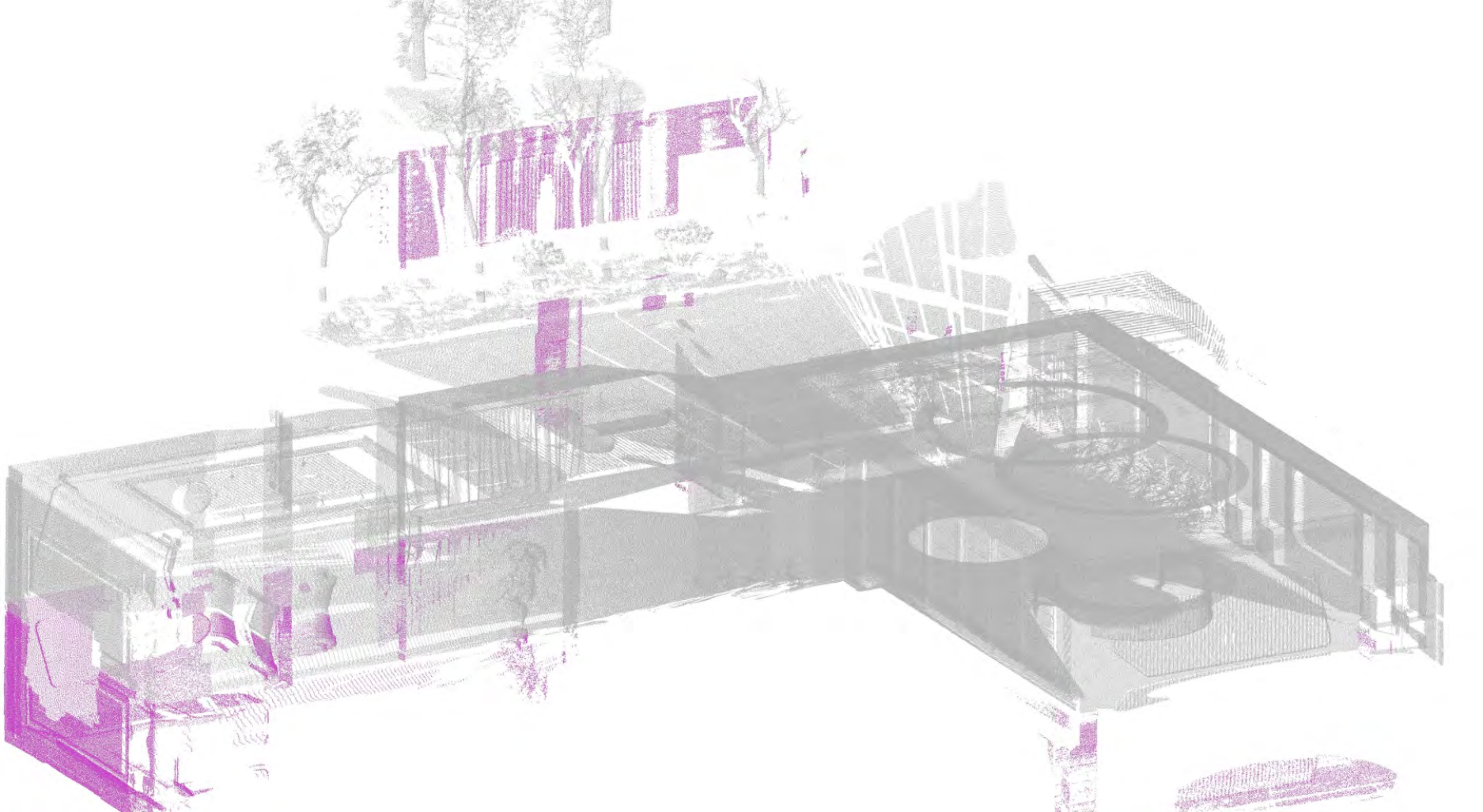}}
     \vspace{3pt}
     \centerline{(a) Ground truth}
 \end{minipage}
    \begin{minipage}{0.24\linewidth}
     \vspace{3pt}
     \centerline{\includegraphics[width=\textwidth]{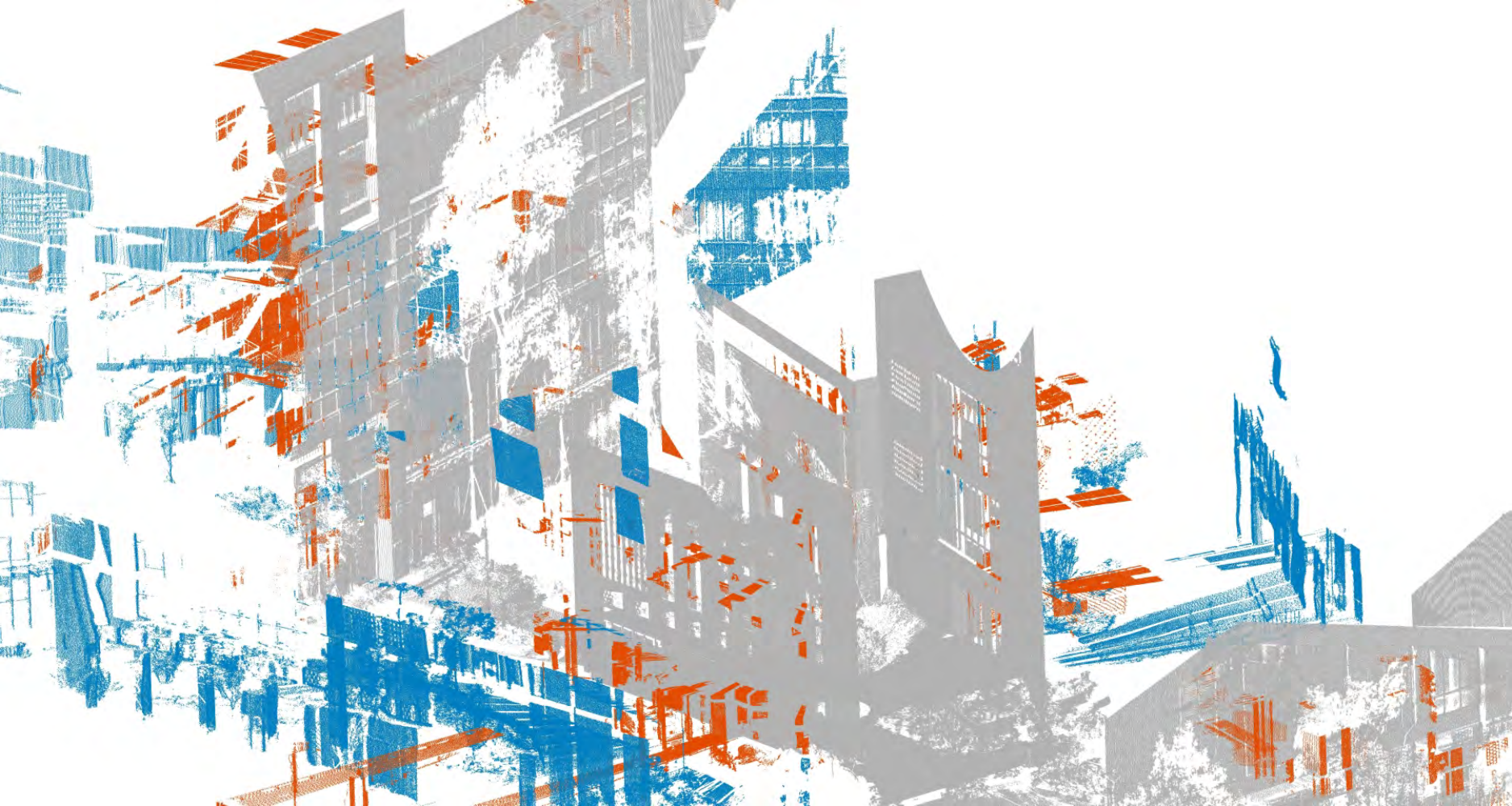}}
     \vspace{3pt}
     \centerline{\includegraphics[width=\textwidth]{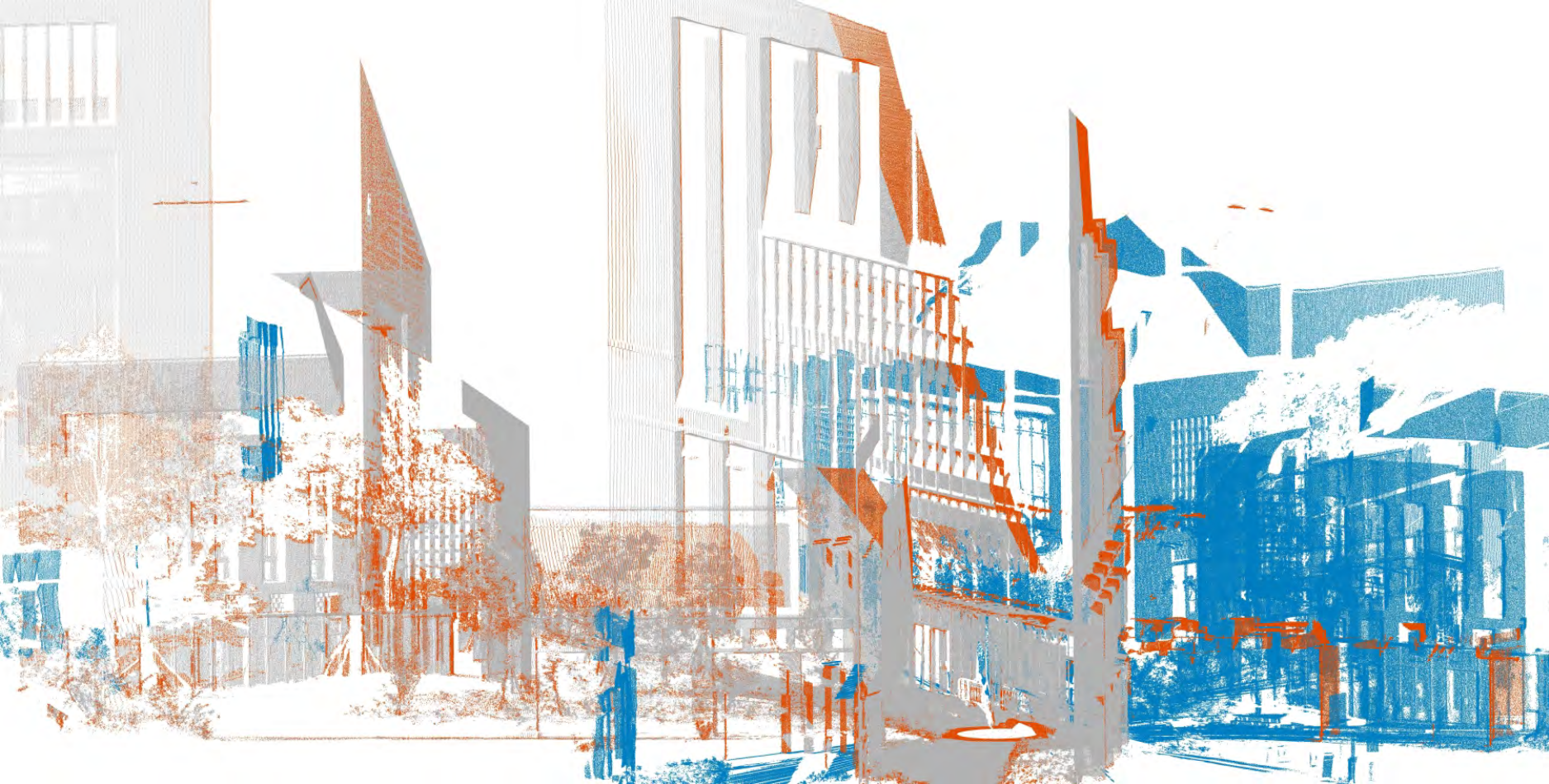}}
     \vspace{3pt}
     \centerline{\includegraphics[width=\textwidth]{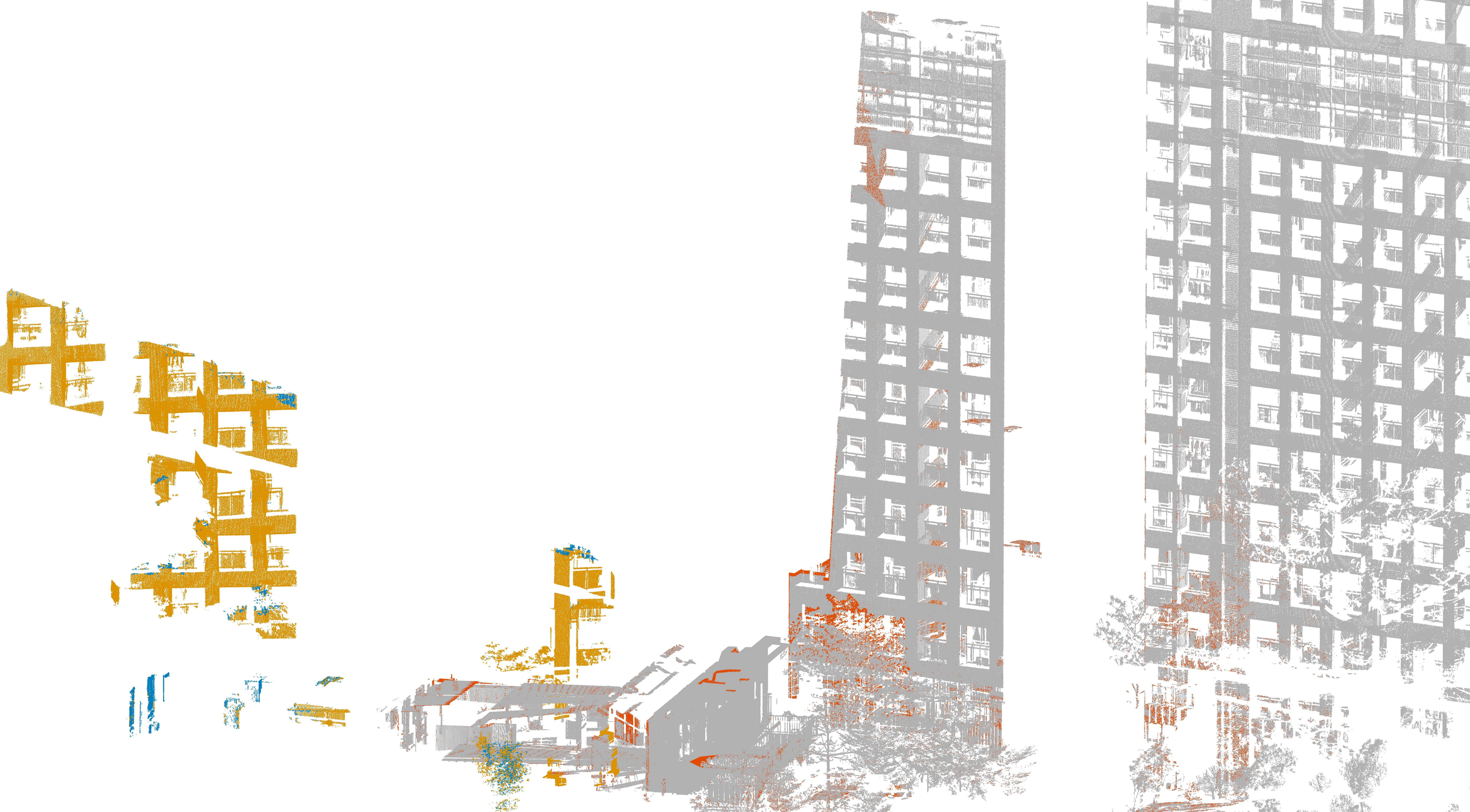}}
     \vspace{3pt}
     \centerline{\includegraphics[width=\textwidth]{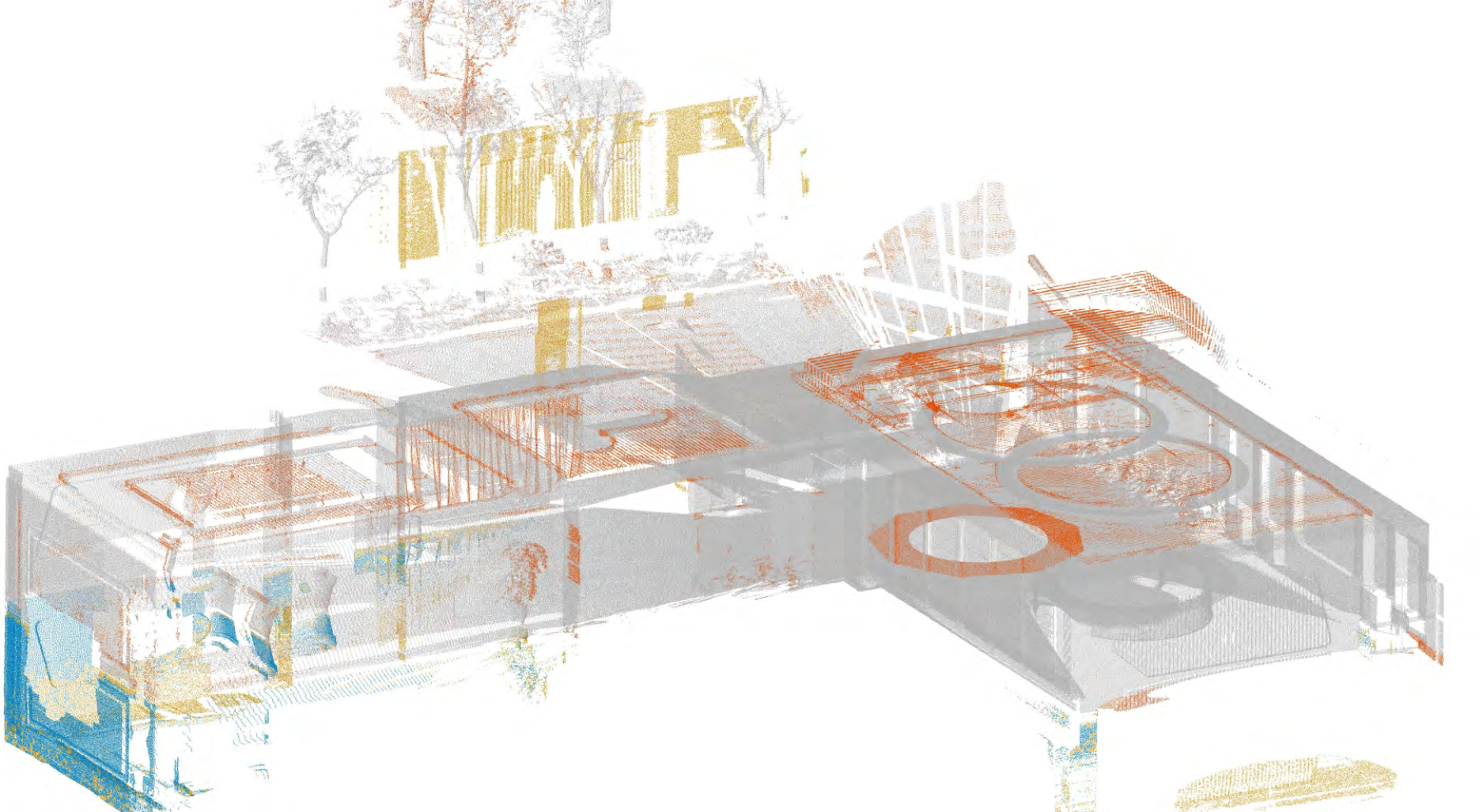}}
     \vspace{3pt}
     \centerline{(b) FARO SCENE}
 \end{minipage}
    \begin{minipage}{0.24\linewidth}
     \vspace{3pt}
     \centerline{\includegraphics[width=\textwidth]{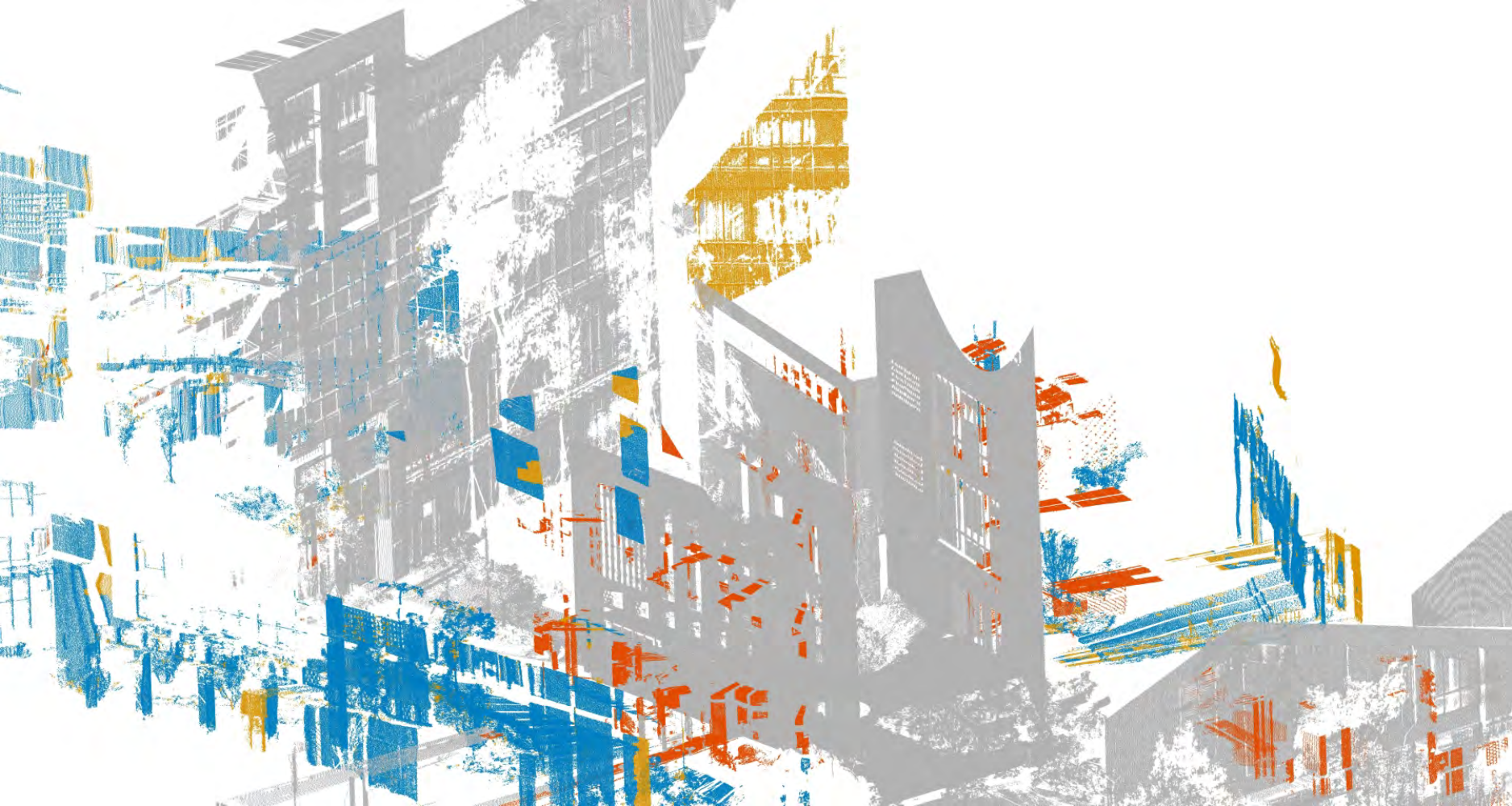}}
     \vspace{3pt}
     \centerline{\includegraphics[width=\textwidth]{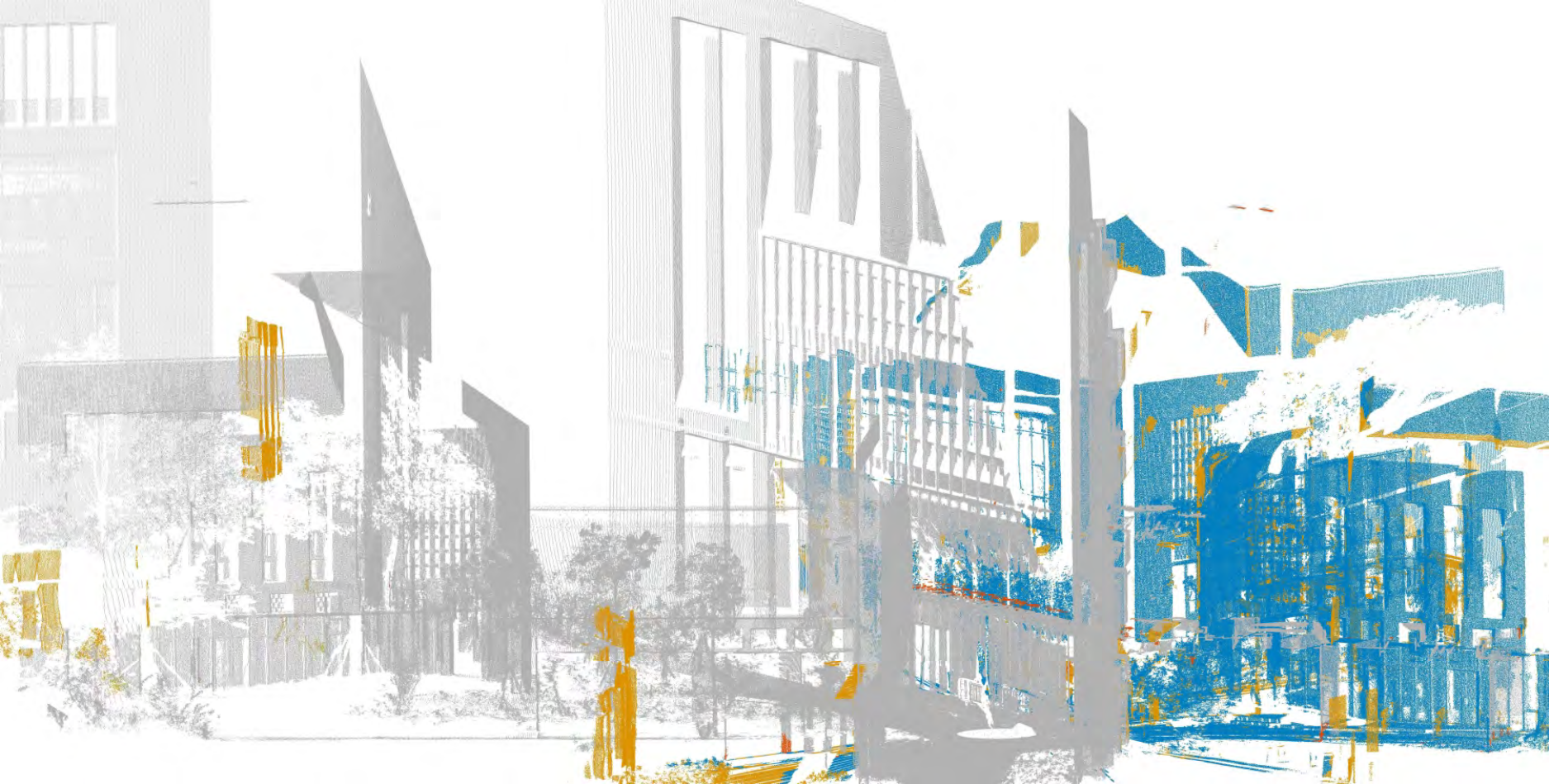}}
     \vspace{3pt}
     \centerline{\includegraphics[width=\textwidth]{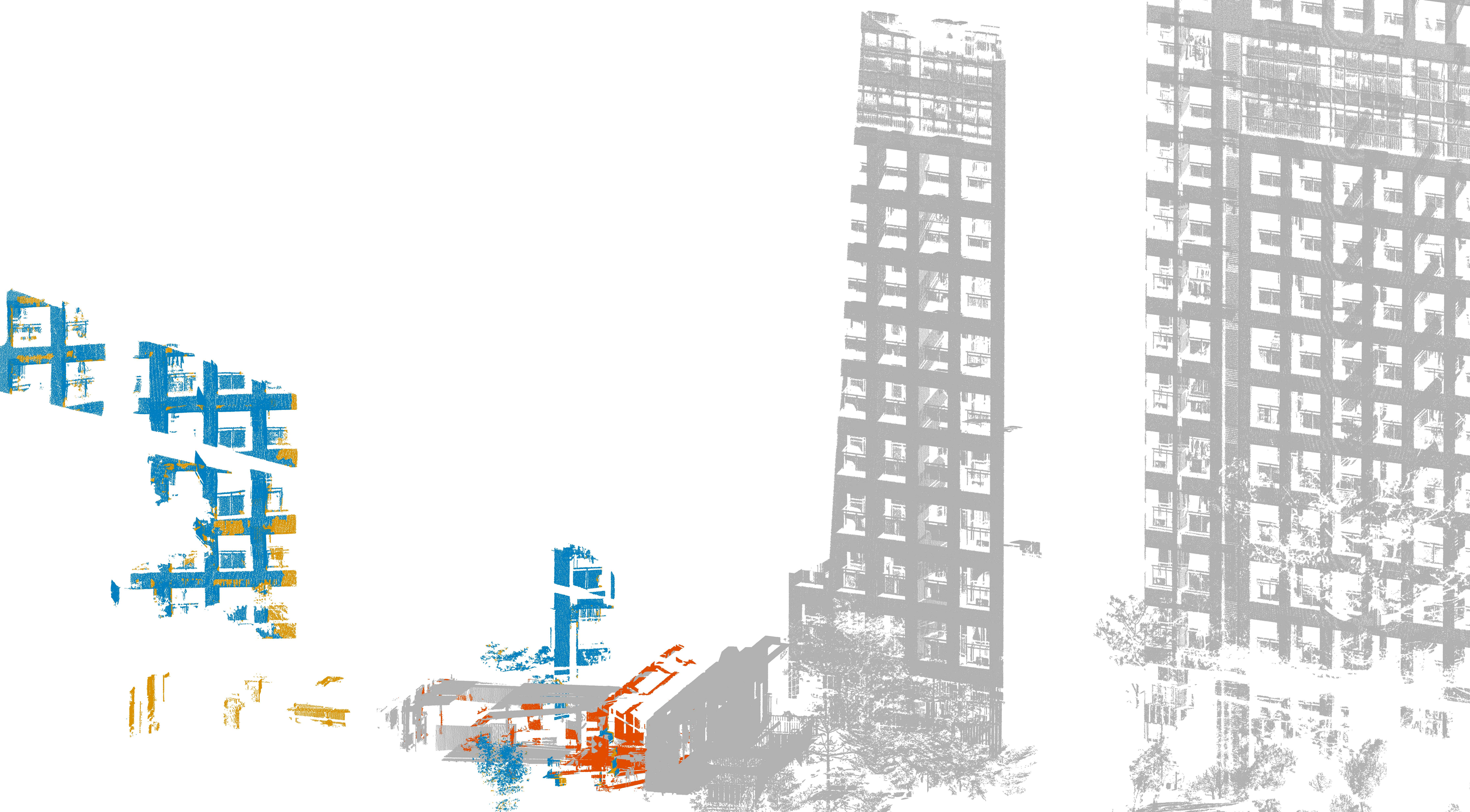}}
     \vspace{3pt}
     \centerline{\includegraphics[width=\textwidth]{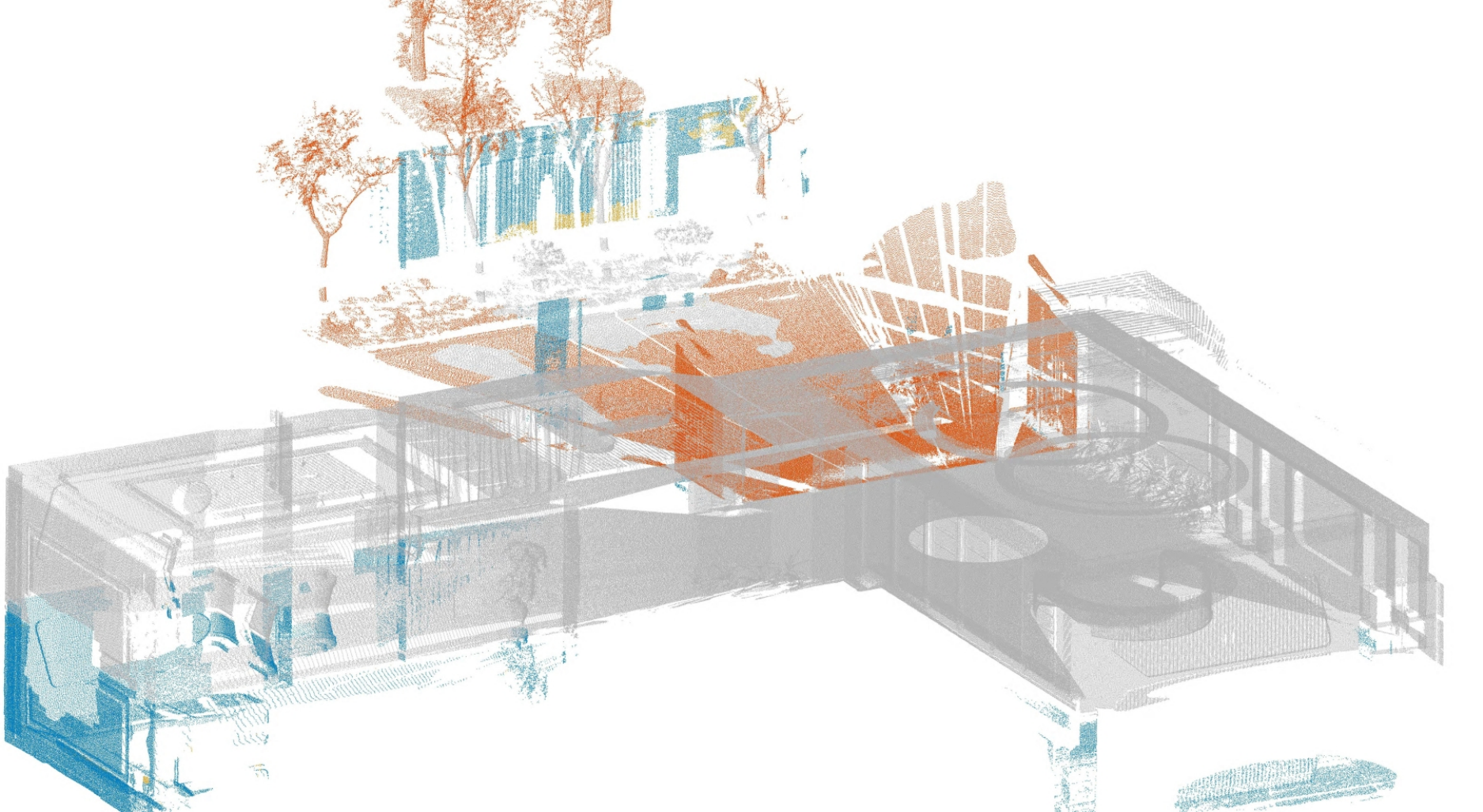}}
     \vspace{3pt}
     \centerline{(c) \cite{RN64}}
 \end{minipage}
     \begin{minipage}{0.24\linewidth}
     \vspace{3pt}
     \centerline{\includegraphics[width=\textwidth]{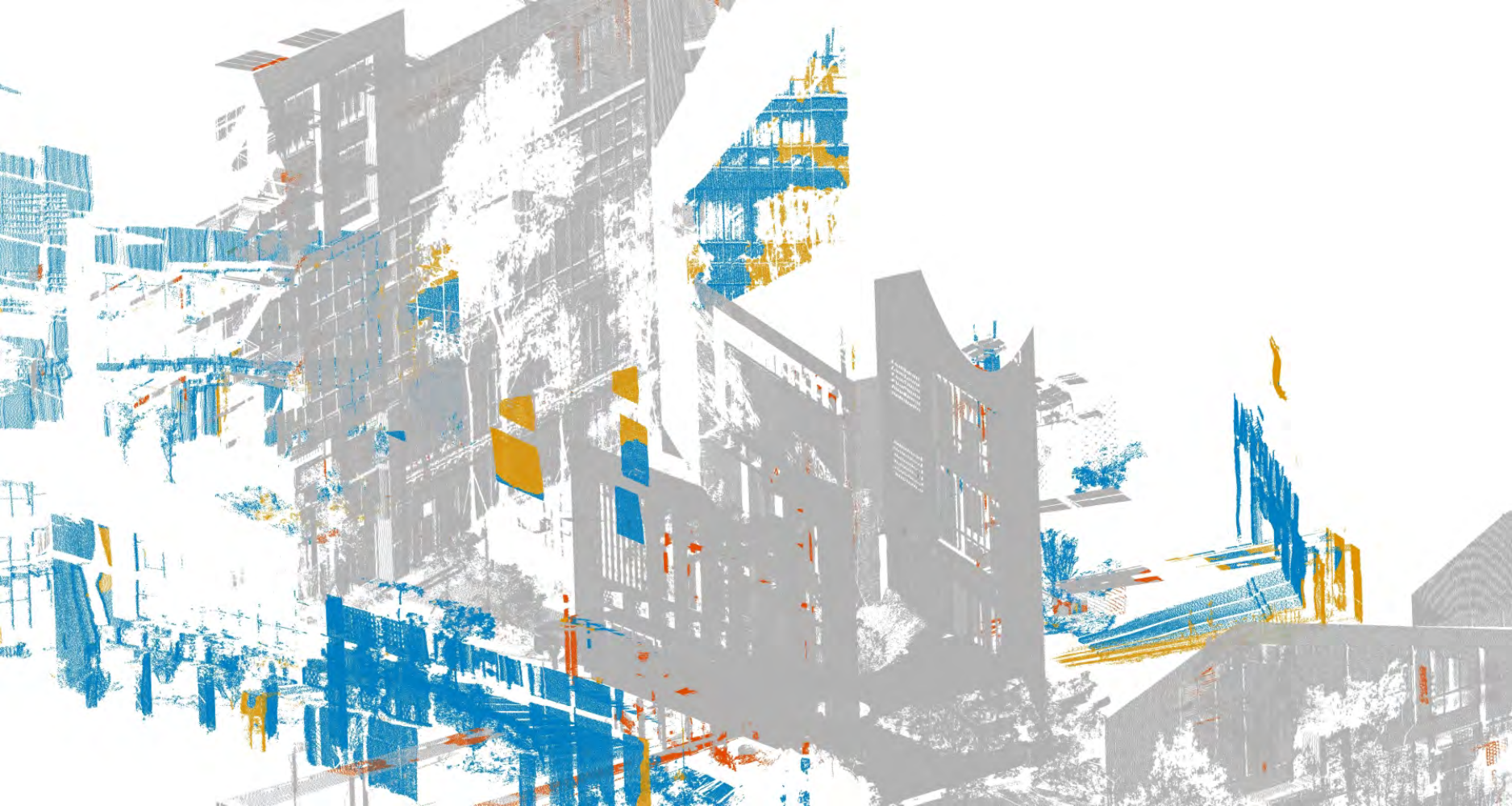}}
     \vspace{3pt}
     \centerline{\includegraphics[width=\textwidth]{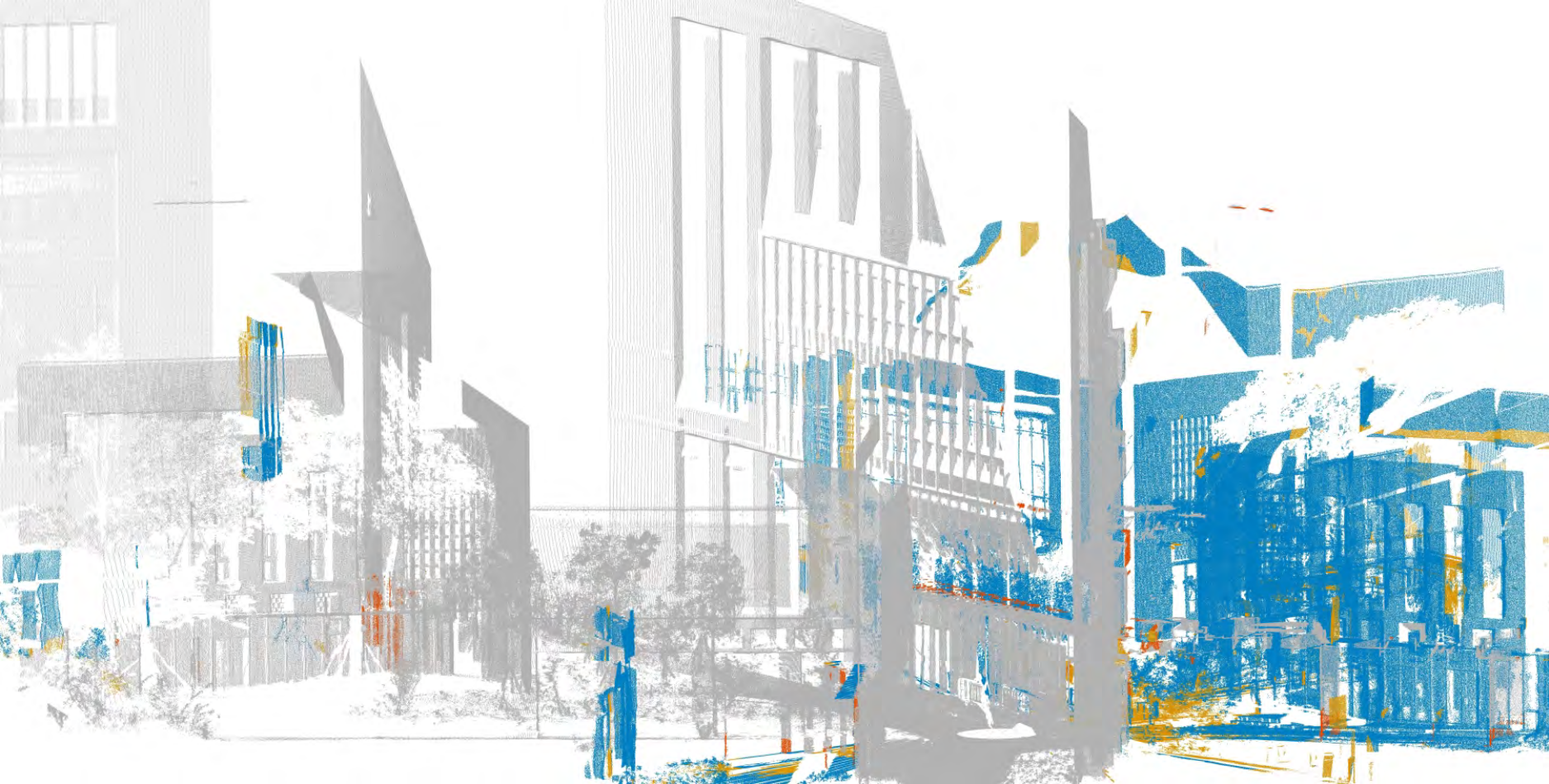}}
     \vspace{3pt}
     \centerline{\includegraphics[width=\textwidth]{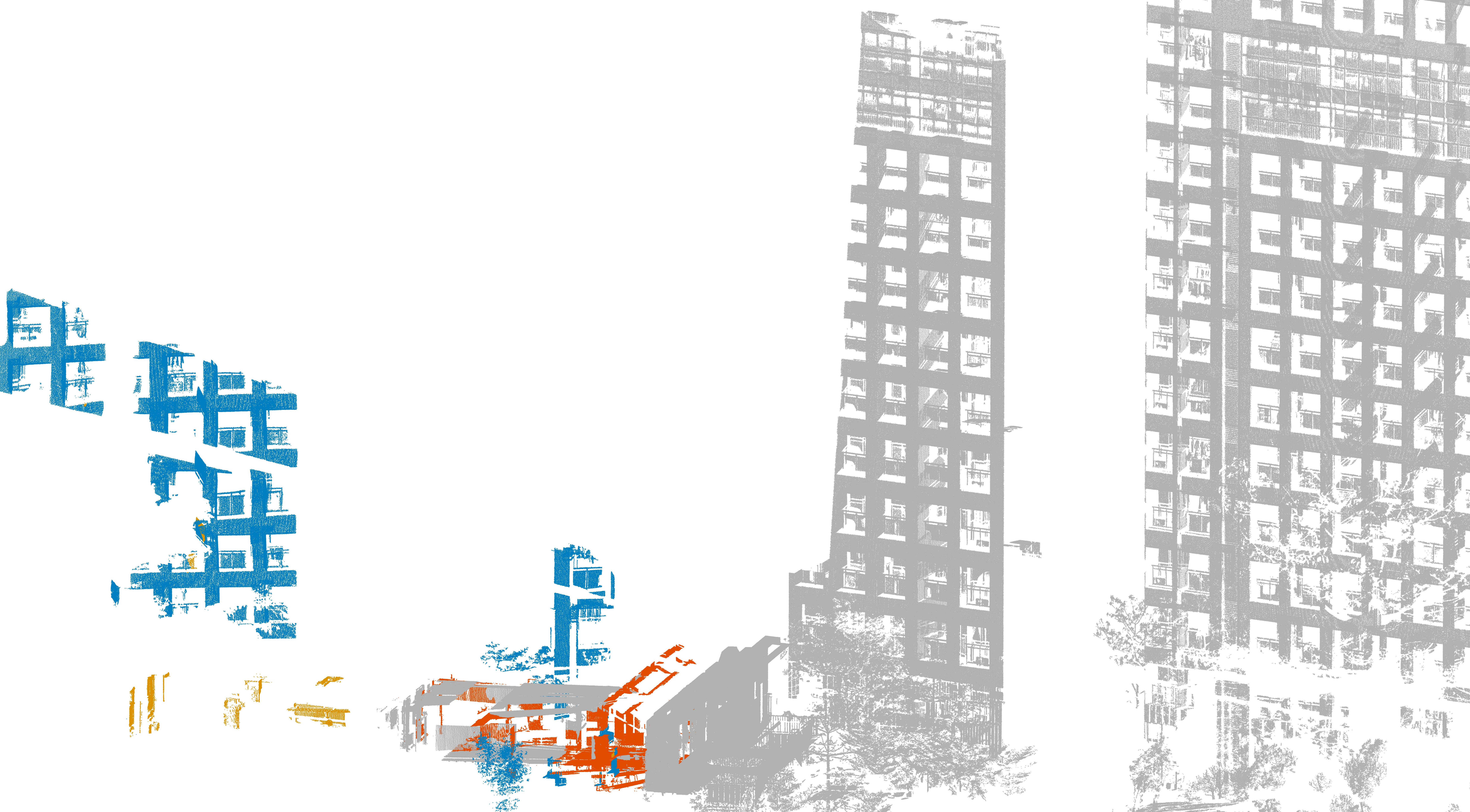}}
     \vspace{3pt}
     \centerline{\includegraphics[width=\textwidth]{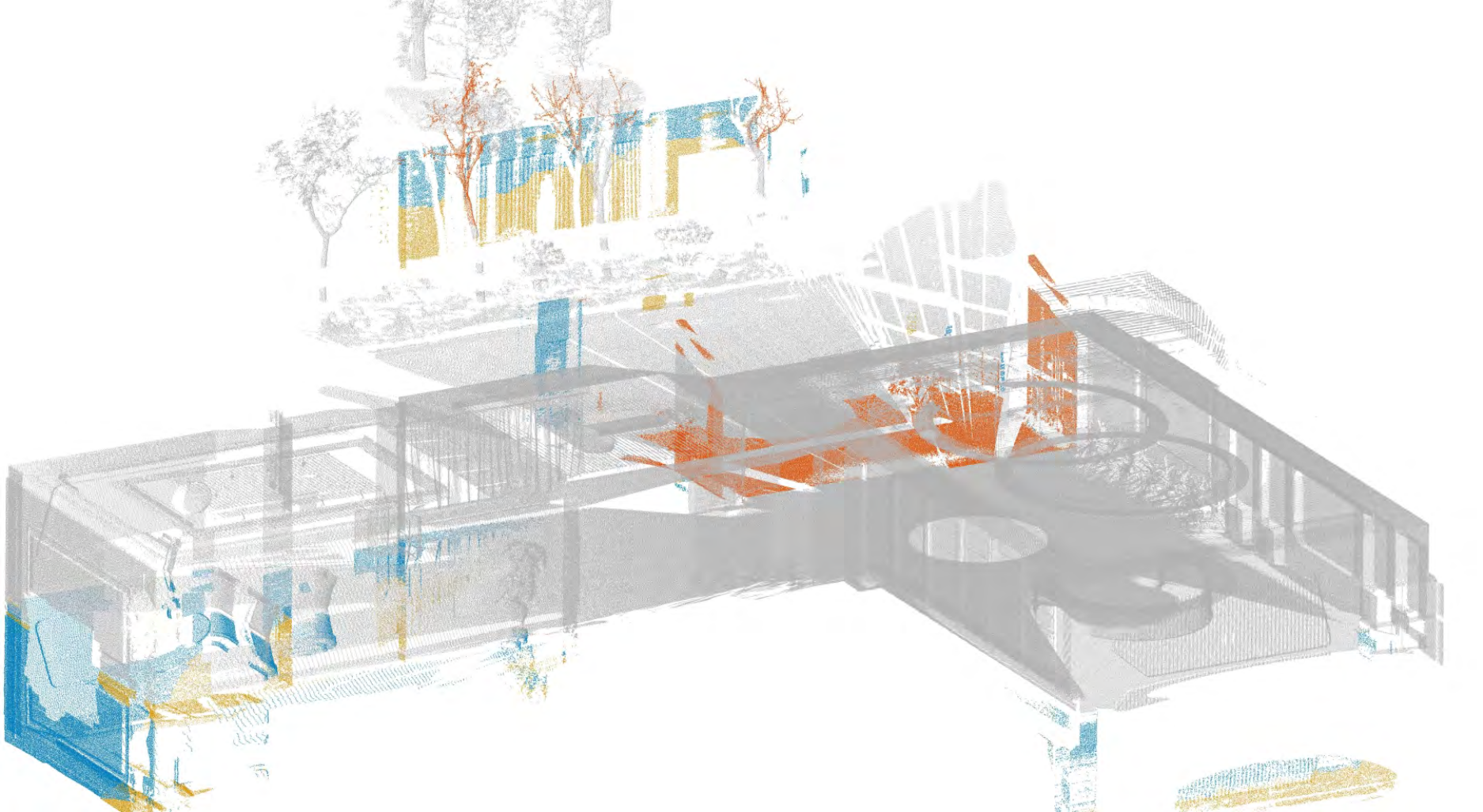}}
     \vspace{3pt}
     \centerline{(d) Proposed}
 \end{minipage}
 \begin{minipage}{0.98\linewidth}
    \vspace{3pt}
    \centerline{\includegraphics[width=\textwidth]{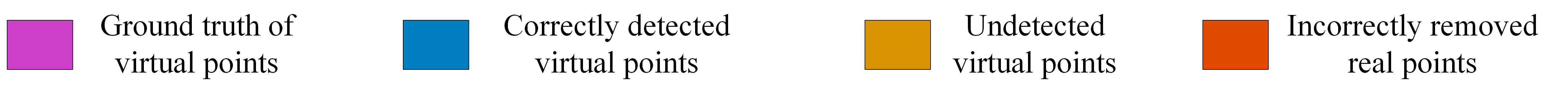}}
\end{minipage}
	\caption{Comparison of the virtual point removal performance from different methods.}
	\label{fig:com}
\end{figure*}
Fig.\ref{fig:result} displays the effectiveness of the proposed algorithm on four extensive point cloud models. Fig.\ref{fig:result}a depicts the original point cloud. The first example is an outdoor scene featuring buildings with large glass curtain walls, trees, and gardens. The last two examples are primarily indoor scenes that include reflective surfaces such as glass and tiled floors. The ground truth for virtual points is presented in Fig.\ref{fig:result}b, which has been manually adjusted. Reflective surfaces identified by the proposed method are shown in Fig.\ref{fig:result}c. These models contain intricate reflection structures. Using the proposed reflective surface detection algorithm, most reflective surfaces in the scenes are accurately identified. However, in certain instances, the estimated reflective surfaces may be incomplete, with some missing parts along the edges. This occurs because the proposed algorithm might exclude some planar clusters located at the edges during the plane fitting process. However, this does not affect the subsequent virtual point detection, as our
method relies on the precision of the estimated normal vectors rather than the completeness of the estimated plane. In indoor environments, besides glass, other reflective surfaces such as smooth tiles can cause reflection noise. In scenes 3 and 4, the proposed algorithms have successfully detected reflective planes made of non-glass materials, in addition to the common glass planes. Fig.\ref{fig:result}d displays both the detected virtual points and those that were not removed. It also shows the real points that were incorrectly removed as virtual points by the proposed method. Removing most of the virtual points while minimizing the removal of real points is a challenging task. However, the proposed algorithm effectively removes most of the reflected noise while maintaining a good performance in preserving real points. Scene 1 is a typical and complex model featuring four large glass curtain walls made up of numerous small glass planes. These complex glass planes generate a significant amount of reflected noise. The presence of many similar trees and building structures in front of the glass, along with some real points behind the glass, adds to the challenge. Despite this, the proposed method performs well in Scene 1, successfully removing most of the virtual points.

\subsection{Comparison of Noise Detection}
To assess the effectiveness of the proposed algorithm, we evaluate it against the denoising method offered by FARO SCENE \citep{FARO} and the method proposed in \cite{RN64}. We adhere to the parameter settings used in FARO SCENE as described in \cite{RN60}.

\subsubsection{Qualitative Comparison}

In previous work \citep{RN63,RN64}, fitting a glass plane requires the corresponding real and virtual points. In the case of missing corresponding real points, the glass plane detection will not succeed. Nevertheless, our newly proposed reflective plane detection method leverages data like echo intensity to identify reflection regions directly, eliminating the need for corresponding real and virtual points. 

Fig.\ref{fig:com} illustrates the performance of virtual point removal for the aforementioned methods across four scenes. It is evident that in the evaluated scenes, the FARO SCENE results contain numerous undetected virtual points. In comparison, the approach by \cite{RN64} eliminates most virtual points, although some small fragments of virtual noise remain undetected. This could be attributed to the absence of mirror symmetry properties in the FPFH feature descriptor they utilized. The proposed method demonstrates superior performance among the three methods. This is due to two factors: firstly, our algorithm precisely estimates reflective planes; secondly, we have enhanced the feature descriptor and distance metric for the real reflection scheme. With these improvements, the performance of the proposed method has been further enhanced.
To demonstrate the information loss regarding the original geometric structure of the point cloud by the three methods, we also assess the preservation of real points by these methods. The erroneously removed real points are highlighted in red-orange in Fig.\ref{fig:com}. It is evident that FARO SCENE eliminates numerous isolated real points as virtual points, leading to a significant loss of structure-forming geometric information of the original point cloud and substantially degrading its quality. While the approach by \cite{RN64} alleviates this issue to some degree, there is still a loss of small segments of the real point cloud, such as trees and floors. Specifically, in the fourth scene with highly similar floor interference, \cite{RN64} entirely removes the floor in front of the glass. Overall, our method efficiently removes virtual points while significantly reducing the loss of geometric information in the point cloud.
\begin{table*}[htbp]
\centering
\caption{Comparison of quantitative performance. The last row presents the average values, with the best performing results highlighted in \textbf{bold}. ODR, IDR, FPR, and FNR are shown as percentages (\%).}\label{table:odr}
\resizebox{\textwidth}{!}{
\begin{tabular}{@{}ccccccccccccccccc@{}}
\toprule
\multirow{2}{*}{\textbf{}} & \multicolumn{3}{c}{\textbf{ODR}(\%)}                                                                                                                                                                                                              & \multicolumn{3}{c}{\textbf{IDR}(\%)}                                                                                                                                                                                                              & \multicolumn{3}{c}{\textbf{FPR}(\%)}                                                                                                                                                                                                              & \multicolumn{3}{c}{\textbf{FNR}(\%)}                                                                                                                                                                                                              \\ \cmidrule(r){2-4} \cmidrule(r){5-7} \cmidrule(r){8-10} \cmidrule(r){11-13} 
                           & \begin{tabular}[c]{@{}c@{}}FARO SCENE\end{tabular} & \begin{tabular}[c]{@{}c@{}}\cite{RN64}\end{tabular} & \begin{tabular}[c]{@{}c@{}}Proposed\end{tabular} & \begin{tabular}[c]{@{}c@{}}FARO SCENE\end{tabular} & \begin{tabular}[c]{@{}c@{}}\cite{RN64}\end{tabular} & \begin{tabular}[c]{@{}c@{}}Proposed\end{tabular}& \begin{tabular}[c]{@{}c@{}}FARO SCENE\end{tabular} & \begin{tabular}[c]{@{}c@{}}\cite{RN64}\end{tabular} & \begin{tabular}[c]{@{}c@{}}Proposed\end{tabular}& \begin{tabular}[c]{@{}c@{}}FARO SCENE\end{tabular} & \begin{tabular}[c]{@{}c@{}}\cite{RN64}\end{tabular} & \begin{tabular}[c]{@{}c@{}}Proposed\end{tabular} \\ \midrule
Scan 01                          & 99.46                                               & 79.10                                                      & 87.06                                                    & 57.88                                                       & 93.58                                               & 98.44                                                      & 42.12                                                    & 6.42                                                       & 1.56                                               & 0.54                                                      & 20.90                                                    & 12.94                                                       \\
Scan 02                          & 99.99                                               & 52.40                                                      & 88.46                                                    & 67.24                                                       & 96.37                                               & 98.23                                                      & 32.76                                                    & 3.63                                                       & 1.77                                               & 0.01                                                      & 47.60                                                    & 11.54                                                       \\
Scan 03                          & 99.98                                               & 6.78                                                      & 83.02                                                    & 59.56                                                       & 99.18                                               & 99.61                                                      & 40.44                                                    & 0.28                                                       & 0.39                                               & 0.15                                                      & 93.22                                                    & 16.98                                                       \\
Scan 04                          & 99.89                                               & 75.12                                                      & 87.30                                                    & 75.97                                                       & 99.28                                               & 98.68                                                      & 24.03                                                    & 0.72                                                       & 1.32                                               & 0.11                                                      & 24.88                                                    & 12.70                                                       \\
Scan 05                          & 99.65                                               & 50.24                                                      & 88.52                                                    & 72.36                                                       & 97.49                                               & 97.48                                                      & 27.64                                                    & 2.51                                                       & 2.52                                               & 0.35                                                      & 49.76                                                    & 11.48                                                       \\
Scan 06                          & 99.73                                               & 95.64                                                      & 84.20                                                    & 71.55                                                       & 87.91                                               & 98.25                                                      & 28.45                                                    & 12.09                                                       & 1.75                                               & 0.27                                                      & 4.36                                                    & 15.80                                                       \\ 
Scan 07                          & 14.17                                               & 68.47                                                      & 59.15                                                    & 85.16                                                       & 93.31                                               & 92.63                                                      & 14.84                                                    & 6.69                                                       & 7.37                                               & 85.83                                                      & 31.53                                                    & 40.85                                                       \\
Scan 08 &9.77 &77.99 &91.34 &85.94 &97.49 &96.70 &14.06 &2.51 &3.31 &90.23 &22.01 &8.66\\
Scan 09 &3.56 &99.47 &85.44 &90.47 &58.84 &86.76 &9.53 &41.16 &13.24 &96.44 &0.53 &14.56\\
Scan 10 &3.73 &47.81 &41.83 &84.12 &74.95 &84.31 &15.88 &25.05 &15.69 &96.27 &52.19 &58.17\\
Scan 11 &47.45 &97.43 &73.20 &86.68 &78.09 &94.56 &13.32 &21.91 &5.44 &52.55 &2.57 &26.80\\
Scan 12 &36.03 &93.12 &84.17 &91.06 &89.88 &85.62 &8.94 &10.12 &14.38 &63.97 &6.88 &15.83\\
Average &59.45 &70.30 &\textbf{79.47} &77.33 &88.86 &\textbf{94.27} &22.67 &11.09 &\textbf{5.73} &40.56 &29.70 &\textbf{20.53}\\
\bottomrule
\end{tabular}
}
\end{table*}

\begin{table}[t]
\centering
\caption{Comparison of accuracy performance. The last row represents the average accuracy, with the highest value highlighted in \textbf{bold}. Accuracy percentages are provided as \% values.}\label{table:accuracy}
\resizebox{\linewidth}{!}{
\begin{tabular}{@{}ccccc@{}}
\toprule
\multirow{2}{*}{} & \multicolumn{3}{c}{\textbf{Accuracy}(\%)}                              \\ \cmidrule(l){2-4} 
                  & \begin{tabular}[c]{@{}c@{}}FARO SCENE\end{tabular} & \begin{tabular}[c]{@{}c@{}}\cite{RN64}\end{tabular} & \begin{tabular}[c]{@{}c@{}}Proposed\end{tabular} \\ \midrule
Scan 01                 &65.55 & 90.94                    &96.37             \\
Scan 02                 &67.95 & 95.41                    &98.02             \\
Scan 03                 &60.25 & 97.60                    &99.33             \\
Scan 04                 &83.75 & 91.42                    &95.00             \\
Scan 05                 &77.66 & 88.32                    &95.74             \\
Scan 06                 &74.02 & 88.58                    &97.01             \\
 Scan 07&84.05 &92.92 &92.10 \\
 Scan 08&82.60 &96.63 &96.46 \\
 Scan 09&84.09 &61.82 &86.66 \\
 Scan 10&66.49 &69.00 &75.00 \\
 Scan 11&83.63 &79.59 &92.90 \\
 Scan 12&87.73 &90.08 &85.53 \\
Average           &76.48 & 86.86           &\textbf{92.51}                      \\ \bottomrule
\end{tabular}
}
\end{table}

\begin{table}[t]
\centering
\caption{SNR performance comparison. The last row shows the average SNR, with the highest SNR emphasized in \textbf{bold}. SNR values are given in dB.}\label{table:snr}
\resizebox{\linewidth}{!}{
\begin{tabular}{@{}ccccc@{}}
\toprule
                   &                                & \multicolumn{3}{c}{\textbf{SNR(dB)}}                                                                                                                                                                \\ \cmidrule(l){3-5} 
\multirow{-2}{*}{} & \multirow{-2}{*}{Original SNR} & \begin{tabular}[c]{@{}c@{}}FARO SCENE\end{tabular} & \begin{tabular}[c]{@{}c@{}}\cite{RN64}\end{tabular} & \begin{tabular}[c]{@{}c@{}}Proposed\end{tabular} \\ \midrule
Scan 01                  & 6.52& 3.74& 9.56& 13.52\\
Scan 02                  & 16.51& 4.85& 13.29& 16.93\\
Scan 03                  & 17.57& 3.93& 17.19& 21.64\\
Scan 04                  & 3.17& 6.18& 8.96& 11.28\\
Scan 05                  & 6.18& 5.57& 8.39& 12.77\\
Scan 06                  & 10.17& 5.46& 9.03& 14.85\\
Scan 07                  & 18.00& 7.90& 11.43& 10.96\\
Scan 08                  & 13.39& 7.40& 14.53& 14.32\\
Scan 09                  & 11.01& 7.65& 3.85& 8.42\\
Scan 10                 & 5.51& 3.67& 4.01& 4.94\\
Scan 11                 & 10.74& 7.51& 6.55& 11.14\\
Scan 12                 & 11.91& 8.84& 9.76& 8.12\\
Average            & 10.89& 6.06& 9.71& \textbf{12.41}\\ \bottomrule
\end{tabular}
}
\end{table}

\subsubsection{Quantitative Comparison}
Table \ref{table:odr} presents the comparison results among the discussed methods based on four evaluation criteria. On average, our method outperforms the other two methods. Compared to FARO SCENE, our method performs worse in the ODR and FNR metrics for outdoor scenes. This is because FARO SCENE excessively removes real points to eliminate more virtual points, an extreme approach that results in the loss of geometric structure information of the point cloud. Regarding the method introduced in \cite{RN64}, our method demonstrates significant advantages in both indoor and outdoor scenes across all four metrics. Overall, the proposed method achieves superior performance, as shown in Table \ref{table:accuracy}. Using the proposed reflective region detection algorithm and the combination of the RE-LSFH feature descriptor and the Hausdorff distance, our method further enhances the accuracy by 5.65\% compared to \cite{RN64}.

Table \ref{table:snr} illustrates a comparison of their effectiveness in enhancing point cloud quality through the computation of the proposed SNR values. The results indicate that the proposed method yields the highest and solely positive improvement in SNR. In contrast, the other two methods somewhat diminish the quality of the point cloud, while the proposed method excels in maintaining the original point cloud data.

\subsection{Ablation Study}
We conduct ablation studies on our designs in the proposed algorithm. The ablation studies involve two main designs: the proposed algorithm for detecting reflective regions and the combination of feature descriptors and distance metrics for identifying virtual points.

The effectiveness of detecting reflective regions greatly influences the accuracy of virtual point detection. To prove the effectiveness of our proposed reflective region detection algorithm, we used the F-measure score \citep{RN64} to quantitatively show the superiority of our method in identifying reflective region points. The results are presented in Table \ref{table:ablation F}. It is evident that our algorithm significantly outperforms the one described in \cite{RN64} in terms of both precision and the F-measure. The traditional algorithm for detecting reflective regions relied on a simplified assumption that laser beams are backscattered only once with single return against non-reflective areas. However, in practical situations, non-reflective areas could produce multiple laser echoes, leading to numerous points from these regions being mistakenly identified as reflective region points.

To assess the efficacy of the proposed RE-LSFH feature descriptors combined with the Hausdorff distance, we performed an ablation study on various combinations of feature descriptors and distance metrics. As illustrated in Table \ref{table:ablation acc} and Table \ref{table:ablation snr}, we examined three distinct combinations of feature descriptors and distance metrics and evaluated their performance in terms of accuracy and SNR.

\begin{table}[htbp]
\centering
\caption{Ablation study on the proposed reflective region detection algorithm}
\label{table:ablation F}
\resizebox{\linewidth}{!}{
\begin{tabular}{@{}cccc@{}}
\toprule
Method   & Precision & Recall & F-measure \\ \midrule
\cite{RN64}        &0.2055           &0.5167        &0.2163           \\
Proposed           &0.7758           &0.8347        &0.7803           \\ \bottomrule
\end{tabular}
}
\end{table}
\begin{table}[t]
\centering
\caption{Ablation study on the accuracy of components within virtual point detection and removal module. The best accuracy is in \textbf{bold}.}
\label{table:ablation acc}
\resizebox{\linewidth}{!}{
\begin{tabular}{@{}ccccc@{}}
\toprule
\multirow{2}{*}{} & \multicolumn{3}{c}{\textbf{Accuracy}(\%)}                              \\ \cmidrule(l){2-4} 
                  & \begin{tabular}[c]{@{}c@{}}FPFH+Helinger\end{tabular} & \begin{tabular}[c]{@{}c@{}}RE-LSFH+Helinger\end{tabular} & \begin{tabular}[c]{@{}c@{}}RE-LSFH+Hausdorff\end{tabular} \\ \midrule
Scan 01                 &94.73 & 95.04                    &96.37             \\
Scan 02                 &98.19 & 98.18                    &98.02             \\
Scan 03                 &99.16 & 99.03                    &99.33             \\
Scan 04                 &91.01 & 91.20                    &94.98             \\
Scan 05                 &89.91 & 92.11                    &95.74             \\
Scan 06                 &96.12 & 96.50                    &97.01             \\
 Scan 07&91.85 &93.35 &92.10 \\
 Scan 08&95.38 &96.05 &96.46 \\
 Scan 09&85.81 &86.44 &86.66 \\
 Scan 10&72.42 &74.12 &75.00 \\
 Scan 11&88.70 &92.97 &92.90 \\
 Scan 12&84.92 &86.34 &85.53 \\
Average           &90.68 & 91.78           &\textbf{92.51}                      \\ \bottomrule
\end{tabular}
}
\end{table}
\begin{table}[t]
\centering
\caption{Ablation study on the SNR of components within virtual point detection and removal module. The best SNR is in \textbf{bold}.}
\label{table:ablation snr}
\resizebox{\linewidth}{!}{
\begin{tabular}{@{}ccccc@{}}
\toprule
\multirow{2}{*}{} & \multicolumn{3}{c}{\textbf{SNR(dB)}}                              \\ \cmidrule(l){2-4} 
                  & \begin{tabular}[c]{@{}c@{}}FPFH+Helinger\end{tabular} & \begin{tabular}[c]{@{}c@{}}RE-LSFH+Helinger\end{tabular} & \begin{tabular}[c]{@{}c@{}}RE-LSFH+Hausdorff\end{tabular} \\ \midrule
Scan 01                 &11.91 & 12.17                    &13.52             \\
Scan 02                 &17.32 & 17.31                    &16.93             \\
Scan 03                 &20.69 & 20.04                    &21.64             \\
Scan 04                 &8.75 & 8.85                    &11.28             \\
Scan 05                 &9.03 & 10.09                    &12.77             \\
Scan 06                 &13.71 & 14.15                    &14.85             \\
 Scan 07&10.82 &11.70 &10.96 \\
 Scan 08&13.16 &13.84 &14.32 \\
 Scan 09&8.15 &8.35 &8.42 \\
 Scan 10&4.52 &4.80 &4.94 \\
 Scan 11&9.12 &11.18 &11.14 \\
 Scan 12&7.94 &8.38 &8.12 \\
Average           &11.26 & 11.74           &\textbf{12.41}                      \\ \bottomrule
\end{tabular}
}
\end{table}

The Hellinger distance is utilized as the distance metric for both the first and second groups, with the feature descriptors being the traditional FPFH feature and the newly proposed RE-LSFH feature respectively. By analyzing the experimental data from the first and second groups, it is evident that the proposed RE-LSFH feature descriptor exhibits superior feature discrimination in matching mirror symmetry points, thanks to the inclusion of the mirror symmetry property. Additionally, to validate the effectiveness of the introduced Hausdorff distance, we kept the feature descriptor constant in the second and third experimental sets and conducted an ablation study on the distance metric. The results indicate that the Hausdorff distance outperforms the Helinger distance. This improvement is attributed to the presence of significant deformed reflection noise in the test scenes, which the Hausdorff distance effectively mitigates.

\section{Conclusion}
This study focuses on removing reflection noise in TLS 3D point clouds. We present, for the first time, a reflection plane detection algorithm based on geometry-optical principles, utilizing the physical characteristics embedded in the TLS point clouds. This algorithm detects points in the reflective area by analyzing the optical reflection model. By processing the point cloud features, our research provides insights and recommendations for future studies. To achieve reflection-invariant features, we propose the RE-LSFH descriptor. In real-world scenarios, reflective surfaces are often neither smooth nor flat, causing significant interference. The use of the RE-LSFH descriptor and the Hausdorff distance has reduced the impact of similar/symmetrical building structures and irregular reflective surfaces on the virtual point detection algorithm.

We constructed a real-world point cloud dataset 3DRN consisting of 12 TLS point cloud data, including indoor and outdoor environments. These environments are heavily affected by reflection noise, contain complex glass structures with various levels of transparency and different shapes, and have highly repetitive and symmetrical features, making them particularly difficult. The experiments performed on this dataset indicate that the proposed approach can directly obtain dense and reliable reflective area points in 3D space based on optical physical properties. The approach improves the precision and recall of reflection area points by 57.03\% and 31.80\% respectively, enabling accurate and complete estimation of reflection surface. Furthermore, the experiments show that the proposed algorithm significantly surpasses the leading method in both qualitative and quantitative evaluation results, achieving 9.17\% and 5.65\% increase in ODR and accuracy respectively.

Future research could focus on the multiple reflection caused by thick glass. In addition, removing reflection noise from curved glass is also a big challenge.

\section*{CRediT authorship contribution statement}
\textbf{Li Fang:}Writing – review \& editing, Writing – original draft, Methodology, Supervision, Project administration, Funding acquisition. \textbf{Tianyu Li: }Writing – review \& editing, Writing – original draft, Methodology, Software, Validation. \textbf{Yanghong Lin: }Writing – review \& editing, Writing – original draft, Methodology, Software, Validation, Visualization. \textbf{Shudong Zhou: }Writing – review \& editing, Investigation, Data Curation, Formal analysis, Visualization. \textbf{Wei Yao: }Writing – review \& editing, Writing – original draft, Methodology, Supervision, Funding acquisition. 

\section*{Declaration of competing interest}
The authors declare that they have no known competing financial interests or personal relationships that could have appeared to influence the work reported in this paper.

\section*{Acknowledgements}
The research was supported by the National Natural Science Foundation of China (NSFC) under Grant number 42101359 and 42171361. 



\bibliographystyle{elsarticle-harv} 
\bibliography{ref}





\end{CJK}
\end{document}